\documentclass[11pt]{article}

\usepackage{placeins}
\usepackage[preprint]{acl}

\usepackage{times}
\usepackage{latexsym}
\usepackage{graphicx}
\usepackage{fvextra}
\usepackage{subcaption}
\usepackage{placeins}

\usepackage{listings}
\usepackage[T1]{fontenc}
\usepackage{morefloats}

\usepackage[utf8]{inputenc}

\usepackage{microtype}

\usepackage{array}
\usepackage{tabularx}
\usepackage[table]{xcolor}  
\usepackage{colortbl}
\usepackage{tcolorbox}
\usepackage{enumitem}

\usepackage{inconsolata}

\usepackage{graphicx}
\usepackage{booktabs}
\usepackage{multirow}
\usepackage{array}
\usepackage{ragged2e}

\usepackage[english,bidi=default]{babel}

\usepackage{cuted}
\tcbuselibrary{listings,breakable}
\ifxetex\babelfont{rm}{TeXGyreTermesX}\fi
\ifluatex\babelfont{rm}{TeXGyreTermesX}\fi

\babelprovide[import]{arabic}
\babelfont[*arabic]{rm}[
  Path=fonts/,              
  UprightFont = Vazirmatn-Regular.ttf,
  BoldFont    = Vazirmatn-Bold.ttf 
]{Vazirmatn}

\newcolumntype{C}[1]{>{\centering\arraybackslash}m{#1}}   
\newcolumntype{L}[1]{>{\raggedright\arraybackslash}m{#1}} 
\newcolumntype{J}[1]{>{\justifying\arraybackslash}m{#1}}

\newcolumntype{E}{>{\centering\arraybackslash\columncolor{enrow}}X}  
\newcolumntype{F}{>{\centering\arraybackslash\columncolor{farow}}X}  
\newcolumntype{A}{>{\centering\arraybackslash\columncolor{arow}}X}   
\newcolumntype{Y}{>{\centering\arraybackslash\columncolor{clrow}}X}  

\definecolor{ovrow}{HTML}{E0F2F1} 
\newcolumntype{O}{>{\centering\arraybackslash\columncolor{ovrow}}X}

\definecolor{enrow}{RGB}{204, 226, 240}  
\definecolor{farow}{RGB}{250, 236, 204}  
\definecolor{clrow}{RGB}{204, 236, 227}  
\definecolor{arow}{RGB}{246, 223, 204}  

\definecolor{enline}{RGB}{190,210,240}
\definecolor{faline}{RGB}{210,190,240}
\definecolor{clline}{RGB}{190,230,210}

\tcbuselibrary{skins, breakable, theorems, listings, hooks}

\tcbset{
  promptstyle/.style={
    enhanced,
    breakable,
    float*=h,
    boxrule=1.5pt,
    arc=2mm,
    outer arc=2mm,
    width=\textwidth,
    left=4pt,right=4pt,top=4pt,bottom=4pt,
    before skip=4pt,after skip=4pt,
    fonttitle=\bfseries
  },
  responsestyle/.style={
    enhanced,
    breakable,
    float*=h,
    boxrule=2pt,
    arc=2mm,
    outer arc=1.5mm,
    colback=gray!3,
    colframe=gray!45,
    colbacktitle=gray!20,
    coltitle=black,
    fonttitle=\bfseries\footnotesize,
    width=\textwidth,
    left=4pt,right=4pt,top=4pt,bottom=4pt,
    before skip=4pt,after skip=4pt
  },
  gtstyle/.style={
    colback=pink!50,       
    colframe=pink!60!black, 
    arc=2mm,                
    boxrule=1pt,
    left=5pt, right=5pt, top=3pt, bottom=3pt,
    fonttitle=\bfseries\small,
    title=Ground Truth,     
    before skip=10pt        
  }
}

\newtcolorbox{PromptEN}[2][]{promptstyle,
  colback=enrow,
  colframe=enrow!50!black,
  colbacktitle=enrow!60!black,
  coltitle=black,
  title={#1},
  #2
}

\newtcolorbox{PromptGen}[2][]{promptstyle,
  colback=white,
  colframe=enrow!50!black,
  colbacktitle=enrow!60!black,
  coltitle=black,
  title={#1},
  #2
}

\newtcolorbox{PromptFA}[2][]{promptstyle,
  colback=farow,
  colframe=farow!50!black,
  colbacktitle=farow!60!black,
  coltitle=black,
  title={#1},
  #2
}

\newtcolorbox{PromptAR}[2][]{promptstyle,
  colback=arow,
  colframe=arow!50!black,
  colbacktitle=arow!60!black,
  coltitle=black,
  title={#1},
  #2
}

\newtcolorbox{PromptCL}[2][]{promptstyle,
  colback=clrow,
  colframe=clrow!50!black,
  colbacktitle=clrow!60!black,
  coltitle=black,
  title={#1},
  #2
}

\newtcblisting{ModelResponse}[2][]{responsestyle,
  title={Model: #2},
  listing engine=listings,
  listing only,
  breakable,
  listing options={
    escapeinside={(@}{@)},
    breaklines=true,
    breakatwhitespace=false,
    columns=fullflexible
  },
  #1
}

\title{\protect\raisebox{-0.25em}{\protect\includegraphics[height=1.3em]{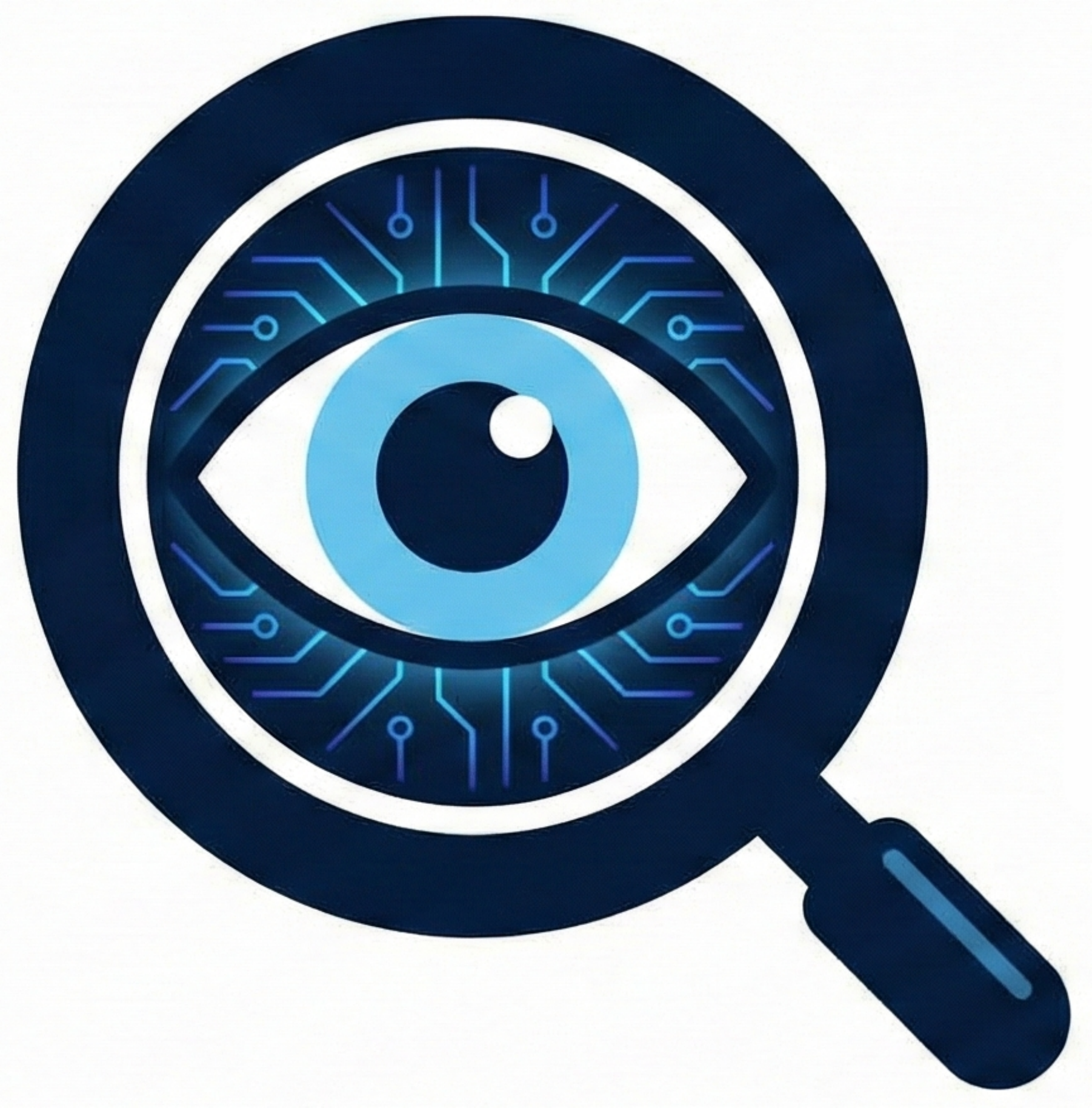}}
    \benchmark{}: A Multilingual Benchmark for Visual Word Puzzle Solving and Image-to-Phrase Reasoning 
    }

\newcommand{\SameEqual}{\textsuperscript{\protect\hyperlink{equal}{*}}}

\author{
  Ali Najar\textsuperscript{1},
  Alireza Mirrokni\textsuperscript{1},
  Arshia Izadyari\textsuperscript{1}\thanks{\protect\hypertarget{equal}{These authors contributed equally.}},
  Sadegh Mohammadian\textsuperscript{1}\SameEqual, \\
  \textbf{Amir Homayoon Sharifizade\textsuperscript{1}\SameEqual,
  Asal Meskin\textsuperscript{1}\SameEqual,
  Mobin Bagherian\textsuperscript{1}\SameEqual,
  Ehsaneddin Asgari\textsuperscript{2}} \\
  \textsuperscript{1}Computer Engineering Department, Sharif University of Technology, Iran \\
  \textsuperscript{2}Qatar Computing Research Institute (QCRI), HBKU, Qatar \\
  \texttt{\{ali.najjar82, alireza.mirrokni28, arshia.izd01, sadegh.mohammadian01\}@sharif.edu} \\
  \texttt{\{amirhoma.shrfzde01, asal.meskin82, mobin.bagherian01\}@sharif.edu, easgari@hbku.edu.qa}
}

\newcommand{\benchmark}{\textsc{Eye-Q}}

\begin{document}
\maketitle

\renewcommand{\thefootnote}{\fnsymbol{footnote}}
\setcounter{footnote}{2} 
\footnotetext{The code is publicly available at \href{https://github.com/llm-lab-org/Eye-Q}{https://github.com/llm-lab-org/Eye-Q}.}


\begin{abstract}
Vision-Language Models (VLMs) have achieved strong performance on standard vision–language benchmarks, yet often rely on surface-level recognition rather than deeper reasoning. We propose visual word puzzles as a challenging alternative, as they require discovering implicit visual cues, generating and revising hypotheses, and mapping perceptual evidence to non-literal concepts in ways that are difficult to solve via literal grounding, OCR-heavy shortcuts, or simple retrieval-style matching. We introduce \benchmark{}, a multilingual benchmark designed to assess this form of complex visual understanding. \benchmark{} contains 1,343 puzzles in which a model observes a conceptually dense scene with a brief description and must infer a specific target word or phrase. The puzzles are intentionally unstructured and cue-implicit, with distractors and contextual relationships that demand selective attention, abstraction, and associative inference. The benchmark spans English, Persian, Arabic, and cross-lingual puzzles. We evaluate state-of-the-art VLMs using an open-ended, human-aligned protocol that probes hypothesis formation and revision under lightweight assistance. Results reveal substantial performance gaps, especially on abstract and cross-lingual puzzles, highlighting limitations in current models’ ability to construct and search over appropriate conceptual representations for flexible image-to-phrase inference; maximum accuracy reaches only 60.27\%. \benchmark{} is publicly available at \href{https://huggingface.co/datasets/llm-lab/Eye-Q}{https://huggingface.co/datasets/llm-lab/Eye-Q}.
\end{abstract}

\section{Introduction}

\begin{figure*}[t!]
    \centering
    \includegraphics[width=1\linewidth]{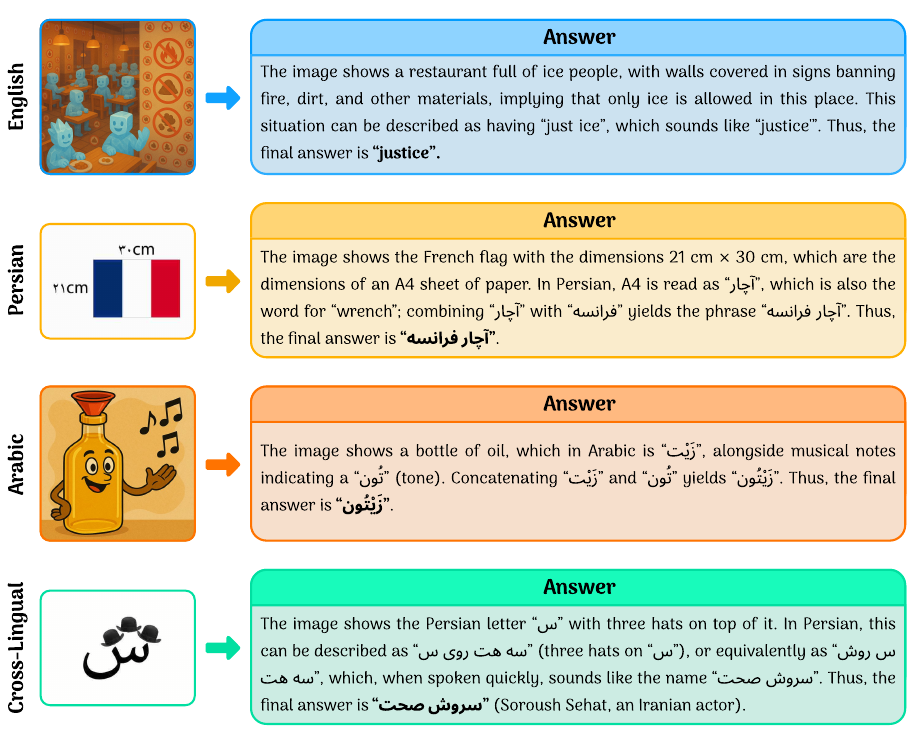}
    \caption{Illustrative \benchmark{} examples from the English, Persian, Arabic, and cross-lingual subsets. Each example shows an input image along with a textual derivation that leads to the target word or phrase.}
    \label{fig:examples_fig}
\end{figure*}

Vision-language models (VLMs) have rapidly evolved into general-purpose assistants that can describe images, answer questions, and follow complex visual instructions \citep{xai2025grok4_modelcard, bai2023qwenvlversatilevisionlanguagemodel, comanici2025gemini25pushingfrontier, openai2025gpt52systemcard, touvron2023llamaopenefficientfoundation}. As these systems become more capable and widely deployed, benchmarking them becomes increasingly important. Yet many existing benchmarks rely on questions with explicit, directly queryable evidence, and a large fraction of their content (or close variants) is available on the web or in widely circulated sources. This can let models succeed via surface recognition, shortcut heuristics, or training-data exposure rather than the reasoning we aim to measure \citep{agrawal-etal-2022-assume, si-etal-2022-language, deng-etal-2024-investigating, li-etal-2024-open-source, li-etal-2024-latest-eval}. A natural next step is to push toward evaluations that are harder to solve by appearance matching or retrieval. In particular, a challenging benchmark should be \textbf{(i) multimodal} in how it encodes meaning, \textbf{(ii) multi-step} in a way that supports hypothesis testing rather than one-shot guessing, and \textbf{(iii) multi/cross-lingual}, measuring performance beyond English scripts, while presenting these criteria in a creative and previously unseen manner.

\noindent\textbf{Multimodality.}
A broad suite of multimodal benchmarks probes higher-level cognition beyond perception, including multi-domain understanding \citep{liu2024mmbenchmultimodalmodelallaround}, compositional reasoning \citep{zhu2025testtimematchingunlockingcompositional, lei-etal-2025-scaffolding}, abstract reasoning \citep{chia-etal-2024-puzzlevqa, wüst2025bongardwonderlandvisualpuzzles}, and scientific understanding \citep{Ma2025CoCoVaCO, yin2025ariseadaptiveresolutionawaremetric}. These benchmarks have been valuable, but many tasks still emphasize literal grounding: recognizing entities, reading text, and answering questions with evidence that is localized and directly queryable. Strong performance therefore does not always imply that a model can discover subtle cues in a complex scene or map perceptual evidence to non-literal concepts. This leaves a gap for evaluations where the signal is implicit.

\noindent\textbf{Multi-step Reasoning.}
Many evaluations reward systematic and rule-following reasoning that narrows toward a single correct solution \citep{cropley-2006-covergent}. This gap appears in puzzles such as rebus-like riddles and picture-word games \citep{gritsevskiy2024rebus, columbus2025, khurdula-etal-2025-beyond, chia-etal-2024-puzzlevqa}. Solving these tasks is abductive and iterative: solvers propose hypotheses about which aspects of the scene are cues, test them, and revise when they fail. In this setting, multi-step reasoning alone is insufficient, because without improving the underlying hypothesis, additional steps can amplify unproductive search. Recent puzzle-oriented evaluations underscore that even strong VLMs struggle when success depends on abstraction, symbolic substitution, and non-literal image-to-language associations \citep{khurdula-etal-2025-beyond, gritsevskiy2024rebus, columbus2025, lee-etal-2025-puzzled, ren2025vgrpbench, guetta2024visual}. However, many of these works rely on visually clean compositions or explicit text, making success largely a matter of OCR and shallow heuristics. Moreover, protocols can under-measure models when they do not allow hypothesis revision. This suggests a need for puzzle-style evaluations that are visually dense, cue-implicit, and paired with a human-aligned setup that supports controlled refinement without changing the task.

\noindent\textbf{Multilinguality.}
Multilinguality further compounds the challenge. While multilingual vision--language resources and evaluations have grown \citep{liu2021marvl, Leng2025CrossWordBenchET, Fabbri2025MultiNRCAC}, puzzle solving remains under-studied beyond English, particularly for non-Latin scripts and culturally grounded expressions. Puzzles often hinge on phonetic, orthographic, and idiomatic mechanisms that do not translate cleanly, thereby offering a rigorous test of reasoning across linguistic contexts. Uniquely, our work introduces the first systematic evaluation of visual word puzzles with Persian answers and cross-lingual puzzles that bridge English visual cues with Persian solutions.

\noindent\textbf{Our Approach.}
We introduce \benchmark{} (Figure~\ref{fig:examples_fig}), a multilingual benchmark for visual word puzzle solving and open-ended image-to-phrase inference. The task is designed to require implicit cue discovery in dense scenes and supports hypothesis revision via lightweight, human-aligned assistance. Even with these supports, performance across proprietary VLMs remains low, especially in cross-lingual settings, highlighting persistent gaps in non-literal vision--language reasoning.

\paragraph{Contributions.}
Our work makes the following contributions:
\begin{itemize}
\item \textbf{Task Formulation.} We introduce visual word puzzle solving as a vision--language reasoning task requiring multi-modal cue integration, multi-step reasoning and search, and multilingual/cross-lingual generalization.

\item \textbf{Benchmark Dataset.} We introduce a multilingual benchmark of 1,343 puzzles spanning English, Persian, Arabic, and cross-lingual settings.

\item \textbf{Cue-Implicit Design.} The puzzles are cue-implicit, unstructured, and conceptually dense, avoiding token-explicit cues (OCR or templated layouts) and requiring abstraction and associative reasoning.

\item \textbf{Evaluation Protocol.} We propose a human-aligned, open-ended evaluation protocol that probes hypothesis formation and revision through lightweight assistance (answer-length hints, partial character reveals, few-shot examples, iterative refinement).

\item \textbf{Empirical Findings.} We benchmark state-of-the-art VLMs and reveal substantial performance gaps, especially on abstract and cross-lingual puzzles, highlighting limits in current models’ ability to construct and search over conceptual representations.
\end{itemize}
\section{Related Work}
\label{sec:related}

\paragraph{VLM Benchmarks Beyond Literal Grounding.}
Vision--language evaluation has long focused on VQA and captioning tasks with explicit, directly grounded evidence \citep{goyal2017making, hudson2019gqa, chen2015microsoft}. Newer benchmarks target compositional or multi-step reasoning and more abstract visual understanding \citep{Kil2024IIMMRIA, Zhou2025ReasoningIA, chia-etal-2024-puzzlevqa}. Yet multiple analyses show that high performance can reflect language priors or dataset shortcuts rather than robust grounding \citep{Yang2025NewDA, Krojer2025ASV, Chi2025ChimeraDS}, motivating evaluations that require implicit cue discovery and non-literal inference.

\paragraph{Visual Wordplay and Rebus-Style Puzzles.}
Puzzle-style benchmarks probe non-literal image-to-phrase mappings. Rebus and visual wordplay datasets encode short phrases via symbolic substitution, layout, and phonetic or semantic composition \citep{gritsevskiy2024rebus, columbus2025, khurdula-etal-2025-beyond, Das2025RebusBUS}. Relevant work studies visual riddles and puns, including culturally grounded variants \citep{guetta2024visual, Chung2024CanVLA, Zhang2024CreatingALA}. However, many settings use clean, token-like cues (text/icons) and analyses report weak hint use and limited hypothesis revision \citep{lee-etal-2025-puzzled, Movva2025ReasoningRHA}.

\paragraph{Hinted and Iterative Evaluation Protocols.}
A growing literature evaluates reasoning under iterative protocols where models revise answers using feedback or critique. Representative methods include self-refinement and reflection-style loops \citep{Shinn2023ReflexionLAA, Madaan2023SelfRefineIRA} and critique-based verification with tools \citep{Gou2023CRITICLLA}. Benchmarks also measure critique-and-correct behavior directly and in multi-turn settings \citep{Wu2024VISCO, Lin2024CriticBenchBLA, Banatt2024WILTAMA}, while other studies emphasize that self-correction is unreliable without strong verifiers and careful reporting \citep{Huang2023LargeLMA, Kamoi2024WhenCLA}. Our evaluation uses lightweight assistance to separate cue-discovery failures from output brittleness.

\paragraph{Multi/Cross-Lingual VLM Evaluation.}
Although most benchmarks are English-centric, recent work expands evaluation across languages and cultures through multilingual suites, cross-lingual VQA, and culturally grounded reasoning datasets \citep{liu2021marvl, pfeiffer-etal-2022-xgqa, Bugliarello2022IGLUE, Romero2024CVQACMA}. Complementary efforts broaden coverage for underrepresented regions and non-Latin scripts, including Arabic-focused resources \citep{Barua2024ChitroJeraARA, Alwajih2024PeacockAFA, Alwajih2025PearlAMA}. Still, multilingual multimodal puzzle solving remains limited, particularly for culturally grounded wordplay and cross-lingual transformations into Persian.

\paragraph{Position of \benchmark{}.}
\benchmark{} connects these threads by evaluating open-ended image-to-phrase inference for non-literal puzzles in cue-implicit, distractor-rich scenes, and by extending the setting to English, Persian, Arabic, and cross-lingual puzzles.

\section{\benchmark{}}
\label{sec:eye-Q}
This section introduces \benchmark{} and the design principles behind its puzzles. We first summarize the benchmark composition and instance format, then describe puzzle design and the data sources used for the monolingual and cross-lingual subsets.

\subsection{Benchmark Overview}
\benchmark{} is a multimodal benchmark designed to evaluate open-ended image-to-phrase reasoning under cue-implicit visual settings and multilingual wordplay. The benchmark contains 1{,}343 picture puzzles organized into four subsets: English, Persian, Arabic, and cross-lingual (Table~\ref{tab:dataset_stats}).

Each puzzle consists of an image and a short textual prompt that specifies the game rule (i.e., the image encodes a hidden target word or short phrase) and requests a single final answer. The task is open-ended: models must generate the target word/phrase rather than select from candidates.

The images are intentionally cue-implicit and conceptually dense: scenes contain multiple objects, attributes, and relationships that may function as semantic cues, often alongside distractors. Cues can involve object orientation, item counts, color, relative size, spatial relations (e.g., containment, overlap, inversion), material properties, posture, and affect. Text may appear in some images, but \benchmark{} is designed to be anti-OCR: solving rarely reduces to simply reading visible text.

\begin{table}[t]
  \centering
  \small
  \setlength{\tabcolsep}{4pt}
  \renewcommand{\arraystretch}{1.2}

  \resizebox{\linewidth}{!}{%
    \begin{tabular}{lccc}
      \rowcolor{black!70}
      \textcolor{white}{\textbf{Subset}} &
      \textcolor{white}{\textbf{\# Puzzles}} &
      \textcolor{white}{\textbf{Source}} \\
      \toprule
      \rowcolor{enrow}
      English       & 300   & Our synthetic puzzles \\
      \rowcolor{farow}
      Persian       & 671   & \textit{Aftabe} game   \\
      \rowcolor{arow}
      Arabic       & 50     & Our synthetic puzzles   \\
      \rowcolor{clrow}
      Cross-lingual & 322   & \textit{Aftabe} game   \\
      \midrule
      Total         & 1{,}343      & --                     \\
      \bottomrule
    \end{tabular}%
  }

  \caption{Basic statistics for \benchmark{}.}
  \label{tab:dataset_stats}
\end{table}

\subsection{Puzzle Design}
The core of \benchmark{} lies in requiring a semantic leap from literal recognition to a non-literal linguistic target. Unlike standard VQA tasks that ask ``What is in the image?'', our puzzles ask ``What does this scene imply or represent?''. Solving typically requires: (i) \textit{cue discovery} (identifying which elements are informative versus distractors), (ii) \textit{relational abstraction} (reasoning over relations and transformations rather than isolated objects), and (iii) \textit{linguistic association} (mapping the inferred concept to an idiom, pun, phonetic resemblance, or conventional phrase in the target language).

Puzzles range from monolingual wordplay to cross-lingual reasoning. In cross-lingual instances, cues naturally suggest words or concepts in one language, while the correct answer is expressed in the other, requiring phonetic, orthographic, or semantic bridging between English and Persian. Figure~\ref{fig:examples_fig} provides representative examples.

\begin{figure*}[t!]
    \centering
    \includegraphics[width=1\linewidth]{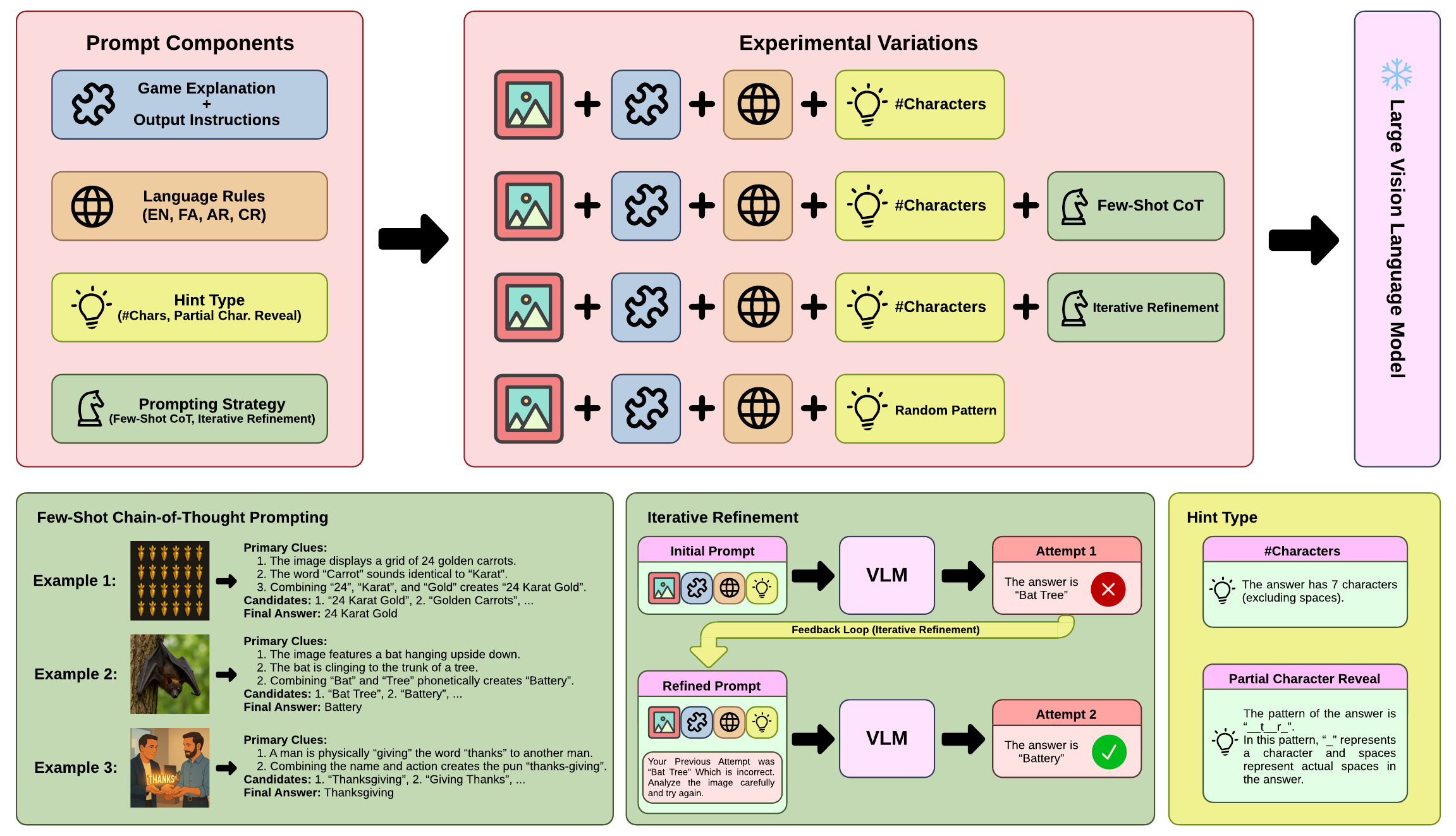}
    \caption{Overview of experimental variations in \benchmark{}. We instantiate prompts by combining a shared base template, game explanation and subset-specific language rules, with optional hinting and prompting strategy modules.}
    \label{fig:experiments_var}
\end{figure*}

\subsection{Data Collection and Ground Truth}
\label{subsec:data_collection}
\benchmark{} is built from two sources: (i) the mobile game \textit{Aftabe} (Appendix~\ref{aftabe_mobile_game}), and (ii) English and Arabic puzzles designed by the authors and rendered into images using text-to-image generation models. We have obtained permission from \textit{Aftabe} to use and redistribute the curated puzzles as part of \benchmark{}.

\paragraph{Persian and Cross-Lingual Puzzles.}
We curated Persian and cross-lingual puzzles from \textit{Aftabe}, which contains visual puzzles with Persian answers. For each selected instance, we preserve the original image and the game-provided intended solution as ground truth. Because puzzles can admit multiple interpretations, we manually filter out items that are ambiguous or under-specified, retaining puzzles with a clear intended reading and a unique target answer. We then categorize each retained puzzle as Persian or cross-lingual based on whether solving requires an English--Persian bridge.

\paragraph{English and Arabic Puzzles.}
We build the English and Arabic subsets by first selecting a target word or short phrase and writing a concise scene description intended to lead a human solver to that target. We then render the scene using text-to-image models (GPT~\citep{openai2025gpt52systemcard} and Nano Banana~\citep{raisinghani2025nanobananapro}). All generated images are manually reviewed, and we retain only those that clearly support the intended solution and do not contain rendering artifacts or alternative plausible interpretations. Because the target answer is fixed during design and each image is filtered for clarity, these instances have controlled ground-truth labels that are verified to be consistent with the visual content.

\paragraph{Evaluation Hints.}
Lightweight assistance signals such as answer-length hints and partial character reveal are not separate annotation fields in \benchmark{}. However, they are directly derivable from the ground-truth answer (e.g., character count or revealing a fixed fraction of characters), so any user of the dataset can apply the same hinting schemes. In our experiments (Section~\ref{sec:experiments}), we use these derived hints as controlled evaluation variants to provide human-aligned assistance without changing the underlying task.

\begin{figure*}[t!]
    \centering
    \includegraphics[width=1\linewidth]{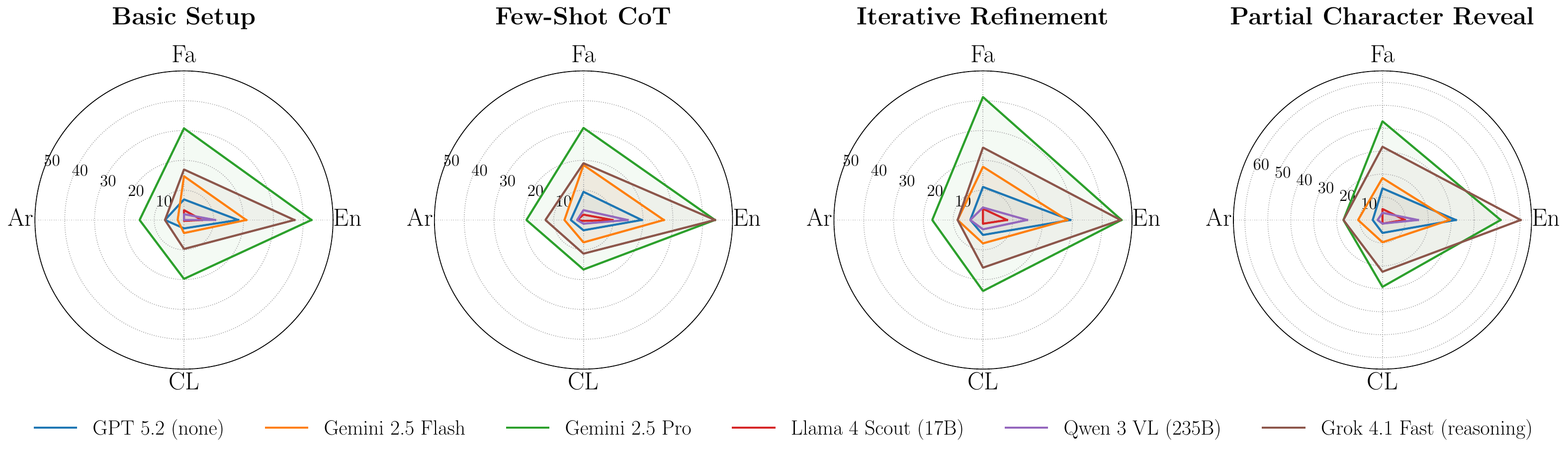}
    \caption{Accuracy(\%) of six LVLMs across four prompt variants and four language subsets.}
    \label{fig:accuracy_results}
\end{figure*}

\section{Experiments}
\label{sec:experiments}

This section evaluates vision--language models on \benchmark{}. We compare multiple models under four prompt variants and report accuracy across the benchmark’s language subsets. We include both proprietary and open-source LVLMs. Our goal is to measure open-ended puzzle solving under a human-aligned setup that supports hypothesis formation and revision, while keeping the underlying task unchanged.

\subsection{Models}
\label{subsec:models}
We evaluate a set of proprietary and open-source LVLMs: GPT-5.2~\citep{openai2025gpt52systemcard}, Gemini~2.5~Flash~\citep{comanici2025gemini25pushingfrontier}, Gemini~2.5~Pro~\citep{comanici2025gemini25pushingfrontier}, Grok~4.1~Fast reasoning~\citep{xai2025grok4_modelcard}, Llama~4~Scout~\citep{touvron2023llamaopenefficientfoundation}, and Qwen~3~VL~\citep{bai2023qwenvlversatilevisionlanguagemodel}. All experiments use the default inference configuration provided by each model endpoint or release, including each model’s default decoding strategy and default visual preprocessing.

\subsection{Prompt Design and Input Format}
\label{subsec:prompting}
For each puzzle, we query the model with a single image and a text prompt instantiated from a fixed base template, shown in Figure~\ref{fig:experiments_var}. The base template includes: (i) a brief explanation of the game objective, and (ii) subset-specific instructions for English, Persian, Arabic, or cross-lingual puzzles. Across all experiments, we keep the base instructions and output format consistent and vary only the experiment-specific modules described below. We provide the exact prompts used for the experiments in Appendix~\ref{app:prompt_options}.

\paragraph{Decoding and scoring.}
All main results reported in the paper use each model’s default decoding configuration rather than enforcing a shared decoding strategy such as greedy or fixed-temperature sampling. We report exact-match accuracy against the ground-truth answer, using the model's final answer string with only minimal cleanup, specifically trimming whitespace and removing Persian and Arabic diacritics (A'rab).

\paragraph{Experimental Variations.}
Using the shared base template, we evaluate four prompting variants:

\begin{enumerate}
    \item \textbf{Basic.}
    We append an orthographic hint indicating the target answer length in characters to narrow the search space, and ask the model to output a single final answer.

    \item \textbf{Few-Shot Chain-of-Thought.}
    We prepend three solved examples to the prompt. Each demonstration includes an image, the correct answer, and a concise description of the intended image--answer relation. Demonstrations are sampled from the same subset pool as the test puzzle, English, Persian, Arabic, or cross-lingual, and exclude the test instance. For fair comparison, the selected demonstration set is fixed across models for each evaluated sample.

    \item \textbf{Iterative Refinement.}
    We first query the model using the basic prompt. If the answer is incorrect, we append the model’s previous guess and a short instruction to revise, then re-query. We allow up to two such revisions, for a total of three attempts. A puzzle is counted as correct if any attempt matches the ground truth.

    \item \textbf{Partial Character Reveal.}
    We reveal a randomly selected 25\% subset of the ground-truth answer’s non-space characters and mask the remainder, for example with underscores, then ask the model to output the full answer. The revealed positions are sampled with a fixed random seed to ensure reproducibility.
\end{enumerate}

Across all variants, the input image is unchanged and the core task remains open-ended image-to-phrase inference; only the prompting strategy and lightweight hinting described above are varied.

\section{Results}
\label{sec:results}

Figure~\ref{fig:accuracy_results} reports accuracy for six LVLMs across the four language subsets and four prompt variants described in Section~\ref{subsec:prompting}, following the evaluation protocol in Section~\ref{sec:experiments}. Each value is the normalized exact-match accuracy over all puzzles in a subset, where normalization applies minimal surface-form cleanup (e.g., trimming punctuation/whitespace artifacts and lowercasing when applicable). Exact numerical values for all model--subset--prompt combinations are provided in Appendix~\ref{app:overall_accuracy}. We additionally report controlled temperature sweeps in Appendix~\ref{app:temperature_ablation}.

\subsection{Overall Performance}
Across all models and prompt variants, accuracy remains far from high, indicating that \benchmark{} poses a challenging open-ended image-to-phrase inference problem. Even the best-performing models struggle substantially on the harder subsets. Considering the best result achieved on each subset across all evaluated models and prompt variants, the peak accuracy is 60.27\% on English (Grok~4.1~Fast, Partial Character Reveal), 43.03\% on Persian (Gemini~2.5~Pro, Partial Character Reveal), 19.15\% on Arabic (Gemini~2.5~Pro, Few-Shot CoT), and 29.15\% on cross-lingual puzzles (Gemini~2.5~Pro, Partial Character Reveal). Thus, while English can reach moderate performance under supportive prompting, Arabic and cross-lingual puzzle solving remain particularly difficult even for the strongest models.

Performance differences across model families are also clear. Proprietary models (GPT-5.2, Gemini~2.5, Grok~4.1) dominate overall, while open-source models (Llama~4~Scout, Qwen~3~VL) remain low, including near-zero performance on Arabic and cross-lingual subsets in multiple settings. This highlights persistent weaknesses in multilingual cue discovery and non-literal phrase inference.

\subsection{Assistance Effects}
Prompt variants provide lightweight, human-aligned assistance, but they do not close the gap. Averaged across models and language subsets, iterative refinement improves accuracy over the basic setup from 11.53\% to 15.63\%, suggesting that allowing hypothesis revision can recover from some initial misreadings. Partial character reveal improves further to 18.11\% on average.

These assistance variants also directly reduce a common concern in open-ended evaluation: that low exact-match accuracy might be dominated by output-space brittleness (large lexical search space or superficial near-misses). Partial character reveal constrains the output space without supplying answer candidates, and refinement gives models multiple chances to correct small mistakes without turning the task into candidate verification. Despite this, accuracy remains low even under supportive settings: for example, Arabic stays below 10\% on average even with partial character reveal, and cross-lingual performance remains around 12\% on average. This persistence suggests that the dominant failure mode is upstream of surface form: models often fail to identify which elements in a complex, distractor-rich scene constitute the right cues and to construct the correct abstract interpretation linking the scene to the intended hidden phrase. In this sense, \benchmark{} stresses a different bottleneck than OCR-heavy or visually clean rebus-style settings, where reading explicit tokens and combining them compositionally can account for much of the solution.

\begin{figure}[t!]
    \centering
    \includegraphics[width=1\linewidth]{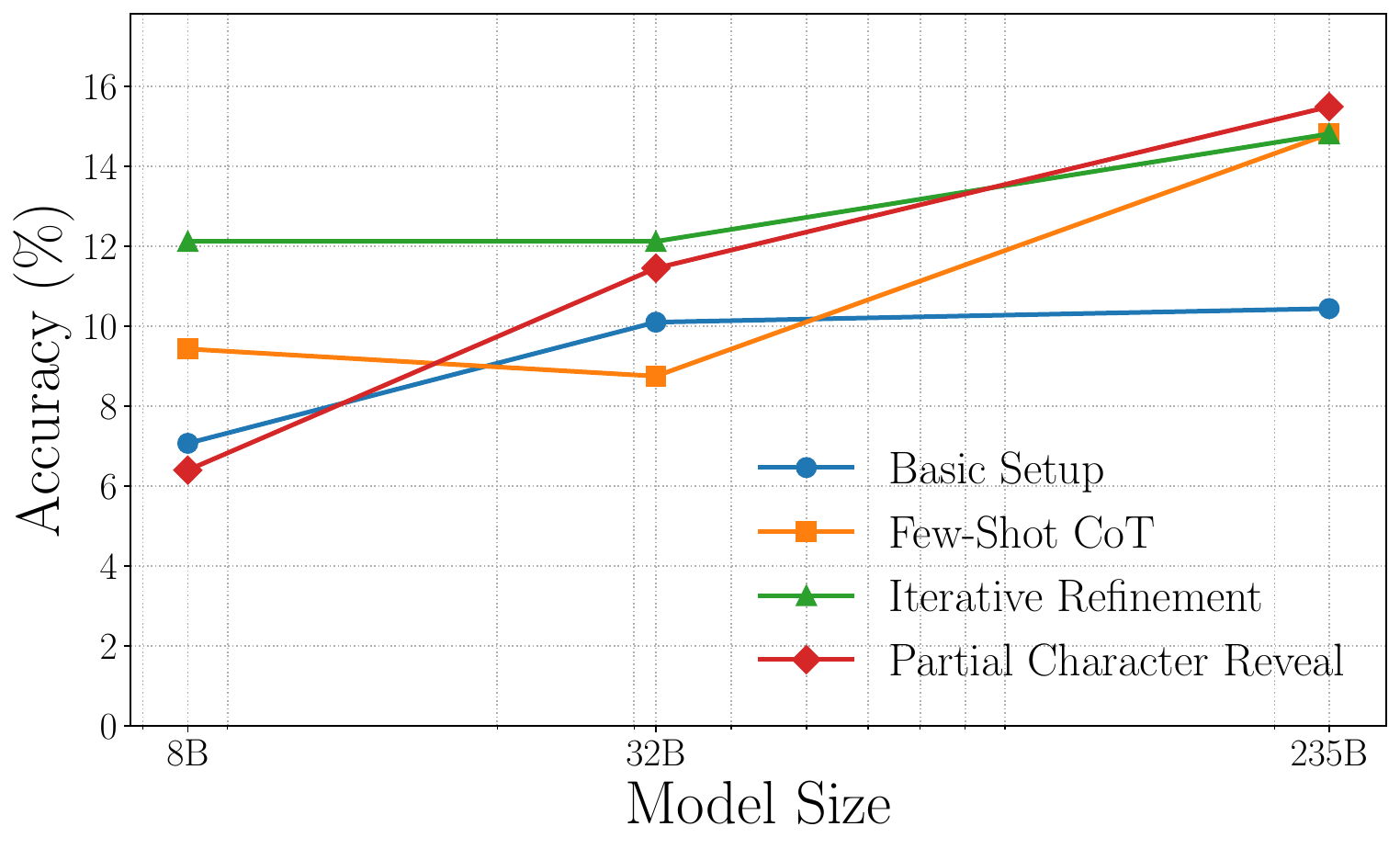}
    \caption{English-subset accuracy versus model size for the Qwen3-VL family (8B, 32B, 235B-A22B) under four prompting variants.}

    \label{fig:acc_vs_size_en}
\end{figure}

\begin{figure}[t!]
    \centering
    \includegraphics[width=1\linewidth]{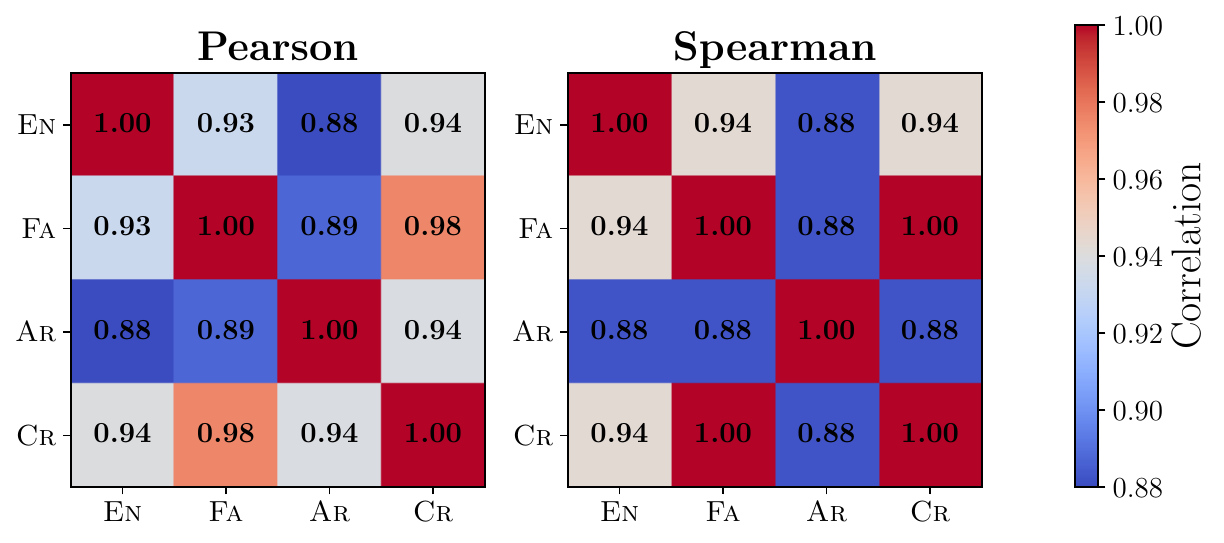}
    \caption{Correlation of model accuracies across language subsets in the Basic setup. Pearson (left) and Spearman (right) correlations are computed across models using subset-level accuracies for English, Persian, Arabic, and cross-lingual puzzles.}
    \label{fig:basic_correlation}
\end{figure}

\begin{figure*}[t!]
    \centering
    \includegraphics[width=1\linewidth]{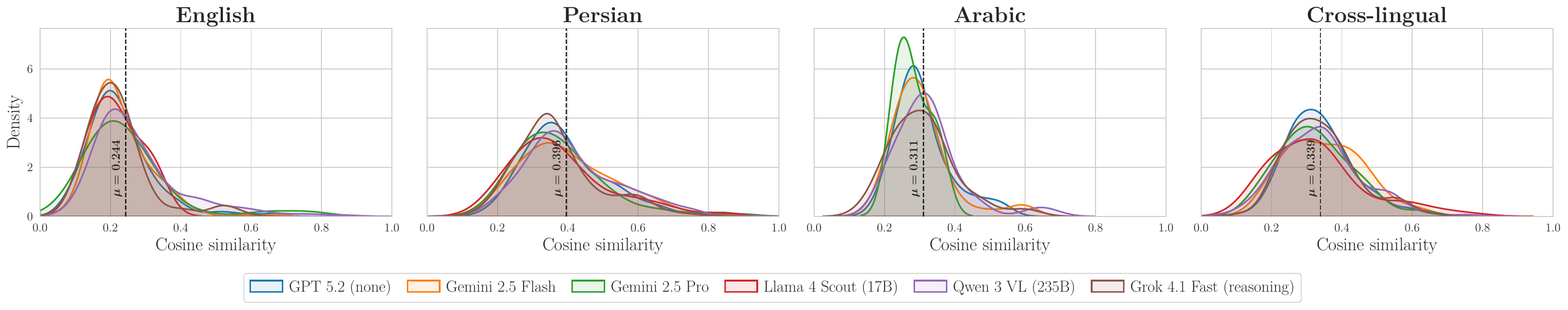}
    \caption{Density estimates of cosine similarity between model outputs and ground-truth answers under iterative refinement. Similarities are computed between each model's final refined answer and the ground-truth answer using OpenAI \texttt{text-embedding-3-large} embeddings~\citep{openai_text_embedding_3_large_docs}. Curves are estimated over failure cases only and shown across English, Persian, Arabic, and cross-lingual subsets; higher values indicate closer semantic alignment. The concentration of mass at low similarity suggests that many errors are not near-miss paraphrases.}
    \label{fig:density_refinement}
\end{figure*}

\subsection{Model Size Effect}
To probe the effect of scale in a controlled, same-family setting, Figure~\ref{fig:acc_vs_size_en} plots accuracy as a function of model size for Qwen3-VL on the English subset under all four prompt variants (Qwen3-VL-8B-Instruct, Qwen3-VL-32B-Instruct, Qwen3-VL-235B-A22B-Instruct). Larger models consistently perform better, and the gains tend to be most visible under assistance variants that encourage revision or constrain the output space. However, even at the largest size, accuracy remains far from saturated, reinforcing that simply scaling parameters is insufficient to overcome the core challenge posed by \benchmark{}: cue discovery and abstraction in complex, cue-implicit scenes.

Results for the same Qwen3-VL scaling analysis on Persian, Arabic, and cross-lingual subsets are provided in Appendix~\ref{app:model_size_effect}.

\subsection{Language Coupling}
\label{sec:correlation_basic}
To test whether model strengths transfer across languages, we analyze language coupling by correlating model accuracies between language subsets in the basic setup. Figure~\ref{fig:basic_correlation} reports Pearson and Spearman correlations computed across models using subset-level accuracies. Correlations are high overall, indicating that relative model ranking is largely consistent across languages.

The tightest coupling is between Persian and cross-lingual performance (Pearson 0.98, Spearman 1.00), suggesting that models strong on Persian puzzles tend also to be strong on cross-lingual puzzles that require bridging Persian and English. In contrast, Arabic shows weaker coupling with other subsets (with correlations around 0.88 in multiple pairings), consistent with Arabic being the most challenging setting for current models and potentially requiring additional robustness beyond what drives performance in English and Persian.

\subsection{Semantic Near-Miss Analysis}
Exact-match evaluation can in principle underestimate performance if models frequently produce answers that are semantically close to the ground truth but differ in surface form. To test this, we measure semantic proximity between model outputs and ground-truth answers using cosine similarity in an embedding space. Specifically, we compute cosine similarity between embeddings of the model's final answer under iterative refinement and the ground-truth answer. We compute this similarity over failure cases only and visualize density estimates across language subsets in Figure~\ref{fig:density_refinement}. Additional plots for other prompting variants show a similar trend and are provided in Appendix~\ref{app:similarity_density}.

The density estimates are concentrated at low similarity values across all four language settings, indicating that incorrect outputs are typically not semantically close to the target phrase. This suggests that low accuracy on \benchmark{} is not primarily driven by minor linguistic differences or near-miss paraphrases penalized by strict string matching. Instead, the dominant failure mode appears to arise earlier: models often fail to identify the right cues amid distractors and to form the correct abstract interpretation linking the complex scene to the intended hidden phrase.

\section{Conclusion}
We introduced \benchmark{}, a multilingual benchmark for visual word puzzle solving and open-ended image-to-phrase inference. \benchmark{} contains 1{,}343 cue-implicit puzzles spanning English, Persian, Arabic, and cross-lingual Persian–English settings. Unlike prior puzzle-style evaluations that often center on visually clean rebus compositions or explicit text cues, \benchmark{} emphasizes visually dense scenes with distractors, where success requires discovering which elements function as clues and composing a non-literal conceptual interpretation.

Across six state-of-the-art LVLMs, performance remains low. Even with human-aligned assistance such as few-shot CoT, iterative refinement, and partial character reveal, models continue to struggle, particularly on Arabic and cross-lingual subsets. A semantic similarity analysis of failures further suggests that errors are typically not near-miss paraphrases, reinforcing that the bottleneck is not only output-space brittleness but also cue selection and abstraction. We hope \benchmark{} will provide a useful testbed for developing and measuring multimodal systems that better support hypothesis generation, revision, and multilingual reasoning beyond literal grounding.

\section*{Limitations}

First, on rare occasions where alternative interpretations exist, \benchmark{} prioritizes the intended ground truth to maintain rigorous evaluation metrics.

Second, dataset generation relies on contributors who are fluent in the target languages. This  can make it harder to scale the benchmark to additional languages, dialects, or low-resource varieties while keeping difficulty and style consistent.

Third, reliable human evaluation is also non-trivial. To assess correctness fairly across languages, each language subset should ideally be evaluated by native speakers or annotators with comparable proficiency. This increases coordination effort and cost, and it can introduce variability between annotator groups, especially when borderline cases depend on subtle linguistic judgments.

If these challenges are addressed, the benchmark's coverage is likely to improve, and its evaluations to become more reliable and reproducible.

\section*{Acknowledgments}
We thank the developers of the \textit{Aftabe} mobile game for creating an engaging collection of visual word puzzles that inspired and enabled the Persian and cross-lingual portions of \benchmark{}.


\bibliography{custom}

\clearpage

\appendix

\label{sec:appendix}

\section{Aftabe Mobile Game}
\label{aftabe_mobile_game}
The Persian and cross-lingual subsets of \benchmark{} are sourced from \textit{Aftabe}, a popular Iranian puzzle game released in 2014. The game achieved viral success in Iran due to its creative use of Persian wordplay and abstract visual riddles. It is available on Google Play at: \url{https://play.google.com/store/apps/details?id=com.freeapks.aftabe}.

\section{Additional Results}
\FloatBarrier
This appendix reports additional results that complement the main experiments and provide further context for the observed trends.

\subsection{Overall accuracy}
\label{app:overall_accuracy}
For completeness, we report the exact accuracies corresponding to Figure~\ref{fig:accuracy_results} in Table~\ref{tab:full_accuaracy_table}.

\begin{table*}[t!]
  \centering
  \renewcommand{\arraystretch}{1.5}
  \setlength{\tabcolsep}{1pt} 
  \scriptsize

  \resizebox{\linewidth}{!}{
    \begin{tabularx}{1.25\linewidth}{
        l
        *{4}{E F A Y} 
    }
      \rowcolor{black!70}
      \textcolor{white}{\textbf{Model}} &
      \multicolumn{4}{c}{\textcolor{white}{\textbf{Basic setup}}} &
      \multicolumn{4}{c}{\textcolor{white}{\textbf{Few-Shot CoT}}} &
      \multicolumn{4}{c}{\textcolor{white}{\textbf{Iterative Refinement}}} &
      \multicolumn{4}{c}{\textcolor{white}{\textbf{Partial Character Reveal}}} \\
      
      & \textbf{En} & \textbf{Fa} & \textbf{Ar} & \textbf{CL} 
      & \textbf{En} & \textbf{Fa} & \textbf{Ar} & \textbf{CL} 
      & \textbf{En} & \textbf{Fa} & \textbf{Ar} & \textbf{CL} 
      & \textbf{En} & \textbf{Fa} & \textbf{Ar} & \textbf{CL} \\
      \toprule

      GPT 5.2 (none)   & 18.18 & 6.90 & \underline{6.38} & 2.82 & 19.53 & 9.45 & 4.26 & 3.45 & 29.29 & 11.09 & 4.26 & 5.02 & 31.99 & 13.79 & 4.26 & 5.64 \\
      Gemini 2.5 Flash & 20.88 & 14.69 & 2.13 & 4.39 & \underline{26.94} & 18.44 & 6.38 & 7.52 & 28.28 & 17.84 & \underline{8.51} & 7.84 & 29.29 & 18.29 & \underline{10.64} & 9.72 \\
      Gemini 2.5 Pro   & \textbf{42.76} & \textbf{30.73} & \textbf{14.89} & \textbf{19.75} & \textbf{44.11} & \textbf{30.88} & \textbf{19.15} & \textbf{16.61} & \textbf{46.46} & \textbf{41.23} & \textbf{17.02} & \textbf{23.82} & \underline{51.52} & \textbf{43.03} & \textbf{17.02} & \textbf{29.15} \\
      Llama 4 Scout (17B) & 6.06 & 3.30 & 0.00 & 0.31 & 9.76 & 1.80 & 2.13 & 0.31 & 8.08 & 3.77 & 0.00 & 1.25 & 9.76 & 4.80 & 0.00 & 1.57 \\
      Qwen 3 VL (235B) & 10.44 & 1.95 & 0.00 & 0.00 & 14.81 & 3.30 & 2.13 & 1.25 & 14.81 & 4.20 & 4.26 & 3.13 & 15.49 & 3.15 & 2.13 & 1.57 \\
      Grok 4.1 Fast (reasoning) & \underline{37.04} & \underline{16.94} & \underline{6.38} & \underline{9.72} & \textbf{44.11} & \underline{19.04} & \underline{12.77} & \underline{11.29} & \underline{46.13} & \underline{24.29} & \underline{8.51} & \underline{15.99} & \textbf{60.27} & \underline{31.93} & \textbf{17.02} & \underline{22.57} \\

      \bottomrule
    \end{tabularx}
  }
  \caption{Accuracy(\%) of models for four prompt types across languages. The best performance in each column is in \textbf{bold} and the second best is \underline{underlined}.}
  \label{tab:full_accuaracy_table}
\end{table*}

\subsection{Temperature Ablation}
\label{app:temperature_ablation}

Table~\ref{tab:temp_ablation} shows that performance is strongly and consistently sensitive to decoding temperature, with clear differences across models. For the most capable model, Gemini~2.5~Pro, accuracy is relatively stable at low temperature but declines steadily as $T$ increases. This monotonic degradation matches the intuition that, in a constrained exact-match evaluation, additional sampling noise mainly perturbs reasoning traces that would otherwise remain correct. Grok~4.1~Fast follows the same overall trend from $T{=}0.01$ to $T{=}1.0$, but drops sharply at $T{=}2.0$, suggesting a distinct failure regime once stochasticity becomes large.

By contrast, several weaker or more generation-sensitive models exhibit an inverted-U pattern: moderate randomness improves results, while higher temperature is harmful. Gemini~2.5~Flash and Qwen~3~VL reach their best performance at $T{=}1.0$, and GPT-5.2 displays a milder version of the same effect. Llama~4~Scout gains slightly at low temperature but breaks down completely at very high $T{=}5.0$. Overall, these findings suggest that modest stochasticity can help less reliable models recover from early incorrect decisions, whereas stronger reasoning-focused models are best evaluated under near-deterministic greedy decoding.

\begin{table}[t!]
    \centering
    \renewcommand{\arraystretch}{1.3} 
    \setlength{\tabcolsep}{5pt}       
    \fontsize{6.6pt}{8pt}\selectfont
    
    \begin{tabular}{l cc cc cc}
        \toprule
        
        \rowcolor{black!70}
        \textcolor{white}{\textbf{Model}} & 
        \textcolor{white}{\textbf{$T$}} & \textcolor{white}{\textbf{Acc (\%)}} & 
        \textcolor{white}{\textbf{$T$}} & \textcolor{white}{\textbf{Acc (\%)}} & 
        \textcolor{white}{\textbf{$T$}} & \textcolor{white}{\textbf{Acc (\%)}} \\
         
        \midrule
        
        \rowcolor{enrow}
        GPT 5.2 (none)      & 0.01 & 18.86 & 1.0 & 19.53 & 2.0 & 17.17 \\
        \rowcolor{enrow}
        Gemini 2.5 Flash    & 0.01 & 23.91 & 1.0 & \underline{26.94} & 2.0 & \underline{23.91} \\
        \rowcolor{enrow}
        Gemini 2.5 Pro      & 0.01 & \textbf{47.14} & 1.0 & \textbf{44.11} & 2.0 & \textbf{42.76} \\
        \rowcolor{enrow}
        Llama 4 Scout (17B) & 0.01 & 7.74  & 0.15 & 9.76  & 5.0 & 0.00  \\
        \rowcolor{enrow}
        Qwen 3 VL (235B)    & 0.01 & 13.80 & 1.0 & 14.81 & 2.0 & 8.75 \\
        \rowcolor{enrow}
        Grok 4.1 Fast       & 0.01 & \underline{46.13} & 1.0 & \textbf{44.11} & 2.0 & 0.34 \\
        
        \bottomrule
    \end{tabular}
    \caption{English task accuracy in the Few-Shot CoT setup. The table presents results for different temperature ($T$) values, indicating the specific $T$ used alongside the corresponding accuracy. The best performance in each column is in \textbf{bold} and the second best is \underline{underlined}.}
    \label{tab:temp_ablation}
\end{table}





\subsection{Model Size Effect}
\label{app:model_size_effect}

\begin{figure}[t!]
    \centering
    \includegraphics[width=1\linewidth]{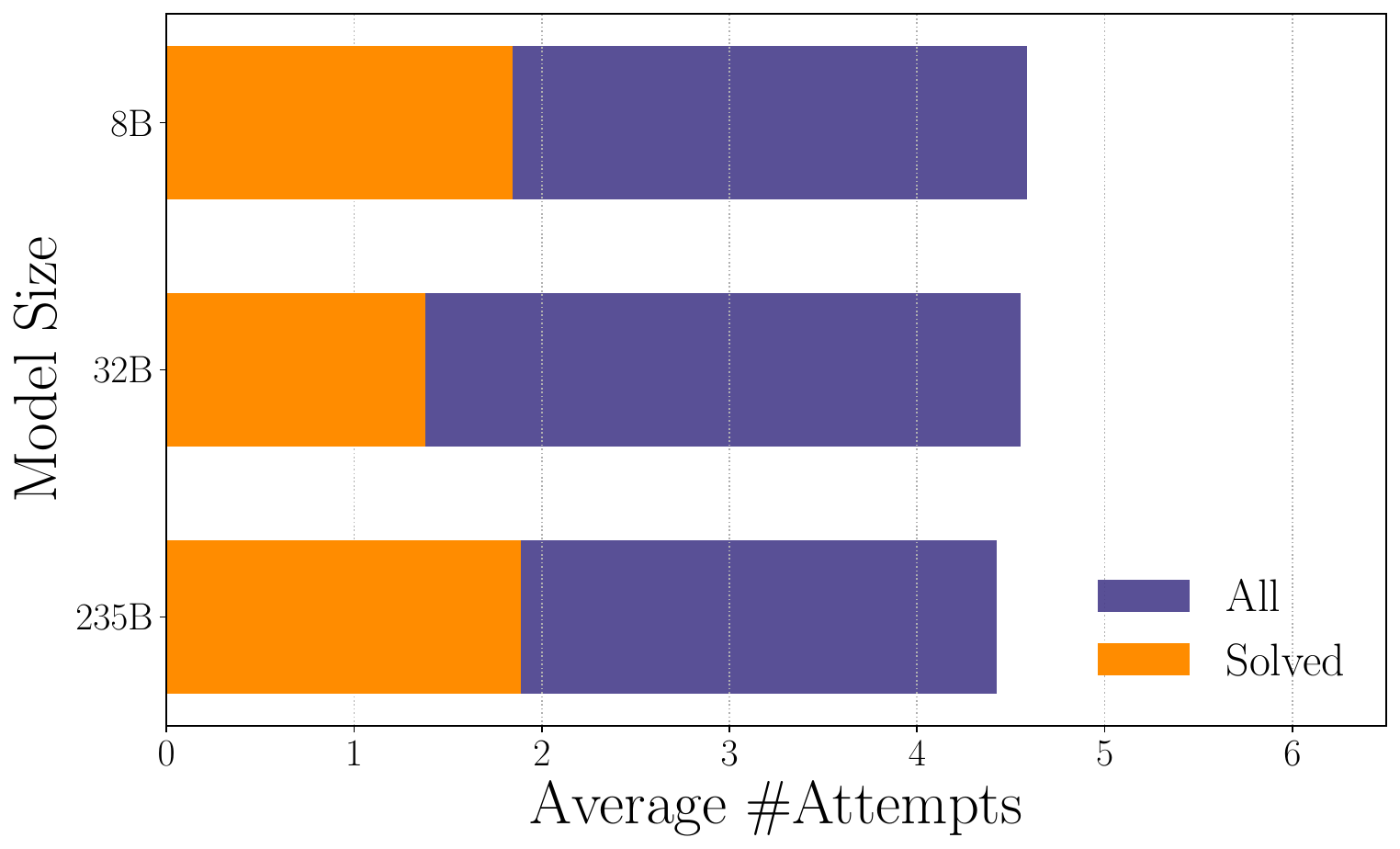}
    \caption{Average number of refinement attempts vs.\ model size for Qwen3-VL on the English subset. The plot reports the mean number of attempts under a five-attempt iterative refinement setting as a function of model size (Qwen3-VL-8B, 32B, 235B), shown for two aggregation regimes: averaging over solved puzzles only, and averaging over all puzzles, where unsolved instances are assigned the maximum of five attempts.}
    \label{fig:attempts_vs_size}
\end{figure}

\paragraph{Attempt Scaling with Model Size.}
Beyond final accuracy, we analyze how model size affects the process of iterative reasoning by measuring the number of refinement attempts used under a five-attempt iterative refinement setting on the English subset only. Figure~\ref{fig:attempts_vs_size} reports two complementary statistics as a function of model size: (i) the average number of attempts conditioned on solved puzzles only, and (ii) the average number of attempts over all puzzles, where unsolved instances are assigned the maximum of five attempts.

When conditioning on solved puzzles, the average number of attempts exhibits a non-monotonic trend, it first decreases from 8B to 32B models, then increases for the largest 235B model. The initial decrease suggests that moderately larger models can identify the correct interpretation more quickly, requiring fewer revision steps to reach a valid solution. The subsequent increase at the largest scale reflects a different regime where larger models appear able to sustain longer chains of productive reasoning, using additional refinement steps to recover from partial hypotheses and eventually converge on the correct answer. In contrast, when averaging over all puzzles, the mean number of attempts decreases monotonically with model size. This trend is driven by the sharp reduction in unsolved puzzles at larger scales, which otherwise dominate the mean by exhausting all five attempts.

\begin{figure*}[t!]
    \centering
    \includegraphics[width=1\linewidth]{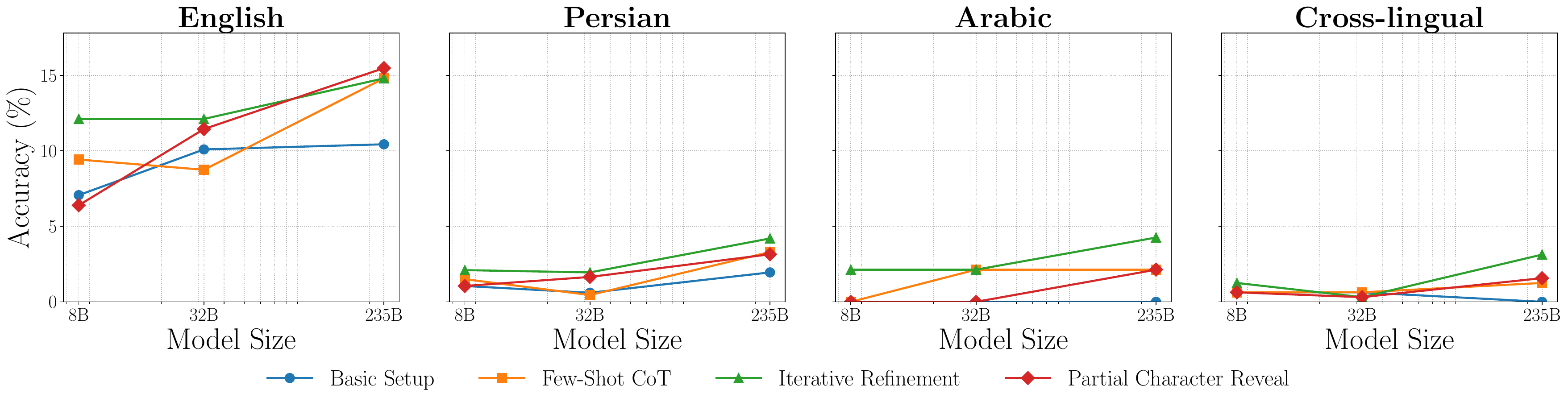}
    \caption{Accuracy vs.\ model size for Qwen3-VL across language subsets. Each panel reports accuracy on one subset as a function of model size (Qwen3-VL-8B, 32B, 235B) under four prompt variants: Basic, Few-shot CoT, Iterative Refinement, and Partial Character Reveal.}
    \label{fig:acc_vs_size_all}
\end{figure*}

\paragraph{Accuracy Scaling with Model Size.}
Figure~\ref{fig:acc_vs_size_all} extends the controlled scaling analysis of the Qwen3-VL family to all language subsets. We plot accuracy versus model size for Qwen3-VL-8B, 32B, and 235B under the same four prompt variants.

Across subsets, scaling generally improves performance, but the magnitude and stability of gains depend on both language and prompting. The English subset shows the most consistent and monotonic improvement with size across variants, with particularly clear gains under iterative refinement and partial character reveal, suggesting that larger models benefit more from settings that enable hypothesis revision or restrict the output space. Persian and cross-lingual subsets also trend upward with size, though improvements are smaller and more variant-dependent, indicating that cross-lingual bridging and non-Latin orthography remain challenging even for larger models. Arabic exhibits the weakest and least consistent scaling behavior. The accuracy remains low overall and improvements with size are modest, indicating that the dominant bottleneck is not simply capacity.

Overall, these results suggest that increasing parameters helps within a same-family comparison, yet performance remains far from saturated across all subsets, highlighting that scale alone does not resolve the core reasoning challenges posed by \benchmark{}.

\subsection{Language Coupling}

\begin{figure*}[t!]
    \centering
    \includegraphics[width=1\linewidth]{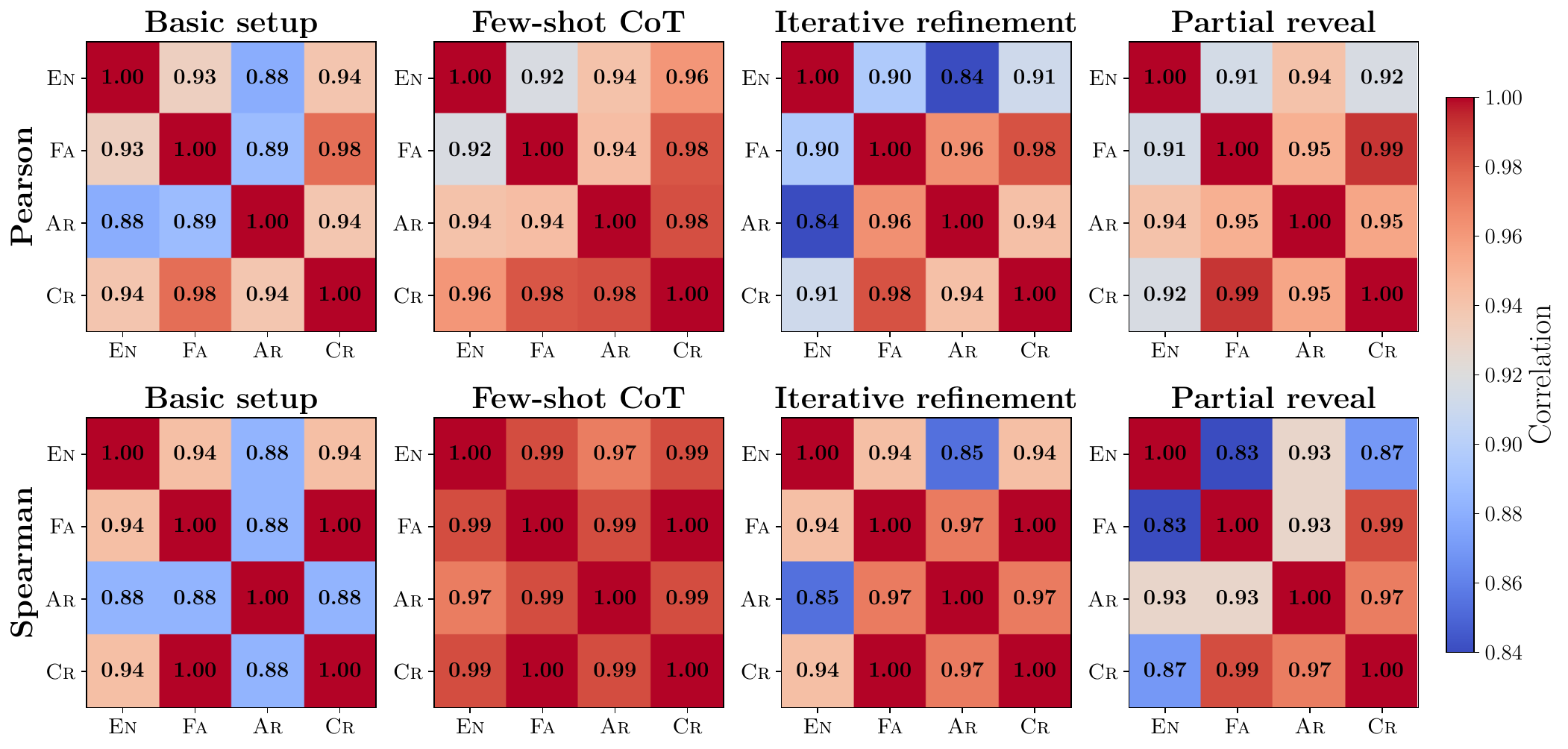}
    \caption{Cross-language coupling across prompt variants. Each matrix shows Pearson (top row) or Spearman (bottom row) correlations computed across models using subset-level accuracies for English (En), Persian (Fa), Arabic (Ar), and cross-lingual (Cr) puzzles, separately for the Basic setup, Few-shot CoT, Iterative Refinement, and Partial Character Reveal. Higher values indicate more consistent model performance across languages under a given evaluation variant.}
    \label{fig:all_correlation}
\end{figure*}

Figure~\ref{fig:all_correlation} extends the language-coupling analysis from the basic setup in Section~\ref{sec:correlation_basic} to all prompting variants. For each variant, we compute Pearson and Spearman correlations across models using subset-level accuracies, producing a $4\times4$ correlation matrix over \{En, Fa, Ar, Cr\}. Across all variants, correlations remain high, indicating that model strengths are largely consistent across languages even when the evaluation protocol changes. In particular, Persian and cross-lingual accuracies remain tightly coupled across variants, suggesting that the abilities driving performance on Persian puzzles also transfer to cross-lingual puzzles. Arabic shows comparatively weaker coupling in several variants, consistent with Arabic being the hardest subset and potentially introducing more variance across models.

\subsection{Failure-Case Similarity Density}
\label{app:similarity_density}

\begin{figure*}[t!]
    \centering
    
    \begin{subfigure}[b]{\textwidth}
        \centering
        \includegraphics[width=\linewidth]{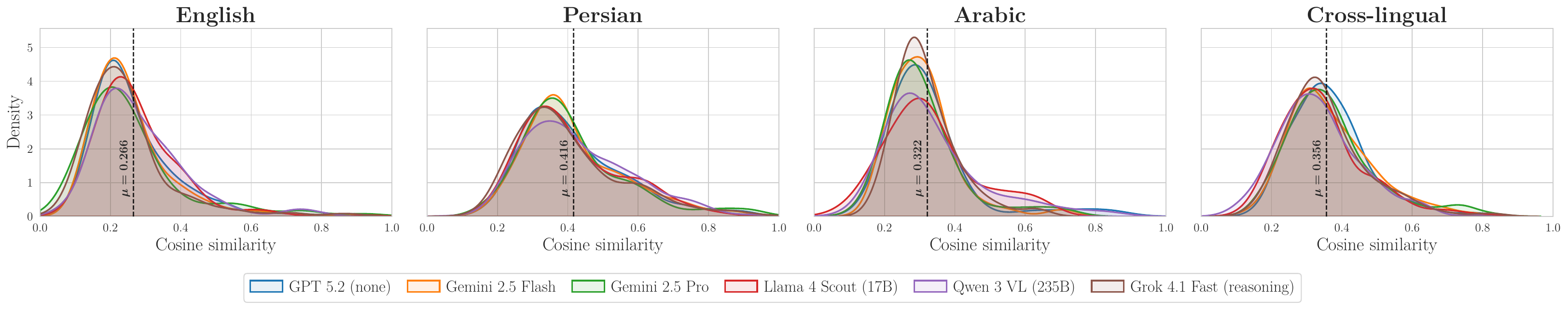}
        \caption{Basic Setup}
        \label{fig:sub1}
    \end{subfigure}
    \hfill 
    
    \begin{subfigure}[b]{\textwidth}
        \centering
        \includegraphics[width=\linewidth]{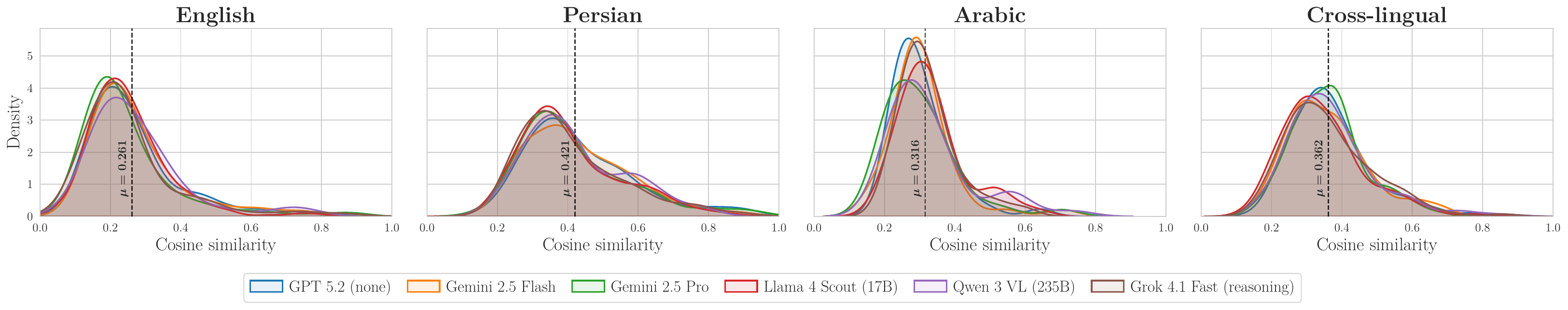}
        \caption{Few-Shot CoT}
        \label{fig:sub2}
    \end{subfigure}
    \hfill 
    
    \begin{subfigure}[b]{\textwidth}
        \centering
        \includegraphics[width=\linewidth]{figures/kde_glass_all_models_ctx0_pass1_char_count.pdf}
        \caption{Iterative Refinement}
        \label{fig:sub3}
    \end{subfigure}
    \hfill 
    
    \begin{subfigure}[b]{\textwidth}
        \centering
        \includegraphics[width=\linewidth]{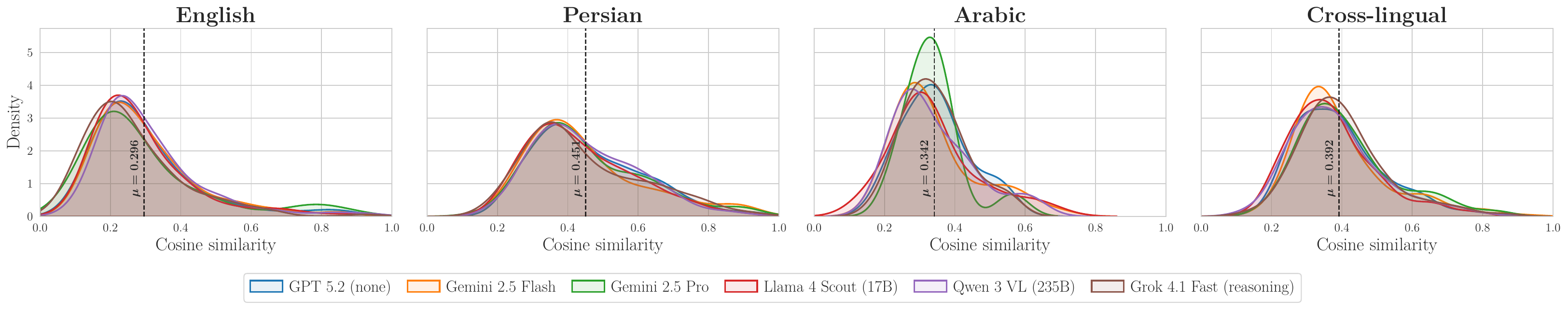}
        \caption{Partial Character Reveal}
        \label{fig:sub4}
    \end{subfigure}
    
    \caption{Semantic similarity of failure cases across evaluation variants. Each row corresponds to a prompting variant and each column to a language subset. Curves show kernel density estimates of cosine similarity between embedding representations of model outputs and ground-truth answers, computed over incorrect predictions only. Dashed lines indicate mean similarity for each plot.}

    \label{fig:density}
\end{figure*}

Figure~\ref{fig:density} extends the near-miss analysis from Figure~\ref{fig:density_refinement} to all evaluation variants (Basic, Few-shot CoT, Iterative Refinement, and Partial Character Reveal). As in the main paper, we compute cosine similarity between model outputs and ground-truth answers in an embedding space and restrict the analysis to failure cases.

Across all variants, the qualitative pattern remains consistent: similarity distributions place most mass at relatively low values, with only small high-similarity tails. This indicates that incorrect outputs are generally not close paraphrases of the ground truth, and that the main conclusions from iterative refinement hold broadly rather than being an artifact of a particular prompting protocol.

Notably, even partial character reveal, which substantially narrows the lexical search space, does not produce a large shift toward high semantic similarity among failures, suggesting that many errors reflect misinterpretations of the underlying visual cues rather than surface-form variation. This indicates that when models fail, they typically fail by selecting the wrong latent concept rather than producing a near-miss wording of the correct one.

\FloatBarrier

\section{More \benchmark{} examples}

\FloatBarrier

\label{extra_examples}

We provide additional qualitative examples of \benchmark{} in Table~\ref{tab:extra_examples_aggregated_EN_PE} and Table~\ref{tab:extra_examples_aggregated_AR_CL}.


\begin{table*}[h]
    \centering
    \small
    \setlength{\tabcolsep}{6pt}
    \renewcommand{\arraystretch}{1.25}

    \begin{tabular}{C{0.16\linewidth} J{0.28\linewidth} !{\vrule width 0.4pt} C{0.16\linewidth} J{0.28\linewidth}}
        \noalign{\hrule height 1pt}
        \rowcolor{black!70}
        \multicolumn{4}{c}{%
          \rule{0pt}{2.2ex}\textbf{\textcolor{white}{Additional \benchmark{} examples}}
        }\\
        \noalign{\hrule height 0.7pt}

        \rowcolor{black!70}
        \multicolumn{1}{c}{\rule{0pt}{2.2ex}\textbf{\textcolor{white}{Image}}} &
        \multicolumn{1}{c}{\rule{0pt}{2.2ex}\textbf{\textcolor{white}{Derivation}}} &
        \multicolumn{1}{c}{\rule{0pt}{2.2ex}\textbf{\textcolor{white}{Image}}} &
        \multicolumn{1}{c}{\rule{0pt}{2.2ex}\textbf{\textcolor{white}{Derivation}}} \\
        \noalign{\hrule height 0.7pt}

        \rowcolor{enrow}
        \multicolumn{4}{c}{\rule{0pt}{2ex}\textbf{English}}\\
        \noalign{\hrule height 0.4pt}

        \rowcolor{enrow}
        \includegraphics[width=\linewidth]{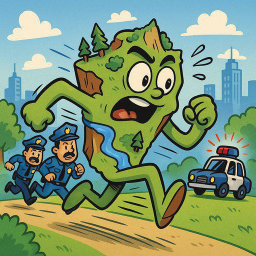} &
        The image shows a green landmass character representing ``land'' running from the cops through an outdoor scene or ``scape''. This suggests combining the two parts as ``land'' + ``scape''. Thus the final answer is \textbf{``landscape''}. &
        \includegraphics[width=\linewidth]{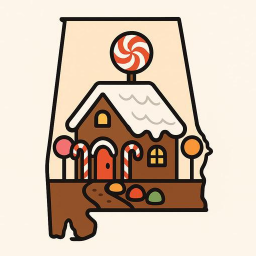} &
        The image shows a ``home'' (a house) decorated with ``sweets'' like candy canes and lollipops, all placed inside an outline shaped like the U.S.\ state of ``Alabama''. This can be read as ``sweet home'' in Alabama, pointing to the familiar phrase. Thus the final answer is \textbf{``Sweet Home Alabama''}. \\
        \noalign{\hrule height 0.7pt}

        \rowcolor{enrow}
        \includegraphics[width=\linewidth]{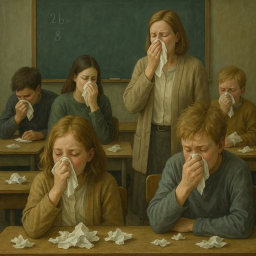} &
        The image shows a ``class'' (students in a classroom) where everyone is ``sick'' (holding tissues and blowing their noses). This can be described as ``class sick'', which sounds like ``classic''. Thus the final answer is \textbf{``classic''}. &
        \includegraphics[width=\linewidth]{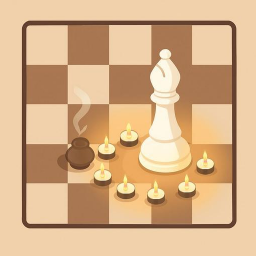} &
        The image shows a chessboard scene in which certain pieces are engaged in a ritual act. This can be read as ``rite chess'', which sounds like ``righteous''. Thus the final answer is \textbf{``righteous''}. \\
        \noalign{\hrule height 0.7pt}

\rowcolor{enrow}
\includegraphics[width=\linewidth]{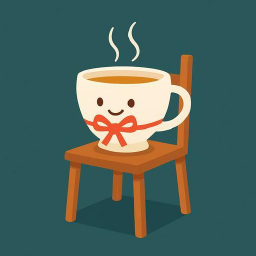} &
The image shows a cup of ``tea'' placed on a ``chair''. This suggests reading the two elements together as ``tea chair'', which sounds like ``teacher''. Thus the final answer is \textbf{``teacher'}. &
\includegraphics[width=\linewidth]{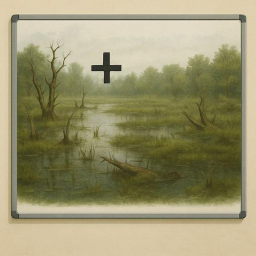} &
The image shows an ``add'' symbol indicating addition together with a muddy or swampy scene representing a ``mire''. Read together as ``add mire'', this sounds like ``admire''. Thus the final answer is \textbf{``admire''}. \\
\noalign{\hrule height 0.7pt}

\rowcolor{enrow}
\includegraphics[width=\linewidth]{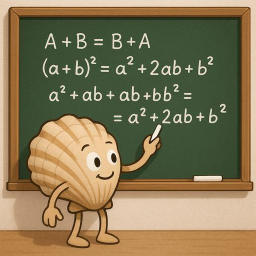} &
The image depicts a ``shell'' alongside a blackboard containing a mathematical ``proof''. Reading these two concepts together gives ``shell proof'', leading to the word ``shellproof''. Thus the final answer is \textbf{``shellproof''}. &
\includegraphics[width=\linewidth]{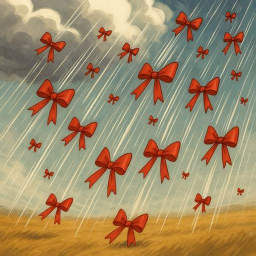} &
The image shows falling ``rain'' together with bow-shaped ribbons resembling a ``bow''. Interpreting these elements together gives ``rain bow'', which forms the word ``rainbow''. Thus the final answer is \textbf{``rainbow''}. \\
\noalign{\hrule height 0.7pt}

        \rowcolor{farow}
        \multicolumn{4}{c}{\rule{0pt}{2ex}\textbf{Persian}}\\
        \noalign{\hrule height 0.4pt}

        \rowcolor{farow}
        \includegraphics[width=\linewidth]{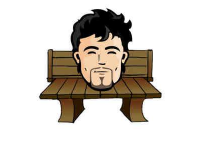} &
        The image shows a head, which in Persian is written \foreignlanguage{arabic}{``سر''}, placed on a bench/seat, suggesting the idea of sitting (\foreignlanguage{arabic}{``نشین''} which relates to the verb "sit" in Persian). Combining \foreignlanguage{arabic}{``سر''} with \foreignlanguage{arabic}{``نشین''} yields \foreignlanguage{arabic}{``سرنشین''}; thus, the final answer is \textbf{\foreignlanguage{arabic}{``سرنشین''}}. &
        \includegraphics[width=\linewidth]{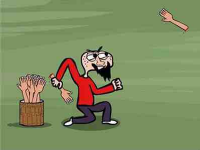} &
        The image shows a hand, which in Persian is written \foreignlanguage{arabic}{``دست''}, together with an action of throwing/putting, which cues \foreignlanguage{arabic}{``انداز''}. Combining \foreignlanguage{arabic}{``دست''} with \foreignlanguage{arabic}{``انداز''} yields \foreignlanguage{arabic}{``دست‌انداز''}; thus, the final answer is \textbf{\foreignlanguage{arabic}{``دست‌انداز''}}. \\
        \noalign{\hrule height 0.7pt}

        \rowcolor{farow}
        \includegraphics[width=\linewidth]{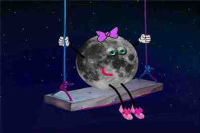} &
        The image shows the moon, which in Persian is written \foreignlanguage{arabic}{``ماه''}, sitting on a swing. In Persian, swinging/rocking is associated with \foreignlanguage{arabic}{``تاب''}. Combining \foreignlanguage{arabic}{``ماه''} with \foreignlanguage{arabic}{``تاب''} yields \foreignlanguage{arabic}{``مهتاب''}; thus, the final answer is \textbf{\foreignlanguage{arabic}{``مهتاب''}}. &
        \includegraphics[width=\linewidth]{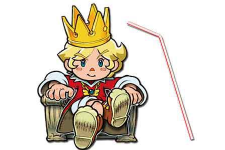} &
        The image shows a straw, which in Persian is written \foreignlanguage{arabic}{``نی''}, together with a king, which cues \foreignlanguage{arabic}{``شاه''}. The king’s blond hair cues \foreignlanguage{arabic}{``بور''}. Combining \foreignlanguage{arabic}{``نی''} with \foreignlanguage{arabic}{``شاه''} and \foreignlanguage{arabic}{``بور''} yields \foreignlanguage{arabic}{``نیشابور''}; thus, the final answer is \textbf{\foreignlanguage{arabic}{``نیشابور''}}. \\
        \noalign{\hrule height 0.7pt}

        \rowcolor{farow}
\includegraphics[width=\linewidth]{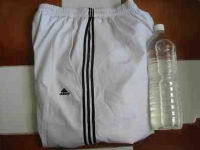} &
The image shows water, which in Persian is written \foreignlanguage{arabic}{``آب''}, placed next to a warm-up pant, which cues \foreignlanguage{arabic}{``گرم‌کن''}. Reading these elements together gives \foreignlanguage{arabic}{``آب‌گرم‌کن''}; thus, the final answer is \textbf{\foreignlanguage{arabic}{``آب‌گرم‌کن''}}. &
\includegraphics[width=\linewidth]{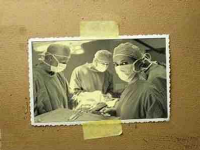} &
The image shows a photograph of a surgical operation, which can be read as \foreignlanguage{arabic}{``عکسِ عمل''}. Interpreting this as ``عکس‌العمل'' leads to the intended word; thus, the final answer is \textbf{\foreignlanguage{arabic}{``عکس‌العمل''}}. \\
\noalign{\hrule height 0.7pt}

\rowcolor{farow}
\includegraphics[width=\linewidth]{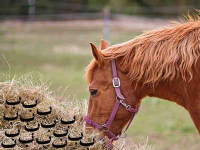} &
The image shows a horse, which in Persian is written \foreignlanguage{arabic}{``اسب''}. The horse is eating the Persian letter \foreignlanguage{arabic}{``ب''}, and since \foreignlanguage{arabic}{``ب''} with \foreignlanguage{arabic}{``خوار''} reads as \foreignlanguage{arabic}{``ب‌خوار''}, this can be interpreted as \foreignlanguage{arabic}{``بخار''}. Reading these elements together yields \foreignlanguage{arabic}{``اسب بخار''}; thus, the final answer is \textbf{\foreignlanguage{arabic}{``اسب بخار''}}. &
\includegraphics[width=\linewidth]{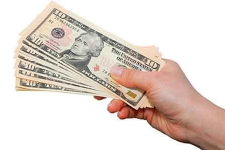} &
The image shows a hand, which in Persian is written \foreignlanguage{arabic}{``دست''}, holding money or property, which cues \foreignlanguage{arabic}{``مال''}. Interpreting these elements together gives \foreignlanguage{arabic}{``دستمال''}; thus, the final answer is \textbf{\foreignlanguage{arabic}{``دستمال''}}. \\
\noalign{\hrule height 0.7pt}

    \end{tabular}

    \caption{Additional illustrative \benchmark{} examples across English and Persian settings. Each puzzle consists of an input image and a short derivation leading to the target word or phrase.}
    \label{tab:extra_examples_aggregated_EN_PE}
\end{table*}

\begin{table*}[t!]
    \centering
    \small
    \setlength{\tabcolsep}{6pt}
    \renewcommand{\arraystretch}{1.25}

    \begin{tabular}{C{0.16\linewidth} J{0.28\linewidth} !{\vrule width 0.4pt} C{0.16\linewidth} J{0.28\linewidth}}
        \noalign{\hrule height 1pt}
        \rowcolor{black!70}
        \multicolumn{4}{c}{%
          \rule{0pt}{2.2ex}\textbf{\textcolor{white}{Additional \benchmark{} examples}}
        }\\
        \noalign{\hrule height 0.7pt}

        \rowcolor{black!70}
        \multicolumn{1}{c}{\rule{0pt}{2.2ex}\textbf{\textcolor{white}{Image}}} &
        \multicolumn{1}{c}{\rule{0pt}{2.2ex}\textbf{\textcolor{white}{Derivation}}} &
        \multicolumn{1}{c}{\rule{0pt}{2.2ex}\textbf{\textcolor{white}{Image}}} &
        \multicolumn{1}{c}{\rule{0pt}{2.2ex}\textbf{\textcolor{white}{Derivation}}} \\
        \noalign{\hrule height 0.7pt}

        \rowcolor{arow}
        \multicolumn{4}{c}{\rule{0pt}{2ex}\textbf{Arabic}}\\
        \noalign{\hrule height 0.4pt}

        \rowcolor{arow}
        \includegraphics[width=\linewidth]{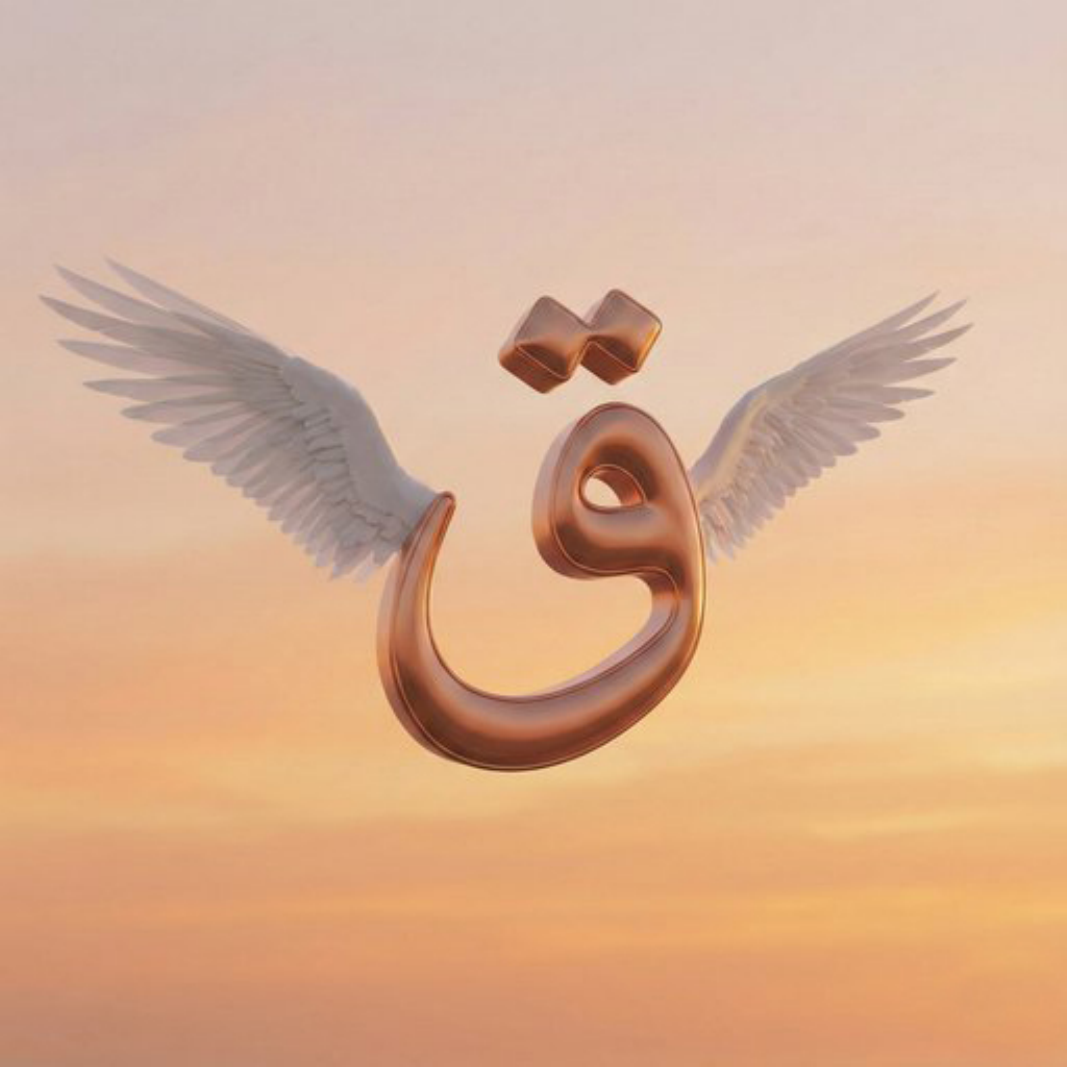} &
        The image shows the Arabic letter \foreignlanguage{arabic}{``ق''} with wings, indicating that it is flying, which cues the verb \foreignlanguage{arabic}{``طَارَ''}. Combining \foreignlanguage{arabic}{``ق''} with \foreignlanguage{arabic}{``طَارَ''} yields \foreignlanguage{arabic}{``قِطَار''}. Thus the final answer is \textbf{\foreignlanguage{arabic}{``قِطَار''}}. &
        \includegraphics[width=\linewidth]{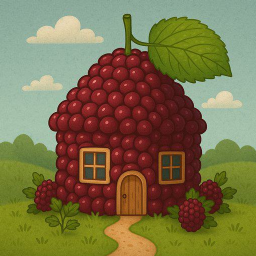} &
        The image shows a house, which in Arabic is \foreignlanguage{arabic}{``بَيْت''}, constructed out of mulberries, which cue the word \foreignlanguage{arabic}{``تُوت''}. Concatenating \foreignlanguage{arabic}{``بَيْت''} and \foreignlanguage{arabic}{``تُوت''} yields \foreignlanguage{arabic}{``بَيْتُوتَة''}. Thus the final answer is \textbf{\foreignlanguage{arabic}{``بَيْتُوتَة''}}. \\
        \noalign{\hrule height 0.7pt}

        \rowcolor{arow}
        \includegraphics[width=\linewidth]{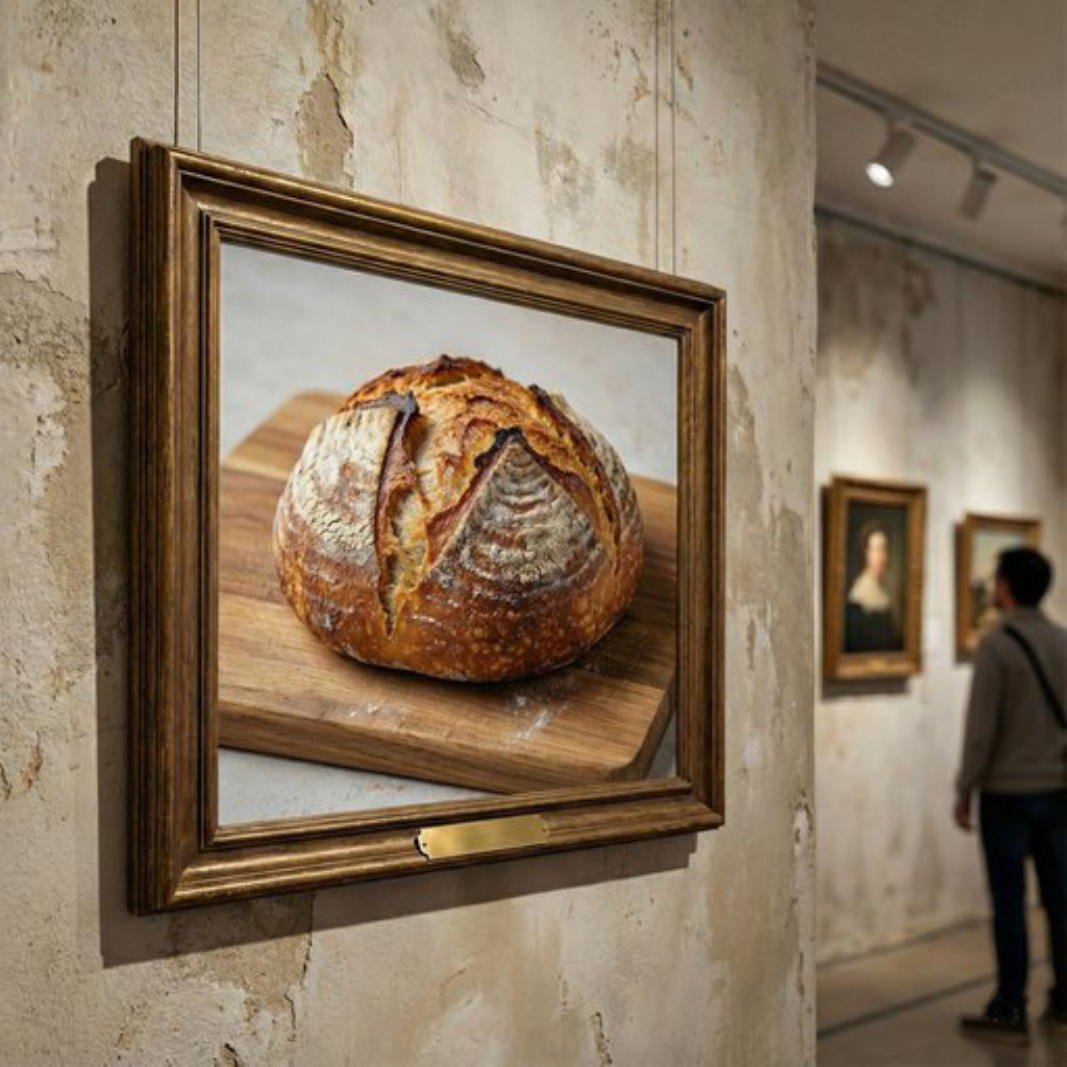} &
        The image shows a museum artwork, which cues the Arabic word \foreignlanguage{arabic}{``فَنّ''}, depicting a loaf of bread, which cues \foreignlanguage{arabic}{``نَان''}. Combining \foreignlanguage{arabic}{``فَنّ''} with \foreignlanguage{arabic}{``نَان''} yields \foreignlanguage{arabic}{``فَنَّان''}. Thus the final answer is \textbf{\foreignlanguage{arabic}{``فَنَّان''}}. &
        \includegraphics[width=\linewidth]{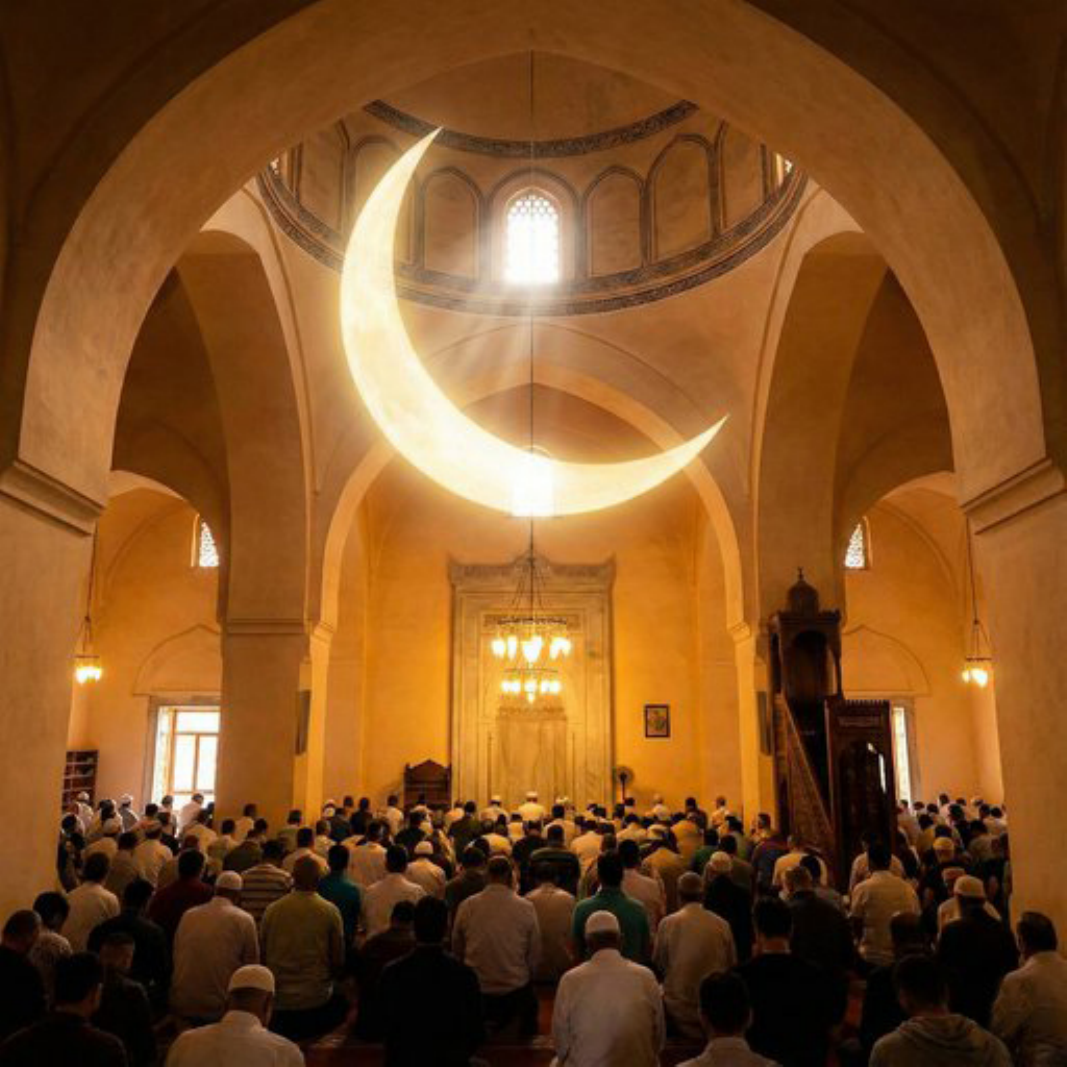} &
       The image shows a bright crescent moon, which cues the Arabic word \foreignlanguage{arabic}{``قَمَر''}, above a mosque interior with worshippers, suggesting \foreignlanguage{arabic}{``دِين''} which means religion. Combining \foreignlanguage{arabic}{``قَمَر''} and \foreignlanguage{arabic}{``دِين''} yields \foreignlanguage{arabic}{``قَمَرُالدِّين''}. Thus, the final answer is \textbf{\foreignlanguage{arabic}{``قَمَرُالدِّين''}}. \\
        \noalign{\hrule height 0.7pt}

        \rowcolor{arow}
\includegraphics[width=\linewidth]{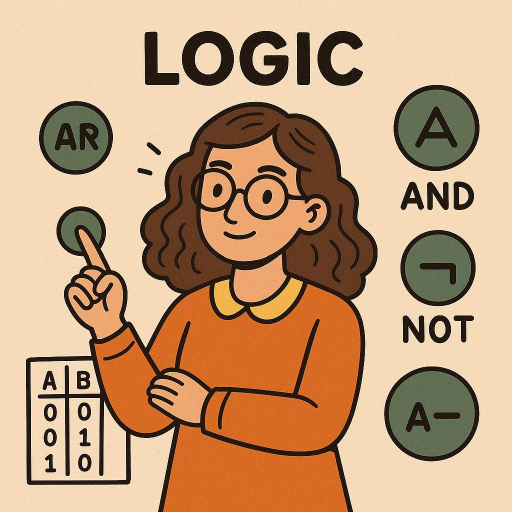} &
The image shows a scene related to logic, which cues the Arabic word \foreignlanguage{arabic}{``مِنْطِق''}. The presence of a feminine marker suggests adding \foreignlanguage{arabic}{``ـة''}. Reading these together gives \foreignlanguage{arabic}{``مِنْطَقَة''}. Thus the final answer is \textbf{\foreignlanguage{arabic}{``مِنْطَقَة''}}. &
\includegraphics[width=\linewidth]{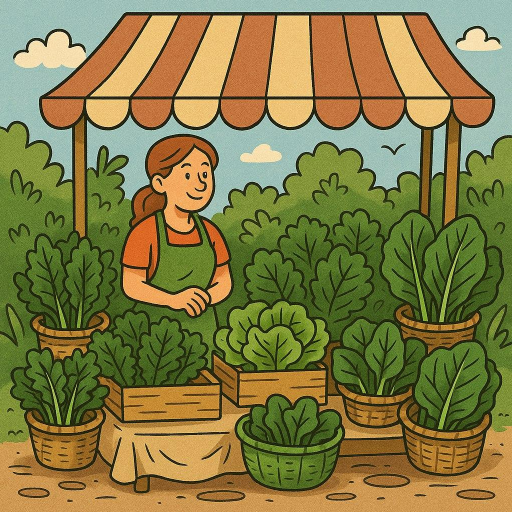} &
The image shows a market, which in Arabic is \foreignlanguage{arabic}{``سُوق''}, filled with fresh plants, which cue \foreignlanguage{arabic}{``طَرِيّ''}. Interpreting these attachively gives \foreignlanguage{arabic}{``سُوقَطْرَى''}. Thus the final answer is \textbf{\foreignlanguage{arabic}{``سُوقَطْرَى''}}. \\
\noalign{\hrule height 0.7pt}

\rowcolor{arow}
\includegraphics[width=\linewidth]{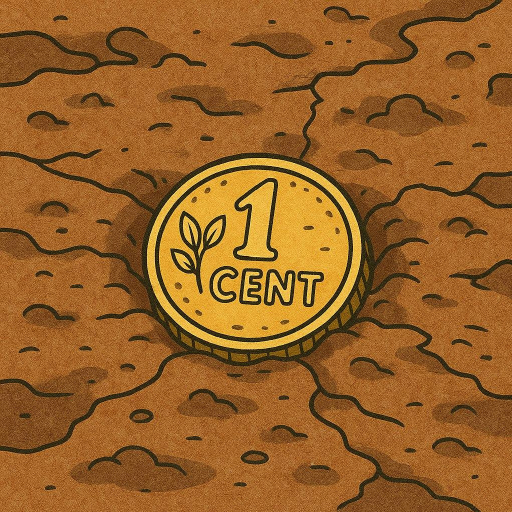} &
The image shows a coin, which cues the Arabic word \foreignlanguage{arabic}{``فِلْس''}, placed in soil or mud, which cues \foreignlanguage{arabic}{``طِين''}. Reading these elements together gives \foreignlanguage{arabic}{``فِلَسْطِين''}. Thus the final answer is \textbf{\foreignlanguage{arabic}{``فِلَسْطِين''}}. &
\includegraphics[width=\linewidth]{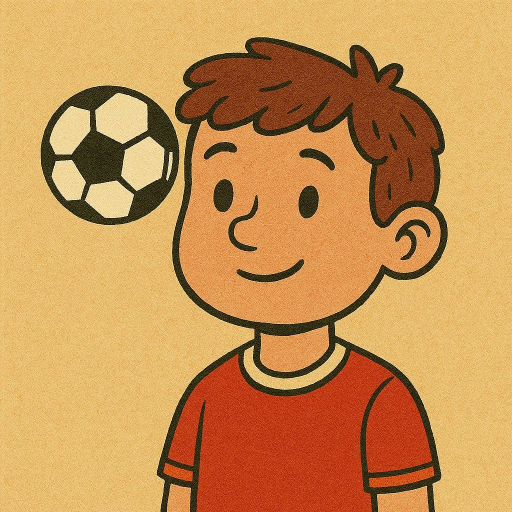} &
The image shows a ball, which cues \foreignlanguage{arabic}{``كُرَة''}, together with a head belonging to a person, which can be read as \foreignlanguage{arabic}{``رَاسُه''} (his head). Interpreting these elements together gives \foreignlanguage{arabic}{``كُرَّاسَة''}. Thus the final answer is \textbf{\foreignlanguage{arabic}{``كُرَّاسَة''}}. \\
\noalign{\hrule height 0.7pt}

        \rowcolor{clrow}
        \multicolumn{4}{c}{\rule{0pt}{2ex}\textbf{Cross-lingual}}\\
        \noalign{\hrule height 0.4pt}

        \rowcolor{clrow}
        \includegraphics[width=\linewidth]{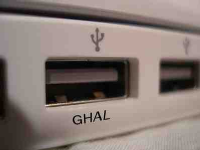} &
        The image shows a USB port, which directly cues the Persian loanword \foreignlanguage{arabic}{``پورت''}. Beneath the port, the label reads GHAL, which cues \foreignlanguage{arabic}{``غال''}. Combining \foreignlanguage{arabic}{``پورت''} with \foreignlanguage{arabic}{``غال''} yields \foreignlanguage{arabic}{``پرتغال''}. Thus, the final answer is \textbf{\foreignlanguage{arabic}{``پرتغال''}}. &
        \includegraphics[width=\linewidth]{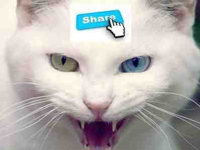} &
        The image shows a Share button, which cues the sound \foreignlanguage{arabic}{``شر''} (from ``share''), together with a cat, which cues \foreignlanguage{arabic}{``کت''}. Combining \foreignlanguage{arabic}{``شر''} with \foreignlanguage{arabic}{``کت''} yields \foreignlanguage{arabic}{``شرکت''}. Thus, the final answer is \textbf{\foreignlanguage{arabic}{``شرکت''}}. \\
        \noalign{\hrule height 0.7pt}

        \rowcolor{clrow}
        \includegraphics[width=\linewidth]{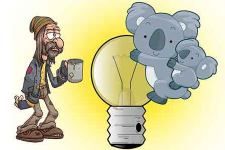} &
        The image shows a koala, which cues \foreignlanguage{arabic}{``کوالا''}, a lamp, which cues \foreignlanguage{arabic}{``لامپ''}, and a poor beggar, which cues \foreignlanguage{arabic}{``پور''} as a sound cue. Combining \foreignlanguage{arabic}{``کوالا''} with \foreignlanguage{arabic}{``لامپ''} and \foreignlanguage{arabic}{``پور''} yields \foreignlanguage{arabic}{``کوالالامپور''}. Thus, the final answer is \textbf{\foreignlanguage{arabic}{``کوالالامپور''}}. &
        \includegraphics[width=\linewidth]{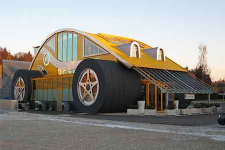} &
        The image shows a car, which cues \foreignlanguage{arabic}{``کار''} as a sound cue, and a house, which cues \foreignlanguage{arabic}{``خانه''}. Combining \foreignlanguage{arabic}{``کار''} with \foreignlanguage{arabic}{``خانه''} yields \foreignlanguage{arabic}{``کارخانه''}. Thus, the final answer is \textbf{\foreignlanguage{arabic}{``کارخانه''}}. \\
        \noalign{\hrule height 1pt}

        \rowcolor{clrow}
\includegraphics[width=\linewidth]{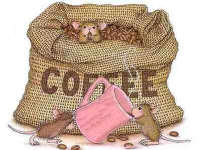} &
The image shows a mouse (\foreignlanguage{arabic}{``موش''}) next to a bag of coffee, cueing the English word ``coffee''. Reading these together yields \foreignlanguage{arabic}{``موش''} + ``coffee'', forming \foreignlanguage{arabic}{``موشکافی''}. Thus, the final answer is \textbf{\foreignlanguage{arabic}{``موشکافی''}}. &
\includegraphics[width=\linewidth]{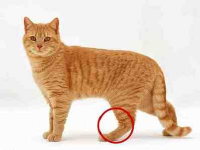} &
The image shows \foreignlanguage{arabic}{``پای گربه''} (a cat’s leg). By replacing \foreignlanguage{arabic}{``گربه''} with the English sound cue ``cat'' (read as \foreignlanguage{arabic}{``کت''}), we obtain \foreignlanguage{arabic}{``پای کت''}, which contracts to \foreignlanguage{arabic}{``پاکت''}. Thus, the final answer is \textbf{\foreignlanguage{arabic}{``پاکت''}}. \\
\noalign{\hrule height 0.7pt}

        \rowcolor{clrow}
\includegraphics[width=\linewidth]{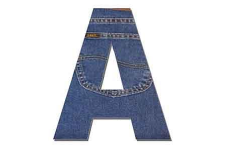} &
The image shows the English letter ``A'' placed on denim material, which cues the Persian word \foreignlanguage{arabic}{``لی''}. Reading these elements together gives ``A'' + \foreignlanguage{arabic}{``لی''}, forming \foreignlanguage{arabic}{``عالی''}. Thus, the final answer is \textbf{\foreignlanguage{arabic}{``عالی''}}. &
\includegraphics[width=\linewidth]{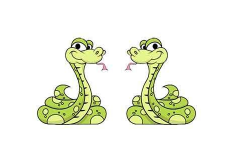} &
The image shows two snakes, and in Persian a snake is \foreignlanguage{arabic}{``مار''}. Reading this as ``two'' + \foreignlanguage{arabic}{``مار''} yields \foreignlanguage{arabic}{``طومار''}. Thus, the final answer is \textbf{\foreignlanguage{arabic}{``طومار''}}. \\
\noalign{\hrule height 1pt}
    \end{tabular}

    \caption{Additional illustrative \benchmark{} examples across Arabic and cross-lingual settings. Each puzzle consists of an input image and a short derivation leading to the target word or phrase.}
    \label{tab:extra_examples_aggregated_AR_CL}
\end{table*}

\FloatBarrier

\section{Prompt Options}
\label{app:prompt_options}

This appendix describes the prompt building blocks used in our experiments. Each query includes one image and a text prompt formed by concatenating a fixed set of reusable modules, as outlined below.

\subsection{Language and subset rules}
\label{app:prompt_language_rules}
We use four rule modules, one for each subset: English, Persian, Arabic, and cross-lingual. Each module specifies the required answer language and any subset-specific constraints. For instance, in the cross-lingual subset, the target answer language is Persian (Farsi), but solving the puzzle may require English knowledge because the image can contain English words, letters, numbers, or concepts. The intended answer may therefore include English elements directly (e.g., via transliteration) or combine English and Persian components.

\subsection{Shared task and output instructions}
\label{app:prompt_game_rules}
All experimental settings include the same core instruction module. This block explains the game setup, states what the model should infer from the image and the accompanying text, and clarifies the expected form of the response. Crucially, it also enforces a strict output constraint to ensure consistent evaluation across models and prompt variants.

\subsection{Hint modules}
\label{app:prompt_hints}
To provide additional guidance without changing the underlying task, we use three lightweight hint modules that can be optionally appended to the base prompt. These hints are designed to constrain the search space or support multi-round solving:
\begin{itemize}[leftmargin=1.3em]
    \item \textbf{Answer-length hint:} provides the answer length in characters, excluding spaces.
    \item \textbf{Partial-reveal hint:} provides a masked pattern where a subset of characters is revealed and all remaining positions are replaced by underscores.
    \item \textbf{Iterative-refinement hint:} provides feedback that includes the model’s previous answer, enabling a revision step in subsequent rounds.
\end{itemize}

\subsection{Few-shot CoT demonstrations}
\label{app:prompt_fewshot}
For the few-shot CoT condition, we prepend three solved demonstrations from the same subset as the target puzzle. We prepare one demonstration block per subset (English, Persian, Arabic, and cross-lingual). Each demonstration includes an image and a concise, structured rationale with two fields: primary clues and candidate answers. The full set of demonstrations used in our experiments is provided below.

\newtcolorbox{promptbox}[2][]{
  colback=gray!3,
  colframe=gray!50,
  coltitle=white,
  fonttitle=\bfseries,
  title={#2},
  boxsep=4pt,
  left=4pt,
  right=4pt,
  top=4pt,
  bottom=4pt,
  colbacktitle=black,
  #1
}

\begin{PromptEN}[English Language Rules]{}
\textbf{LANGUAGE RULES:}
\begin{itemize}
    \item The target answer language is English.
\end{itemize}

\end{PromptEN}

\begin{PromptFA}[Persian Language Rules]{}
\textbf{LANGUAGE RULES:}
\begin{itemize}

    \item The target answer language is Persian (Farsi).

    \item \textbf{CULTURAL LENS:} Do not simply translate English concepts. You must interpret the visual elements through the lens of Persian culture, literature, and common daily idioms.

    \item \textbf{WORDPLAY:} If the image suggests wordplay, prioritize phonetic/semantic connections natural in Persian.
    
\end{itemize}
\end{PromptFA}

\begin{PromptAR}[Arabic Language Rules]{}
\textbf{LANGUAGE RULES:}
\begin{itemize}

    \item The target answer language is Arabic.

    \item CULTURAL LENS: Do not simply translate English concepts. You must interpret the visual elements through the lens of Arabic culture, literature, and common daily idioms.

    \item WORDPLAY: If the image suggests wordplay, prioritize phonetic/semantic connections natural in Arabic.
    
\end{itemize}

\end{PromptAR}

\begin{PromptCL}[Cross-Lingual Rules]{}
\textbf{LANGUAGE RULES:}

\begin{itemize}

    \item The target answer language is Persian (Farsi).

    \item \textbf{ENGLISH KNOWLEDGE REQUIRED:} The puzzle may rely on English words, concepts, letters, or numbers depicted in the image.

    \item You may need to use English elements directly in the Persian answer (transliteration) or combine them with Persian to form the intended phrase.
\end{itemize}
\end{PromptCL}

\begin{PromptGen}[Game Explanation]{}

    You are an expert multi-modal puzzle solver. You solve picture word puzzles.

    \textbf{GAME DESCRIPTION:}
    
    - You will see exactly ONE image per puzzle.\\
    - The image may depict objects, people, scenes, text, icons, or abstract compositions.\\
    - The goal is to infer a SINGLE intended answer: one word or a short phrase.\\
    - The image is a deliberately constructed clue for a linguistic target, NOT a request to describe the scene.\\
    - The intended answer may be:\\
    - a literal word,\\
    - an idiom or proverb,\\
    - a pun or wordplay,\\
    - a common expression,\\
    - a culturally meaningful phrase,\\
    - or a proper noun / named entity (person, place, title, brand, named item).
    
\textbf{GENERAL SOLVING PROCEDURE (follow in order):}
\begin{enumerate}
    \item Identify candidate clue units in the image:
    
    - the most salient objects/entities \\
    - any text, letters, numbers, symbols, or icons \\
    - any repeated motif/pattern
    \item Select ONLY 2–4 PRIMARY clue units:
    
    - prefer central/emphasized/repeated units\\
    - compress repeated motifs into one unit\\
    - ignore minor background details unless they clearly change a primary unit
    \item Hypothesize a simple composition:
    
    - the answer is usually formed by combining or transforming the primary units\\
    - prefer the simplest coherent interpretation with the fewest assumptions
    \item Choose the best final answer:
    
    - it should be natural/common in the target language\\
    - it should explain the primary units as a single intended construction\\
    - prioritize global coherence over matching every local detail
\end{enumerate}
    \textbf{OUTPUT REQUIREMENT:}\\
    - Provide exactly ONE final answer (single word or short phrase).\\
    - If uncertain, choose the most plausible candidate under the simplest coherent interpretation.\\

\end{PromptGen}

\begin{PromptGen}[Output Instructions]{}

    \textbf{OUTPUT FORMAT:}
        Return ONLY a single valid JSON object. Do not output markdown blocks or conversational text.
    
        \{
        \begin{enumerate}[label={}, itemsep=0pt, topsep=0pt]
            \item ``primary\_clues": [``...", ``..."],
            \item ``candidates": [``...", ``...", ``..."],
            \item ``final\_answer": ``..."
        \end{enumerate}
        \}

\end{PromptGen}

\begin{PromptCL}[\#Characters Hint]{}
The answer has <\#Characters> characters (excluding spaces).
\end{PromptCL}

\begin{PromptCL}[Partial Character Reveal Hint]{}
The pattern of the answer is ``\_\_<character>\_\_<character>\_".\\ 
In this pattern, “\_” represents a character and spaces represent actual spaces in the answer.
\end{PromptCL}

\begin{PromptCL}[Iterative Refinement]{}
Your previous attempt was <model\_previous\_answer> which is incorrect. Analyze the image carefully and try again.
\end{PromptCL}

\begin{PromptFA}[Persian Language Few-shot CoT]{}

\begin{center}
\begin{minipage}[t]{0.32\linewidth}
  \centering
  \includegraphics[width=0.8\linewidth]{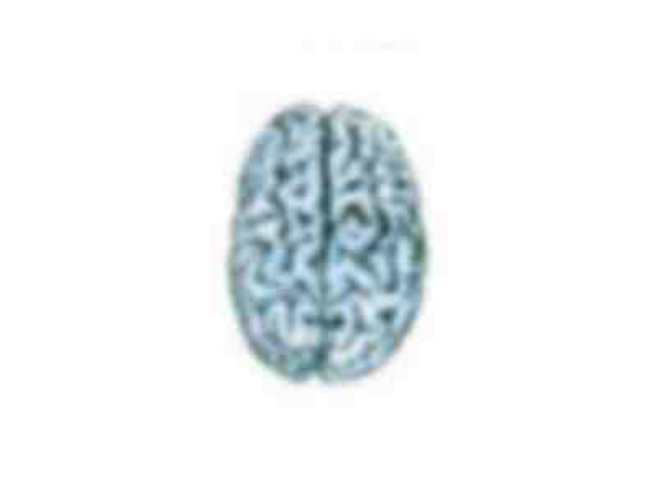}
\end{minipage}\hfill
\begin{minipage}[t]{0.32\linewidth}
  \centering
  \includegraphics[width=0.8\linewidth]{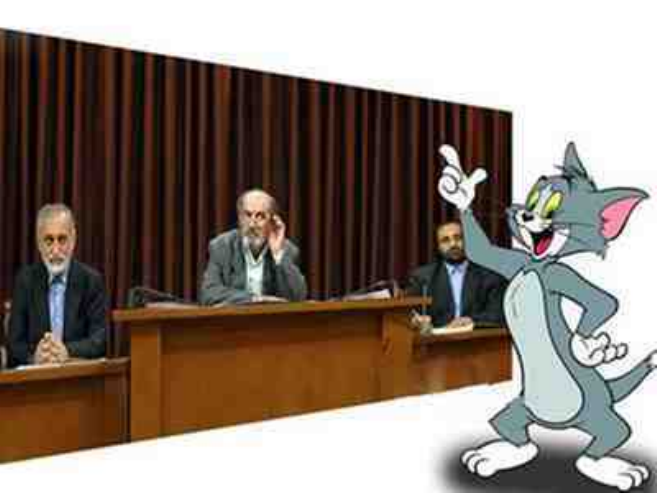}
\end{minipage}\hfill
\begin{minipage}[t]{0.32\linewidth}
  \centering
  \includegraphics[width=0.8\linewidth]{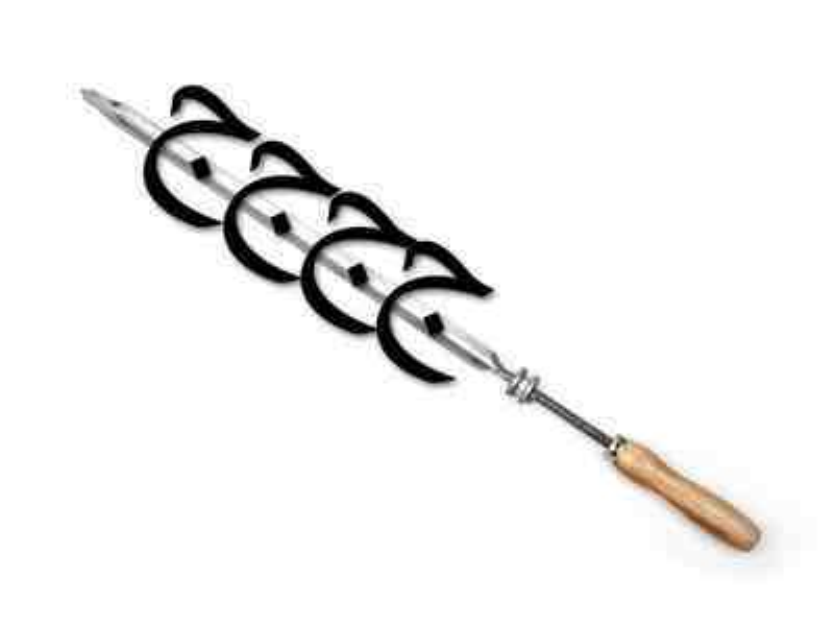}
\end{minipage}
\end{center}

\textbf{Example 1:}

"primary\_clues": [

            "The image displays a human brain, which translates to 'Mokh' \foreignlanguage{Persian}{(مخ)} in Persian.",
            
            "The visual style of the brain is blurred or faded, which translates to the adjective 'Tar' \foreignlanguage{Persian}{(تار)}.",
            
            "Combining the noun 'Mokh' with the adjective 'Tar' creates the construction 'Mokh-Tar'.",
            
            "This forms the common Persian name 'Mokhtar' \foreignlanguage{Persian}{(مختار)}."
            ]
        
        "candidates": [
            \foreignlanguage{Persian}{"مخ مات"}, 
            \foreignlanguage{Persian}{"تارمخ"},  
            \foreignlanguage{Persian}{"مختار"} 
        ]\\

\textbf{Example 2:}

"primary\_clues": [
            "The image shows a courtroom setting with lawyers, representing the legal concept of 'Vekalat' \foreignlanguage{Persian}{(وکالت)} or power of attorney.",
            
            "The cartoon character is Tom from 'Tom and Jerry', whose name in Persian is 'Tam' \foreignlanguage{Persian}{(تام)}.",
            
            "Combining the legal term 'Vekalat' with the character's name 'Tam' creates the phrase 'Vekalat-e Tam' \foreignlanguage{Persian}{(وکالت تام)}.",
            
            "This phrase is a common legal term meaning 'Full power of attorney'."
        ]
        
        "candidates": [
            \foreignlanguage{Persian}{"تام وکیله"},
            \foreignlanguage{Persian}{"وکیل تامی"},
            \foreignlanguage{Persian}{"وکالت تام"}     
        ]\\

\textbf{Example 3:}

"primary\_clues": [
            "The image displays a grilling skewer (seekh) typically used for making kebabs.",
            
            "Instead of meat, the skewer holds multiple instances of the Persian letter 'Che' \foreignlanguage{Persian}{(چ)}.",
            
            "In Persian, 'multiple Che's' or 'several Che's' translates to 'Chand Che' \foreignlanguage{Persian}{(چند چ)}.",
            
            "The phrase 'Chand Che' sounds phonetically very similar to 'Chenjeh' \foreignlanguage{Persian}{(چنجه)}.",
            
            "This creates a pun on 'Kebab Chenjeh' (Lamb Chop Kebab), replacing the meat with letters."
        ]
        
        "candidates": [
            \foreignlanguage{Persian}{"چهار کباب"}, 
            \foreignlanguage{Persian}{"کباب چندچ"}, 
            \foreignlanguage{Persian}{"کباب چنجه"}  
        ]
\end{PromptFA}

\begin{PromptEN}[English Language Few-shot CoT]{}
\textbf{Example 1:}

\begin{center}
\begin{minipage}[t]{0.32\linewidth}
  \centering
  \includegraphics[width=0.8\linewidth]{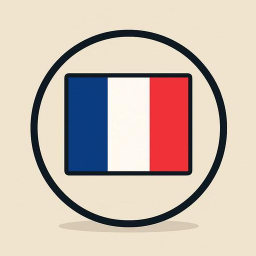}
\end{minipage}\hfill
\begin{minipage}[t]{0.32\linewidth}
  \centering
  \includegraphics[width=0.8\linewidth]{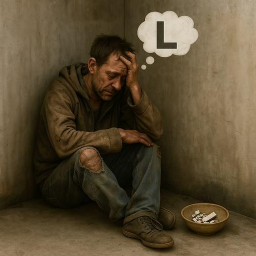}
\end{minipage}\hfill
\begin{minipage}[t]{0.32\linewidth}
  \centering
  \includegraphics[width=0.8\linewidth]{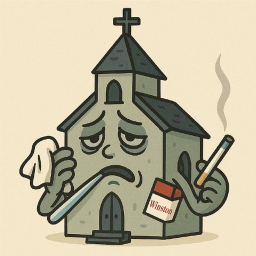}
\end{minipage}
\end{center}

"primary\_clues": [
            "The central image is the flag of France.",
            "The flag is enclosed or surrounded by a circle.",
            "The prefix for 'around' or 'surrounding' is 'Circum-'.",
            "Combining the prefix with the country name creates 'Circum-France'.",
            "Phonetically, this sounds like the geometric term 'Circumference'."
        ]\\
        
        "candidates": [
            "France Circles", 
            "Rounded France", 
            "Circumference"  
        ]\\
        
\textbf{Example 2:}

"primary\_clues": [
            "A man sits in a corner with tattered clothes and a bowl containing broken scraps, symbolizing having lost everything.",
            "His desolate condition and the empty/broken contents of his begging bowl indicate he is 'in need'.",
            "A thought bubble above his head displays the letter 'L'.",
            "Combining his condition ('Need') with the letter ('L') creates the construction 'Need-L'.",
            "Phonetically, this sounds like the word 'Needle'."
        ]
        
        "candidates": [
            "Needs L", 
            "L Wants", 
            "Needle" 
        ]\\
        
\textbf{Example 3:}

"primary\_clues": [
            "The image depicts a personified church building.",
            "The church appears sick, with a thermometer in its mouth and a tissue in its hand, indicating it is 'ill'.",
            "This combines to form 'Church' + 'ill', sounding like 'Churchill'.",
            "The church is also holding a pack of cigarettes clearly labeled 'Winston'.",
            "Combining the brand name with the other elements creates 'Winston Churchill'."
        ]
        
        "candidates": [
            "Sick Church Smokes",
            "Winstons Sick Home",
            "Winston Churchill"
        ]

\end{PromptEN}

\begin{PromptCL}[Cross-Lingual Few-shot CoT]{}

\begin{center}
\begin{minipage}[t]{0.32\linewidth}
  \centering
  \includegraphics[width=0.8\linewidth]{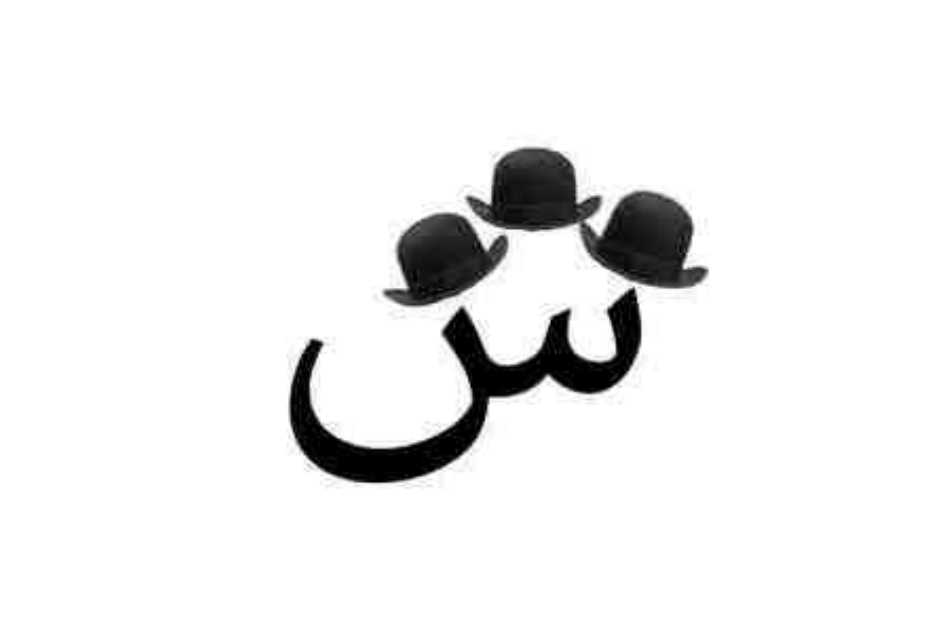}
\end{minipage}\hfill
\begin{minipage}[t]{0.32\linewidth}
  \centering
  \includegraphics[width=0.8\linewidth]{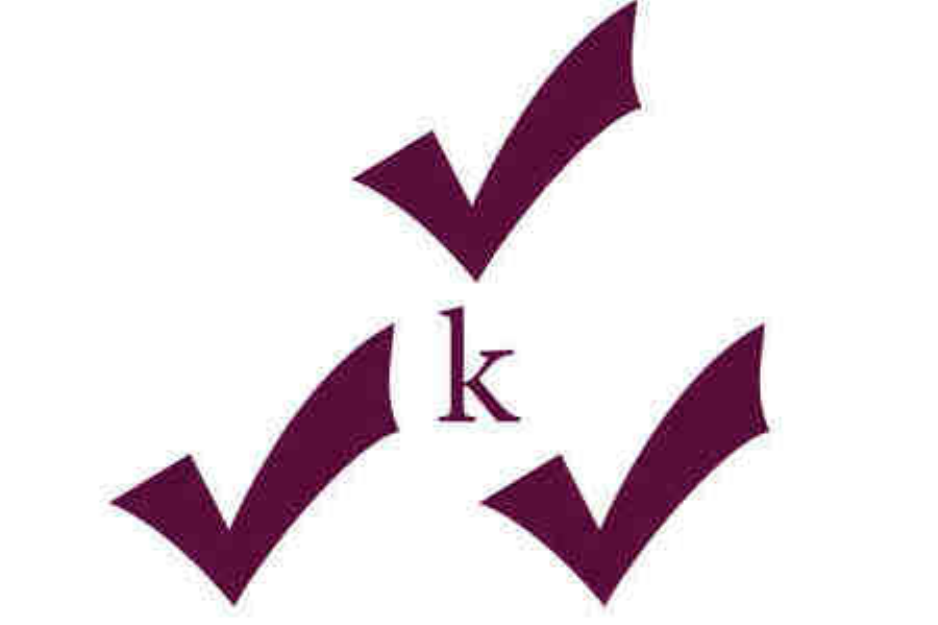}
\end{minipage}\hfill
\begin{minipage}[t]{0.32\linewidth}
  \centering
  \includegraphics[width=0.8\linewidth]{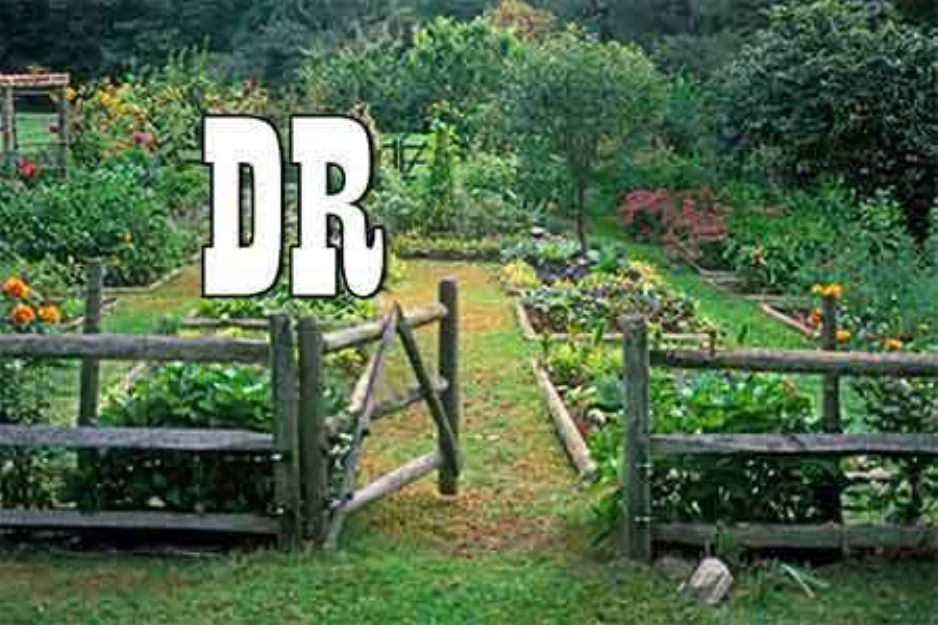}
\end{minipage}
\end{center}

\textbf{Example 1:}

"primary\_clues": [
            "The image shows the Persian letter 'Seen' \foreignlanguage{Persian}{(س)}.",
            "Placed on top ('Roosh' - روش) of the 'Seen' are three ('Se' - \foreignlanguage{Persian}{سه}) hats ('Hat' - \foreignlanguage{Persian}{هت}).",
            "Combining 'Seen' + 'Roosh' creates the name 'Soroush' \foreignlanguage{Persian}{(سروش)}.",
            "Combining 'Se' (Three) + 'Hat' (English word for hat) sounds like 'Sehhat' \foreignlanguage{Persian}{(صحت)}.",
            "Together, they form the name of the famous Iranian director 'Soroush Sehhat' \foreignlanguage{Persian}{(سروش صحت)}."
        ]
        
        "candidates": [
            \foreignlanguage{Persian}{"سه کلاه س"},
            \foreignlanguage{Persian}{"کلاه سه س"},
            \foreignlanguage{Persian}{"سروش صحت"} 
        ]\\

\textbf{Example 2:}

"primary\_clues": [
            "The image shows the letter 'k' surrounded by three checkmarks (ticks).",
            "The letter 'k' is in the middle, which translates to 'Mian' \foreignlanguage{Persian}{(میان)} in Persian.",
            "Combining 'K' + 'Mian' phonetically sounds like 'Kamion' \foreignlanguage{Persian}{(کامیون)}, meaning Truck.",
            "The 'k' is placed among ('La' - \foreignlanguage{Persian}{لا}) the three ('Se' - \foreignlanguage{Persian}{سه}) ticks ('Tick' - \foreignlanguage{Persian}{تیک}).",
            "The phrase 'La-Se-Tick' \foreignlanguage{Persian}{(لا سه تیک)} sounds like 'Lastik' \foreignlanguage{Persian}{(لاستیک)}, meaning Tire.",
            "Together, the visual pun creates 'Lastik Kamion' \foreignlanguage{Persian}{(لاستیک کامیون)}."
        ]
        
        "candidates": [
            \foreignlanguage{Persian}{"لاستیک اتوبوس"},
            \foreignlanguage{Persian}{"لاستیک دوچرخه"},
            \foreignlanguage{Persian}{"لاستیک کامیون"}
        ]\\

\textbf{Example 3:}

"primary\_clues": [
            "The image features a garden, which translates to 'Bagh' \foreignlanguage{Persian}{(باغ)} in Persian.",
            "The letters 'DR' are superimposed on the scene.",
            "Phonetically, the English letters 'DR' (Dee-Ar) sound identical to the Persian word 'Diar' \foreignlanguage{Persian}{(دیار)}, meaning 'Land' or 'Realm'.",
            "Combining the two components yields 'Diar' + 'Bagh'.",
            "This creates a pun on the common phrase 'Diar-e Baghi' \foreignlanguage{Persian}{(دیار باقی)}, which means 'The Eternal Realm' or 'The Afterlife'.",
            "The pun works by substituting the word 'Bagh' (Garden) for the phonetically similar 'Baghi' (Eternal)."
        ]
        
        "candidates": [
            \foreignlanguage{Persian}{"دکتر باغی"},
            \foreignlanguage{Persian}{"باغ دیوار"},
            \foreignlanguage{Persian}{"دیار باقی"}
        ]
\end{PromptCL}

\begin{PromptAR}[Arabic Language Few-shot CoT]{}

\begin{center}
\begin{minipage}[t]{0.32\linewidth}
  \centering
  \includegraphics[width=0.8\linewidth]{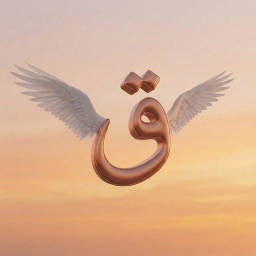}
\end{minipage}\hfill
\begin{minipage}[t]{0.32\linewidth}
  \centering
  \includegraphics[width=0.8\linewidth]{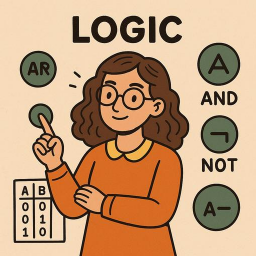}
\end{minipage}\hfill
\begin{minipage}[t]{0.32\linewidth}
  \centering
  \includegraphics[width=0.8\linewidth]{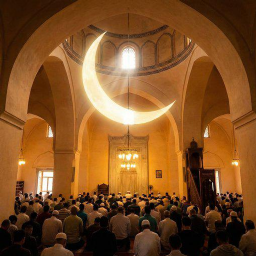}
\end{minipage}
\end{center}

\textbf{Example 1:}

"primary\_clues": [
            "The image features the Arabic letter 'Qaf' \foreignlanguage{Arabic}{(ق)}.",
            "The letter has wings attached to it, indicating the action of flying.",
            "The Arabic verb for 'he flew' is 'Tara' \foreignlanguage{Arabic}{(طار)}.",
            "Combining the letter 'Q' \foreignlanguage{Arabic}{(ق)} with the verb 'Tara' \foreignlanguage{Arabic}{(طار)} creates the sequence 'Q-Tara'.",
            "This forms the word 'Qitar' \foreignlanguage{Arabic}{(قطار)}.",
            "The word 'Qitar' \foreignlanguage{Arabic}{(قطار)} means 'Train' in Arabic."
        ]
        
        "candidates": [
            \foreignlanguage{Arabic}{"قطار"},
            \foreignlanguage{Arabic}{"طائر"},
            \foreignlanguage{Arabic}{"طارق"}
        ]\\

\textbf{Example 2:}

"primary\_clues": [
            "The image features the word 'LOGIC' along with truth tables and logic gates.",
            "The concept of 'Logic' translates to 'Mantiq' \foreignlanguage{Arabic}{(منطق)} in Arabic.",
            "The central character is female, indicating the presence of a 'feminine marker'.",
            "In Arabic morphology, the feminine marker is added as 'Ta Marbuta' \foreignlanguage{Arabic}{(ـة)}.",
            "Combining the word 'Mantiq' \foreignlanguage{Arabic}{(منطق)} with the feminine suffix \foreignlanguage{Arabic}{(ـة)} yields 'Mintaqah' \foreignlanguage{Arabic}{(منطقة)}.",
            "While the base word means Logic, the resulting word 'Mintaqah' \foreignlanguage{Arabic}{(منطقة)} means 'Area' or 'Zone'."
        ]
        
        "candidates": [
            \foreignlanguage{Arabic}{"منطقة"},
            \foreignlanguage{Arabic}{"منطقي"},
            \foreignlanguage{Arabic}{"منظمة"}
        ]\\
        
\textbf{Example 3:}

"primary\_clues": [
            "The image features a large, bright crescent moon inside a structure.",
            "The Arabic word for 'Moon' is 'Qamar' \foreignlanguage{Arabic}{(قمر)}.",
            "The setting is a mosque filled with worshippers, symbolizing 'Religion' or 'Faith', which translates to 'Din' \foreignlanguage{Arabic}{(دين)} in Arabic.",
            "Combining the word for Moon 'Qamar' and Religion 'Din' yields the compound 'Qamar al-Din' \foreignlanguage{Arabic}{(قمر الدين)}.",
            "This forms the term 'Qamar al-Din', which refers to sheets of dried apricot paste."
        ]
        
        "candidates": [
            \foreignlanguage{Arabic}{"قمر الدين"},
            \foreignlanguage{Arabic}{"شمس الدين"},
            \foreignlanguage{Arabic}{"نور الدين"}
        ]

\end{PromptAR}

\FloatBarrier

\section{Model Output Examples}

This section presents qualitative examples of model behavior on our multilingual visual word-puzzle benchmark for two systems, \textbf{Gemini 2.5 Pro} and \textbf{Grok 4.1 Fast}. We evaluated the models in four languages and under four experimental variations, which results in \textbf{16 different setups} for each model. For every setup, we report two examples. The examples are reported by language and then by experimental variation. They are ordered as English, Persian, Arabic, and cross-lingual. The examples are not selected uniformly at random. Instead, we focus on edge cases that are particularly useful for analyzing model capabilities and failure modes. These include borderline correct cases that require non-trivial compositional reasoning, as well as near-miss failures where the model produces a plausible but incorrect answer.
Each example includes the input image, the corresponding variation configuration such as hint type and context settings, the ground-truth answer, and the model output. This structure allows direct comparison across languages and experimental conditions.

\begin{PromptEN}[Basic setup]{sidebyside, lefthand ratio=0.7}
Task: Solve the picture word puzzle from the image.\\
Target language: English.\\
Hint type: char\_count.

\begin{tcolorbox}[gtstyle]
        roadtrip
\end{tcolorbox}

\tcblower

\centering
\includegraphics[width=\linewidth]{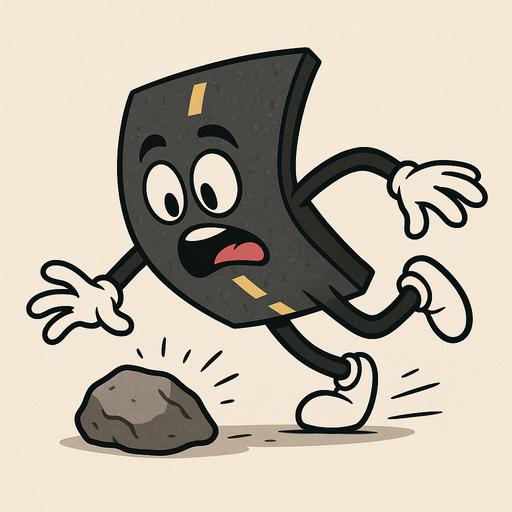}
\end{PromptEN}

\begin{ModelResponse}{gemini-2.5-pro}
{
  "primary_clues": [
    "A character that is a piece of a road", "The action of tripping over a rock"
  ],
  "candidates": [
    "Road trip", "Roadblock", "Hit rock bottom", "A bump in the road"
  ],
  "final_answer": "Road trip"
}
\end{ModelResponse}

\begin{ModelResponse}{grok-4.1-fast}
{
  "primary_clues": [
    "road", "tripping", "rock"
  ],
  "candidates": [
    "rocky road", "roadblock", "roadtrip"
  ],
  "final_answer": "roadtrip"
}
\end{ModelResponse}

\begin{PromptEN}[Basic setup]{sidebyside, lefthand ratio=0.7}
Task: Solve the picture word puzzle from the image.\\
Target language: English.\\
Hint type: char\_count.

\begin{tcolorbox}[gtstyle]
        harry potter
\end{tcolorbox}

\tcblower

\centering
\includegraphics[width=\linewidth]{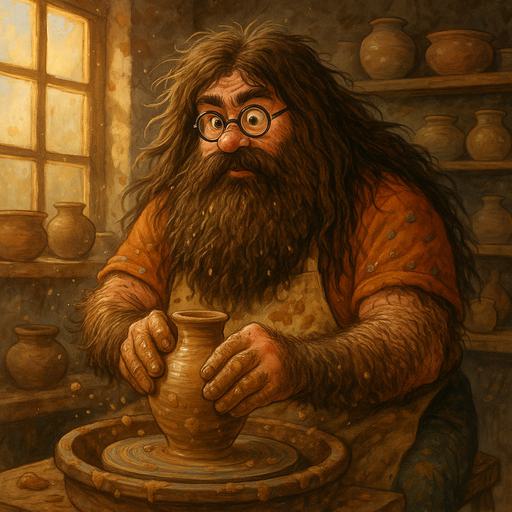}
\end{PromptEN}

\begin{ModelResponse}{gemini-2.5-pro}
{
  "primary_clues": [
    "A hairy man resembling Hagrid", "Making a pot on a potter's wheel"
  ],
  "candidates": [
    "Hairy Potter", "Hagrid's Pottery", "The Potter's Wheel"
  ],
  "final_answer": "Hairy Potter"
}
\end{ModelResponse}

\begin{ModelResponse}{grok-4.1-fast}
{
  "primary_clues": [
    "hairy man", "potter (pottery wheel)"
  ],
  "candidates": [
    "harry potter", "hairy potter", "potter harry"
  ],
  "final_answer": "harry potter"
}
\end{ModelResponse}

\begin{PromptEN}[Example: Few-shot chain-of-thought prompting]{sidebyside, lefthand ratio=0.7}
Task: Solve the picture word puzzle from the image.\\
Target language: English.\\
Hint type: char\_count.

\begin{tcolorbox}[gtstyle]
        Fishtank
\end{tcolorbox}

\tcblower

\centering
\includegraphics[width=\linewidth]{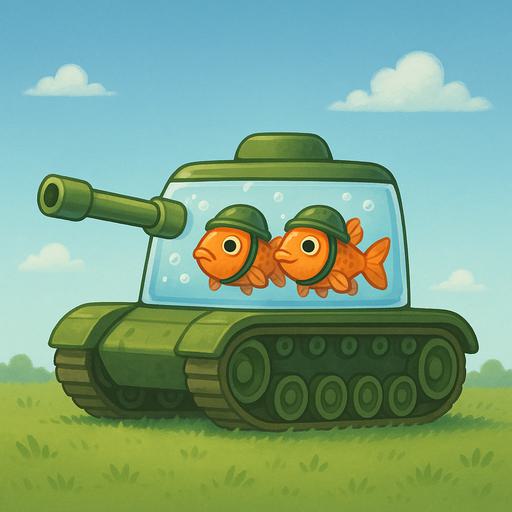}
\end{PromptEN}

\begin{ModelResponse}{gemini-2.5-pro}
{
  "primary_clues": [
    "The turret of the army tank is a glass container holding the fish, which is a fish tank.", 
    "Two fish wearing helmets.", "An army tank."
    
  ],
  "candidates": [
    "Aquarium", "Tank Fish", "Fish Tank"
  ],
  "final_answer": "Fish Tank"
}
\end{ModelResponse}

\begin{ModelResponse}{grok-4.1-fast}
{
  "primary_clues": [
    "military tank with transparent body and bubbles", 
    "two orange fish inside wearing green helmets"
  ],
  "candidates": [
    "army goldfish", "wet soldier fish", "Fishtank"
  ],
  "final_answer": "Fishtank"
}
\end{ModelResponse}

\begin{PromptEN}[Example: Few-shot chain-of-thought prompting]{sidebyside, lefthand ratio=0.7}
Task: Solve the picture word puzzle from the image.\\
Target language: English.\\
Hint type: char\_count.

\begin{tcolorbox}[gtstyle]
        Deserted island
\end{tcolorbox}

\tcblower

\centering
\includegraphics[width=\linewidth]{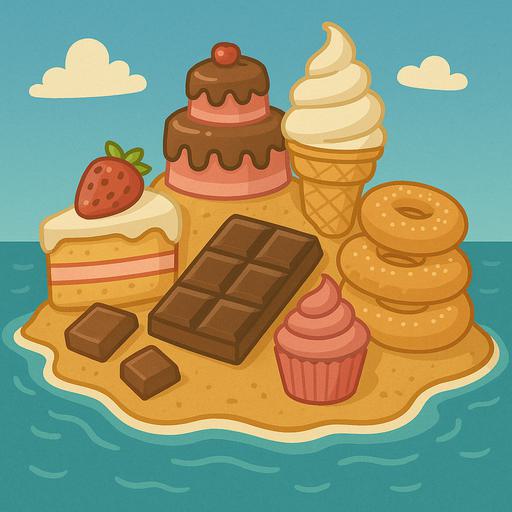}
\end{PromptEN}

\begin{ModelResponse}{gemini-2.5-pro}
{
  "primary_clues": [
    "An island in the ocean.",
    "The island is uninhabited, which makes it a 'deserted' island.",
    "The island is covered entirely with sweets, also known as 'desserts'.",
    "A pun is formed by combining 'dessert' with '-ed' to create      'desserted', which sounds like 'deserted'."
  ],
  "candidates": [
    "Dessert island", "Sweet island", "Desserted island"
  ],
  "final_answer": "Desserted island"
}
\end{ModelResponse}

\begin{ModelResponse}{grok-4.1-fast}
{
  "primary_clues": [
    "a small island surrounded by ocean",
    "the island is made of desserts (cakes, donuts, ice cream, chocolate)",
    "no people present (empty/abandoned)"
  ],
  "candidates": [
    "dessert island", "desert island", "Deserted island"
  ],
  "final_answer": "Deserted island"
}
\end{ModelResponse}

\begin{PromptEN}[Iterative refinement]{sidebyside, lefthand ratio=0.7}
Task: Solve the picture word puzzle from the image.\\
Target language: English.\\
Hint type: char\_count.\\
Hint: The answer has 20 characters (excluding spaces).

\begin{tcolorbox}[gtstyle]
        Scholar of the first sin
\end{tcolorbox}

\tcblower

\centering
\includegraphics[width=\linewidth]{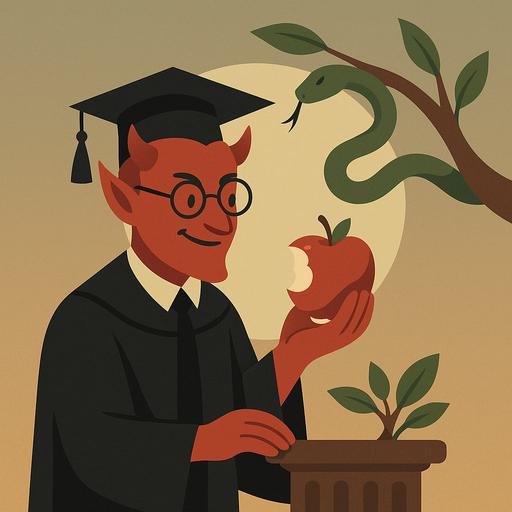}
\end{PromptEN}

\begin{ModelResponse}{gemini-2.5-pro}
"attempts": 
{
  "attempt_idx": 0,
  "model_output": {
    "primary_clues": [
      "A devil figure", "A person wearing a graduation cap and gown", 
      "Symbols of knowledge and temptation (apple, snake)"
    ],
    "candidates": [
      "Devil's Advocate", "Sympathy for the Devil",  "Devil's Graduate School"
    ],
    "final_answer": "Devil's Graduate School"
  },
  "feedback": "The answer 'Devil's Graduate School' is incorrect. Please Step-by-Step."
},
{
  "attempt_idx": 1,
  "model_output": {
    "primary_clues": [
      "A devil figure in academic regalia", "A sapling representing an 'origin'", 
      "Biblical symbols of evil (apple, snake)"
    ],
    "candidates": [
      "On the Origins of Evil", "Knowledge is the root of all evil", "Devil's Advocate"
    ],
    "final_answer": "On the Origins of Evil"
  },
  "feedback": "The answer '' is incorrect. Please Step-by-Step."
},
{
  "attempt_idx": 2,
  "model_output": {
    "primary_clues": [
      "A devil dressed as a scholar or professor", "Symbols of temptation (apple and snake)",
      "A known work of literature with a 20-character title"
    ],
    "candidates": [
      "The Screwtape Letters", "On the Origins of Evil", "Sympathy for the Devil"
    ],
    "final_answer": "The Screwtape Letters"
  }
}
\end{ModelResponse}

\begin{ModelResponse}{grok-4.1-fast}
"attempts": [
{
  "attempt_idx": 0,
  "primary_clues": [
    "horned graduate", "bitten apple", "snake on branch"
  ],
  "candidates": [
    "tree of good and evil", "the tree of good and evil", "an apple for the teacher"
  ],
  "final_answer": "an apple for the teacher"
},
{
  "attempt_idx": 1,
  "primary_clues": [
    "devil graduate", "bitten apple", "snake on branch"
  ],
  "candidates": [
    "forbidden fruit", "tree of good and evil", "the tree of good and evil"
  ],
  "final_answer": "the tree of good and evil"
},
{
  "attempt_idx": 2,
  "primary_clues": [
    "a devil figure wearing academic cap and gown (a scholar/teacher)",
    "a bitten apple", "a snake on a branch (Eden / the first sin)"
  ],
  "candidates": [
    "Scholar of the first sin", "The first sin scholar", "Professor of forbidden fruit"
  ],
  "final_answer": "Scholar of the first sin"
}
\end{ModelResponse}

\begin{PromptEN}[Iterative refinement]{sidebyside, lefthand ratio=0.7}
Task: Solve the picture word puzzle from the image.\\
Target language: English.\\
Hint type: char\_count.\\
Hint: The answer has 9 characters (excluding spaces).

\begin{tcolorbox}[gtstyle]
        Hat-trick
\end{tcolorbox}

\tcblower

\centering
\includegraphics[width=\linewidth]{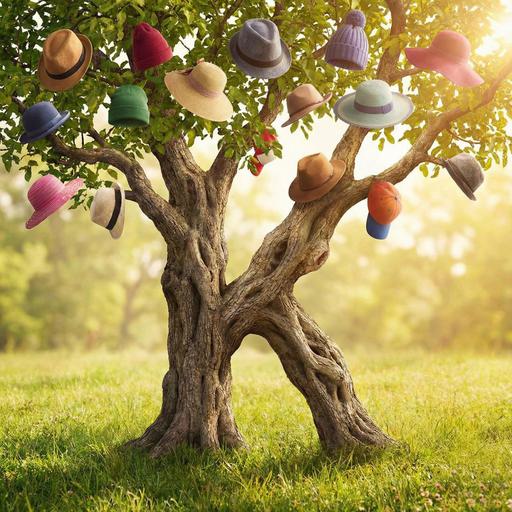}
\end{PromptEN}

\begin{ModelResponse}{gemini-2.5-pro}
"attempts": [
{
  "attempt_idx": 0,
  "model_output": {
    "primary_clues": [
      "A plurality of hats", "A tree", "A tree trunk shaped like the letter 'K'"
    ],
    "candidates": [
      "Hat trick", "Hat-trick", "Hat tricks", "Millinery"
    ],
    "final_answer": "Hat tricks"
  },
  "feedback": "The answer 'Hat tricks' is incorrect. Please Step-by-Step:"
},
{
  "attempt_idx": 1,
  "model_output": {
    "primary_clues": [
      "An impossible scene of hats on a tree, suggesting magic", 
      "A tree trunk shaped like the letter 'K'"
    ],
    "candidates": [
      "Hat tricks", "Family tree", "Magic trick"
    ],
    "final_answer": "Magic trick"
  },
  "feedback": "The answer 'Magic trick' is incorrect. Please Step-by-Step:"
},
{
  "attempt_idx": 2,
  "model_output": {
    "primary_clues": [
      "A plurality of hats", "A tree", "A tree trunk shaped like the letter 'K'"
    ],
    "candidates": [
      "Hat tricks", "Magic trick", "Hattricks"
    ],
    "final_answer": "Hattricks"
  }
}
\end{ModelResponse}

\begin{ModelResponse}{grok-4.1-fast}
  "attempts": [
    {
      "attempt_idx": 0,
      "primary_clues": [
        "tree", "hats"
      ],
      "candidates": [
        "hat tree", "mad hatter", "three hats"
      ],
      "final_answer": "three hats"
    },
    {
      "attempt_idx": 1,
      "primary_clues": [
        "many hats", "K-shaped tree"
      ],
      "candidates": [
        "hat tree", "hat trick", "mad hatter"
      ],
      "final_answer": "mad hatter"
    },
    {
      "attempt_idx": 2,
      "primary_clues": [
        "three hats", "a tree shaped like the letter K (hinting 'K' in 'trick')"
      ],
      "candidates": [
        "Hat-trick", "hat trick", "hat tricks"
      ],
      "final_answer": "Hat-trick"
    }
\end{ModelResponse}

\begin{PromptEN}[Partial character reveal]{sidebyside, lefthand ratio=0.7}
Task: Solve the picture word puzzle from the image.\\
Target language: English.\\
Hint type: shuffle\_chars.\\
Hint pattern: \verb|__t__r___|.

\begin{tcolorbox}[gtstyle]
        Hat-trick
\end{tcolorbox}

\tcblower

\centering
\includegraphics[width=\linewidth]{figures/en_291.jpg}
\end{PromptEN}

\begin{ModelResponse}{gemini-2.5-pro}
{
  "primary_clues": [
    "A collection of various hats", "A tree",
    "The tree's trunk and branches are shaped like the letter 'K'"
  ],
  "candidates": [
    "Hat trick", "Hat tree", "Patronage"
  ],
  "final_answer": "Hat trick"
}
\end{ModelResponse}

\begin{ModelResponse}{grok-4.1-fast}
{
  "primary_clues": [
    "multiple hats", "a tree that looks like a rack/stand", 
    "a clear hint toward the word 'trick' (visual pun setup)"
  ],
  "candidates": [
    "Hat-trick", "hat trick", "hattrick"
  ],
  "final_answer": "Hat-trick"
}
\end{ModelResponse}

\begin{PromptEN}[Partial character reveal]{sidebyside, lefthand ratio=0.7}
Task: Solve the picture word puzzle from the image.\\
Target language: English.\\
Hint type: shuffle\_chars.\\
Hint pattern: \verb|__n___o__|.

\begin{tcolorbox}[gtstyle]
        singapore
\end{tcolorbox}

\tcblower

\centering
\includegraphics[width=\linewidth]{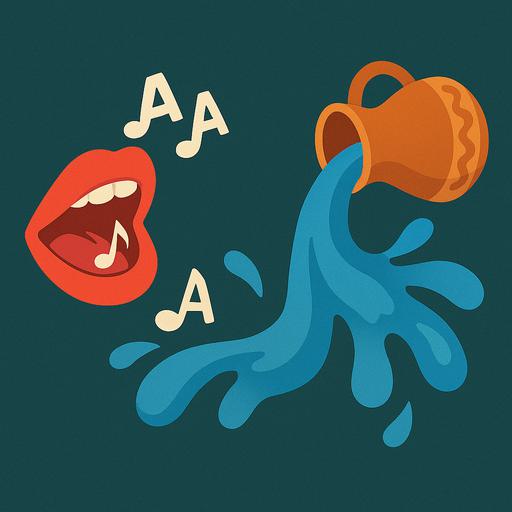}
\end{PromptEN}

\begin{ModelResponse}{gemini-2.5-pro}
{
  "primary_clues": [
    "A mouth singing a sustained note ('A')", "A pitcher pouring a continuous stream of water", 
    "The letter pattern is __n___o__"
  ],
  "candidates": [
    "Tenacious", "Aquarius", "Monotonous"
  ],
  "final_answer": "Tenacious"
}

\end{ModelResponse}

\begin{ModelResponse}{grok-4.1-fast}
{
  "primary_clues": [
    "the wordplay cue 'sing'", "a 'pore' (visible skin pore / pore-like opening)", 
    "the combined sound suggests a place name"
  ],
  "candidates": [
    "singapore", "sing a pore", "singa pore"
  ],
  "final_answer": "singapore"
}
\end{ModelResponse}

\begin{PromptFA}[Basic setup]{sidebyside, lefthand ratio=0.7}
Task: Solve the picture word puzzle from the image.\\
Target language: Persian (Farsi).\\
Hint type: char\_count.\\
Hint: The answer has 7 characters (excluding spaces).

\begin{tcolorbox}[gtstyle]
        \foreignlanguage{arabic}{گلاب پاش}
\end{tcolorbox}

\tcblower

\centering
\includegraphics[width=\linewidth]{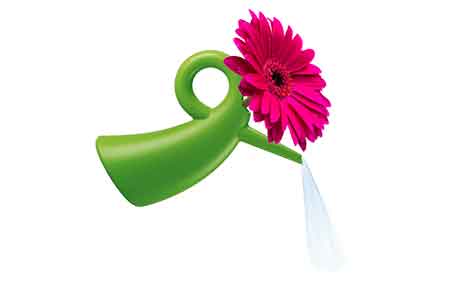}
\end{PromptFA}

\begin{ModelResponse}{gemini-2.5-pro}
{
  "primary_clues": [
    "(@\foreignlanguage{arabic}{گل}@) (flower)", "(@\foreignlanguage{arabic}{آبپاش}@) (watering can)", "(@\foreignlanguage{arabic}{آب}@) (water)"
  ],
  "candidates": [
    "(@\foreignlanguage{arabic}{گلاب}@)", "(@\foreignlanguage{arabic}{آبپاش}@)", "(@\foreignlanguage{arabic}{گلابپاش}@)"
  ],
  "final_answer": "(@\foreignlanguage{arabic}{گلابپاش}@)"
}
\end{ModelResponse}

\begin{ModelResponse}{grok-4.1-fast}
{
  "primary_clues": [
    "(@\foreignlanguage{arabic}{آب‌پاش}@)", "(@\foreignlanguage{arabic}{گل}@)"
  ],
  "candidates": [
    "(@\foreignlanguage{arabic}{گل‌افشان}@)", "(@\foreignlanguage{arabic}{باران گل}@)", "(@\foreignlanguage{arabic}{آبشار گل}@)", "(@\foreignlanguage{arabic}{گلاب پاش}@)"
  ],
  "final_answer": "(@\foreignlanguage{arabic}{گلاب پاش}@)"
}
\end{ModelResponse}

\begin{PromptFA}[Basic setup]{sidebyside, lefthand ratio=0.7}
Task: Solve the picture word puzzle from the image.\\
Target language: Persian (Farsi).\\
Hint type: char\_count.\\
Hint: The answer has 14 characters (excluding spaces).

\begin{tcolorbox}[gtstyle]
        \foreignlanguage{arabic}{دسته گل به آب دادن}
\end{tcolorbox}

\tcblower

\centering
\includegraphics[width=\linewidth]{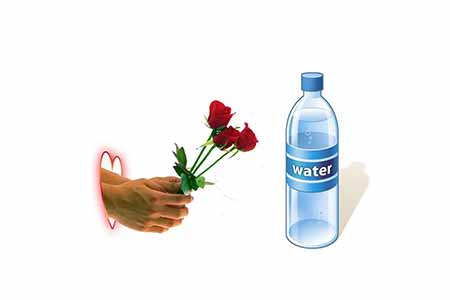}
\end{PromptFA}

\begin{ModelResponse}{gemini-2.5-pro}
{
  "primary_clues": [
    "(@\foreignlanguage{arabic}{دست و گل}@)", "(@\foreignlanguage{arabic}{آب}@)"
  ],
  "candidates": [
    "(@\foreignlanguage{arabic}{دست گل به آب دادن}@)", "(@\foreignlanguage{arabic}{گلاب}@)", "(@\foreignlanguage{arabic}{آب دادن پای گل}@)"
  ],
  "final_answer": "(@\foreignlanguage{arabic}{دست گل به آب دادن}@)"
}
\end{ModelResponse}

\begin{ModelResponse}{grok-4.1-fast}
{
  "primary_clues": [
    "(@\foreignlanguage{arabic}{دسته گل}@)", "(@\foreignlanguage{arabic}{آب}@)"
  ],
  "candidates": [
    "(@\foreignlanguage{arabic}{گل به آب دادن}@)", "(@\foreignlanguage{arabic}{دسته گل به آب دادن}@)", "(@\foreignlanguage{arabic}{آب دادن به گل}@)"
  ],
  "final_answer": "(@\foreignlanguage{arabic}{دسته گل به آب دادن}@)"
}
\end{ModelResponse}

\begin{PromptFA}[Few-shot chain-of-thought prompting]{sidebyside, lefthand ratio=0.7}
Task: Solve the picture word puzzle from the image.\\
Target language: Persian (Farsi).\\
Hint type: char\_count.\\
Context: Few-shot examples are provided in the prompt.

\begin{tcolorbox}[gtstyle]
        \foreignlanguage{farsi}{توپ خانه}
\end{tcolorbox}

\tcblower

\centering
\includegraphics[width=\linewidth]{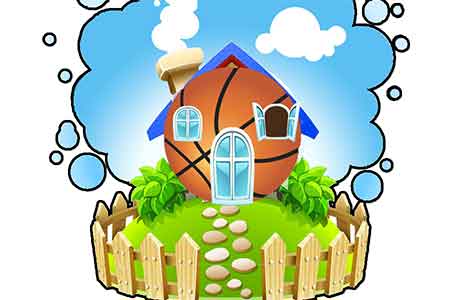}
\end{PromptFA}

\begin{ModelResponse}{gemini-2.5-pro}
{
  "primary_clues": [
    "A house, which is 'Khaneh' ((@\foreignlanguage{farsi}{خانه}@)) in Persian.",
    "A basketball, which is a type of ball, 'Toop'((@\foreignlanguage{farsi}{توپ}@)) in Persian."
  ],
  "candidates": [
    "(@\foreignlanguage{farsi}{خانه توپ}@)", "(@\foreignlanguage{farsi}{منزل بسکتبال}@)", "(@\foreignlanguage{farsi}{توپخانه}@)"
  ],
  "final_answer": "(@\foreignlanguage{farsi}{توپخانه}@)"
}
\end{ModelResponse}

\begin{ModelResponse}{grok-4.1-fast}
{
  "primary_clues": [
    "(@\foreignlanguage{arabic}{خانه‌ای که از توپ (بسکتبال) ساخته شده}@)", "(@\foreignlanguage{arabic}{ترکیب «توپ» + «خانه»}@)"
  ],
  "candidates": [
    "(@\foreignlanguage{arabic}{خانه توپ}@)", "(@\foreignlanguage{arabic}{توپ خانه}@)", "(@\foreignlanguage{arabic}{توپخانه}@)"
  ],
  "final_answer": "(@\foreignlanguage{arabic}{توپخانه}@)"
}
\end{ModelResponse}

\begin{PromptFA}[Few-shot chain-of-thought prompting]{sidebyside, lefthand ratio=0.7}
Task: Solve the picture word puzzle from the image.\\
Target language: Persian (Farsi).\\
Hint type: char\_count.\\
Context: Few-shot examples are provided in the prompt.

\begin{tcolorbox}[gtstyle]
        \foreignlanguage{farsi}{سرپرست}
\end{tcolorbox}

\tcblower

\centering
\includegraphics[width=\linewidth]{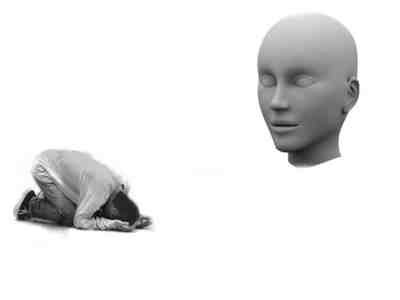}
\end{PromptFA}

\begin{ModelResponse}{gemini-2.5-pro}
{
  "primary_clues": [
    "A person in the act of prostration or worship, which implies 'Parastesh' ((@\foreignlanguage{farsi}{پرستش}@)).",
    "A generic, statue-like head, which represents an 'idol' or 'Bot' ((@\foreignlanguage{farsi}{بت}@)) in Persian."
  ],
  "candidates": [
    "(@\foreignlanguage{farsi}{سر به سجده}@)", "(@\foreignlanguage{farsi}{طاغوت}@)", "(@\foreignlanguage{farsi}{بت‌پرست}@)"
  ],
  "final_answer": "(@\foreignlanguage{farsi}{بت‌پرست}@)"
}
\end{ModelResponse}

\begin{ModelResponse}{grok-4.1-fast}
{
  "primary_clues": [
    "(@\foreignlanguage{arabic}{سرِ انسان (سر)}@)", "(@\foreignlanguage{arabic}{فرد در حال سجده/پرستش (پرست)}@)"
  ],
  "candidates": [
    "(@\foreignlanguage{arabic}{سرپرست}@)", "(@\foreignlanguage{arabic}{سرپرستی}@)", "(@\foreignlanguage{arabic}{پرستار}@)"
  ],
  "final_answer": "(@\foreignlanguage{arabic}{سرپرست}@)"
}
\end{ModelResponse}

\begin{PromptFA}[Iterative refinement]{sidebyside, lefthand ratio=0.7}
Task: Solve the picture word puzzle from the image.\\
Target language: Persian (Farsi).\\
Hint type: char\_count.\\
Hint: The answer has 8 characters (excluding spaces).

\begin{tcolorbox}[gtstyle]
        \foreignlanguage{arabic}{لیست سیاه}
\end{tcolorbox}

\tcblower

\centering
\includegraphics[width=\linewidth]{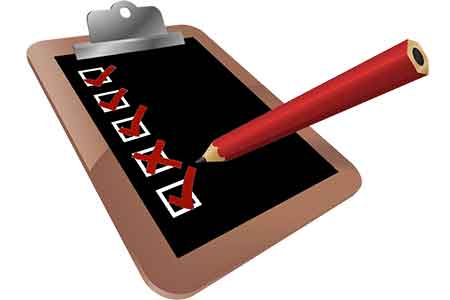}
\end{PromptFA}

\begin{ModelResponse}{gemini-2.5-pro}
"attempts": [
{
  "attempt_idx": 0,
  "model_output": {
    "primary_clues": [
      "A checklist being marked", "A red pencil", "The action of reviewing or checking"
    ],
    "candidates": [
      "(@\foreignlanguage{arabic}{خط قرمز}@)", "(@\foreignlanguage{arabic}{بررسی کردن}@)", "(@\foreignlanguage{arabic}{چک لیست}@)"
    ],
    "final_answer": "(@\foreignlanguage{arabic}{بررسی کردن}@)"
  },
  "feedback": "The answer '(@\foreignlanguage{arabic}{بررسی کردن}@)' is incorrect. Please Step-by-Step:"
},
{
  "attempt_idx": 1,
  "model_output": {
    "primary_clues": [
      "A pencil/pen, which is '(@\foreignlanguage{arabic}{قلم}@)' in Persian",
      "The action of crossing an item out, which signifies finding fault, as in '(@\foreignlanguage{arabic}{ایراد گرفتن}@)'"
    ],
    "candidates": [
      "(@\foreignlanguage{arabic}{قلم گرفتن}@)", "(@\foreignlanguage{arabic}{سرشماری}@)", "(@\foreignlanguage{arabic}{خط قرمز}@)"
    ],
    "final_answer": "(@\foreignlanguage{arabic}{قلم گرفتن}@)"
  },
  "feedback": "The answer '(@\foreignlanguage{arabic}{قلم گرفتن}@)' is incorrect. Please Step-by-Step:"
},
{
  "attempt_idx": 2,
  "model_output": {
    "primary_clues": [
      "A list or form, which can be called '(@\foreignlanguage{arabic}{صورت}@)' in Persian",
      "The action of checking or tallying, related to '(@\foreignlanguage{arabic}{حساب}@)' (accounting)"
    ],
    "candidates": [
      "(@\foreignlanguage{arabic}{صورت حساب}@)", "(@\foreignlanguage{arabic}{قلم گرفتن}@)", "(@\foreignlanguage{arabic}{بررسی کردن}@)"
    ],
    "final_answer": "(@\foreignlanguage{arabic}{صورت حساب}@)"
  }
}
\end{ModelResponse}

\begin{ModelResponse}{grok-4.1-fast}
"attempts": [
{
  "attempt_idx": 0,
  "primary_clues": [
    "(@\foreignlanguage{arabic}{یک لیست/چک‌لیست روی کلیپ‌بُرد}@)", "(@\foreignlanguage{arabic}{علامت‌گذاری روی آیتم‌ها}@)"
  ],
  "candidates": [
    "(@\foreignlanguage{arabic}{چک لیست}@)", "(@\foreignlanguage{arabic}{لیست کامل}@)", "(@\foreignlanguage{arabic}{لیست سیاه}@)"
  ],
  "final_answer": "(@\foreignlanguage{arabic}{چک لیست}@)"
},
{
  "attempt_idx": 1,
  "primary_clues": [
    "(@\foreignlanguage{arabic}{لیست آیتم‌ها}@)", "(@\foreignlanguage{arabic}{علامت‌گذاری به شکل «منفی/رد کردن» (ضربدر/خط زدن)}@)"
  ],
  "candidates": [
    "(@\foreignlanguage{arabic}{لیست ممنوع}@)", "(@\foreignlanguage{arabic}{لیست سیاه}@)", "(@\foreignlanguage{arabic}{خط قرمز}@)"
  ],
  "final_answer": "(@\foreignlanguage{arabic}{لیست ممنوع}@)"
},
{
  "attempt_idx": 2,
  "primary_clues": [
    "(@\foreignlanguage{arabic}{یک «لیست» واضح}@)", "(@\foreignlanguage{arabic}{تأکید تصویری روی «سیاه/منفی‌بودن» لیست (رد شدن/علامت منفی)}@)"
  ],
  "candidates": [
    "(@\foreignlanguage{arabic}{لیست سیاه}@)", "(@\foreignlanguage{arabic}{لیست قرمز}@)", "(@\foreignlanguage{arabic}{لیست ممنوعه}@)"
  ],
  "final_answer": "(@\foreignlanguage{arabic}{لیست سیاه}@)"
}
\end{ModelResponse}

\begin{PromptFA}[Iterative refinement]{sidebyside, lefthand ratio=0.7}
Task: Solve the picture word puzzle from the image.\\
Target language: Persian (Farsi).\\
Hint type: char\_count.

\begin{tcolorbox}[gtstyle]
        \foreignlanguage{arabic}{گرگ باران زده}
\end{tcolorbox}

\tcblower

\centering
\includegraphics[width=\linewidth]{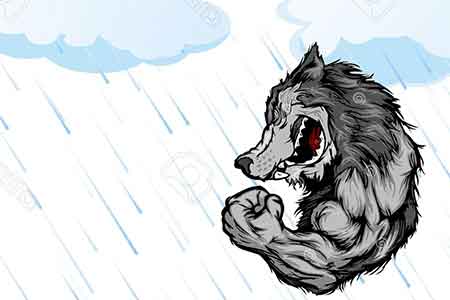}
\end{PromptFA}

\begin{ModelResponse}{gemini-2.5-pro}
"attempts": [
{
  "attempt_idx": 0,
  "model_output": {
    "primary_clues": [
      "(@\foreignlanguage{arabic}{گرگ}@)", "(@\foreignlanguage{arabic}{باران}@)"
    ],
    "candidates": [
      "(@\foreignlanguage{arabic}{گرگ باران دیده}@)", "(@\foreignlanguage{arabic}{گرگ بارش دیده}@)", "(@\foreignlanguage{arabic}{گرگ در باران}@)"
    ],
    "final_answer": "(@\foreignlanguage{arabic}{گرگ بارش دیده}@)"
  },
  "feedback": "The answer '(@\foreignlanguage{arabic}{گرگ بارش دیده}@)' is incorrect. Please Step-by-Step."
},
{
  "attempt_idx": 1,
  "model_output": {
    "primary_clues": [
      "(@\foreignlanguage{arabic}{گرگ}@)", "(@\foreignlanguage{arabic}{باران}@)"
    ],
    "candidates": [
      "(@\foreignlanguage{arabic}{گرگ باران خور}@)", "(@\foreignlanguage{arabic}{گرگ باران دیده}@)", "(@\foreignlanguage{arabic}{گرگ تگرگ دیده}@)"
    ],
    "final_answer": "(@\foreignlanguage{arabic}{گرگ باران خور}@)"
  },
  "feedback": "The answer '(@\foreignlanguage{arabic}{گرگ باران خور}@)' is incorrect. Please Step-by-Step."
},
{
  "attempt_idx": 2,
  "model_output": {
    "primary_clues": [
      "(@\foreignlanguage{arabic}{گرگ}@)", "(@\foreignlanguage{arabic}{باران}@)"
    ],
    "candidates": [
      "(@\foreignlanguage{arabic}{گرگ باران دیده}@)", "(@\foreignlanguage{arabic}{گرگ تگرگ دیده}@)", "(@\foreignlanguage{arabic}{گرگینه}@)"
    ],
    "final_answer": "(@\foreignlanguage{arabic}{گرگ باران دیده}@)"
  }
}
\end{ModelResponse}

\begin{ModelResponse}{grok-4.1-fast}
"attempts": [
{
  "attempt_idx": 0,
  "primary_clues": [
    "(@\foreignlanguage{arabic}{گرگ}@)", "(@\foreignlanguage{arabic}{باران}@)", "(@\foreignlanguage{arabic}{عضله/رگ}@)"
  ],
  "candidates": [
    "(@\foreignlanguage{arabic}{باران سگ گربه}@)", "(@\foreignlanguage{arabic}{باران گربه سگ}@)", "(@\foreignlanguage{arabic}{باران رگباری}@)"
  ],
  "final_answer": "(@\foreignlanguage{arabic}{باران رگباری}@)"
},
{
  "attempt_idx": 1,
  "primary_clues": [
    "(@\foreignlanguage{arabic}{گرگ/سگ}@)", "(@\foreignlanguage{arabic}{باران}@)", "(@\foreignlanguage{arabic}{ابر}@)"
  ],
  "candidates": [
    "(@\foreignlanguage{arabic}{باران سگ گربه}@)", "(@\foreignlanguage{arabic}{رگبار گرگی}@)", "(@\foreignlanguage{arabic}{گرگ بارانی}@)"
  ],
  "final_answer": "(@\foreignlanguage{arabic}{باران سگ گربه}@)"
},
{
  "attempt_idx": 2,
  "primary_clues": [
    "(@\foreignlanguage{arabic}{گرگِ خیس/باران‌خورده}@)", "(@\foreignlanguage{arabic}{باران شدید و ممتد}@)",
    "(@\foreignlanguage{arabic}{تأکید تصویری روی «زده/خورده» شدن (خیس و بارانی بودن)}@)"
  ],
  "candidates": [
    "(@\foreignlanguage{arabic}{گرگ باران زده}@)", "(@\foreignlanguage{arabic}{گرگ زیر باران}@)", "(@\foreignlanguage{arabic}{گرگ خیس}@)"
  ],
  "final_answer": "(@\foreignlanguage{arabic}{گرگ باران زده}@)"
}
\end{ModelResponse}

\begin{PromptFA}[Partial character reveal]{sidebyside, lefthand ratio=0.7}
Task: Solve the picture word puzzle from the image.\\
Target language: Persian (Farsi).\\
Hint type: shuffle\_chars.\\
Hint pattern: \mbox{\foreignlanguage{arabic}{\detokenize{_ی___ک}}}

\begin{tcolorbox}[gtstyle]
        \foreignlanguage{arabic}{سیاتیک}
\end{tcolorbox}

\tcblower

\centering
\includegraphics[width=\linewidth]{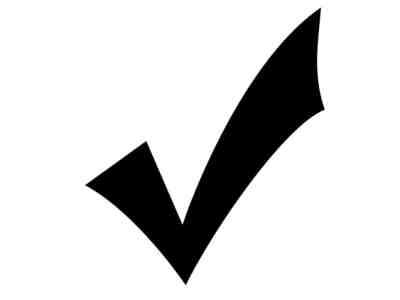}
\end{PromptFA}

\begin{ModelResponse}{gemini-2.5-pro}
{
  "primary_clues": [
    "A check mark, which in Persian is called '(@\foreignlanguage{arabic}{تیک}@)' (tik) and symbolizes correctness or understanding.",
    "A word that fits the letter pattern (a five-letter word where the second letter is '(@\foreignlanguage{arabic}{ی}@)' and the last is '(@\foreignlanguage{arabic}{ک}@)')."
  ],
  "candidates": [
    "(@\foreignlanguage{arabic}{فیزیک}@)", "(@\foreignlanguage{arabic}{کیوسک}@)", "(@\foreignlanguage{arabic}{بی‌درک}@)"
  ],
  "final_answer": "(@\foreignlanguage{arabic}{بی‌درک}@)"
}
\end{ModelResponse}

\begin{ModelResponse}{grok-4.1-fast}
{
  "primary_clues": [
    "(@\foreignlanguage{arabic}{علامت تیک}@)", "(@\foreignlanguage{arabic}{الگوی حروف}@)"
  ],
  "candidates": [
    "(@\foreignlanguage{arabic}{سیاتیک}@)", "(@\foreignlanguage{arabic}{تیک‌تاک}@)", "(@\foreignlanguage{arabic}{فیزیک}@)"
  ],
  "final_answer": "(@\foreignlanguage{arabic}{سیاتیک}@)"
}
\end{ModelResponse}

\begin{PromptFA}[Partial character reveal]{sidebyside, lefthand ratio=0.7}
Task: Solve the picture word puzzle from the image.\\
Target language: Persian (Farsi).\\
Hint type: shuffle\_chars.\\
Hint pattern: \mbox{\foreignlanguage{arabic}{\detokenize{__ ا_ آس___ _ف____}}}

\begin{tcolorbox}[gtstyle]
        \foreignlanguage{arabic}{آب از آسیاب افتادن}
\end{tcolorbox}

\tcblower

\centering
\includegraphics[width=\linewidth]{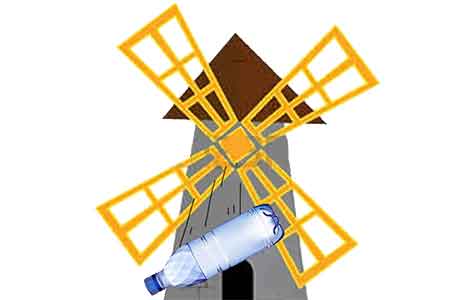}
\end{PromptFA}

\begin{ModelResponse}{gemini-2.5-pro}
{
  "primary_clues": [
    "(@\foreignlanguage{arabic}{آسیاب بادی}@) (windmill)", "(@\foreignlanguage{arabic}{بطری آب}@) (water bottle)", "(@\foreignlanguage{arabic}{افتاده}@) (fallen)"
  ],
  "candidates": [
    "(@\foreignlanguage{arabic}{از آسیاب افتادن}@)", "(@\foreignlanguage{arabic}{آب ها از آسیاب افتادن}@)", "(@\foreignlanguage{arabic}{آب از آسیاب افتاده}@)"
  ],
  "final_answer": "(@\foreignlanguage{arabic}{آب از آسیاب افتاده}@)"
}
\end{ModelResponse}
\FloatBarrier

\begin{ModelResponse}{grok-4.1-fast}
{
  "primary_clues": [
    "(@\foreignlanguage{arabic}{آب}@)", "(@\foreignlanguage{arabic}{آسیاب}@)", "(@\foreignlanguage{arabic}{افتادن}@)"
  ],
  "candidates": [
    "(@\foreignlanguage{arabic}{آب از آسیاب افتادن}@)", "(@\foreignlanguage{arabic}{آب به آسیاب ریختن}@)", "(@\foreignlanguage{arabic}{آب در آسیاب افتادن}@)"
  ],
  "final_answer": "(@\foreignlanguage{arabic}{آب از آسیاب افتادن}@)"
}
\end{ModelResponse}

\begin{PromptAR}[Basic setup]{sidebyside, lefthand ratio=0.7}
Task: Solve the picture word puzzle from the image.\\
Target language: Arabic.\\
Hint type: char\_count.

\begin{tcolorbox}[gtstyle]
        \foreignlanguage{arabic}{بطريق}
\end{tcolorbox}

\tcblower

\centering
\includegraphics[width=\linewidth]{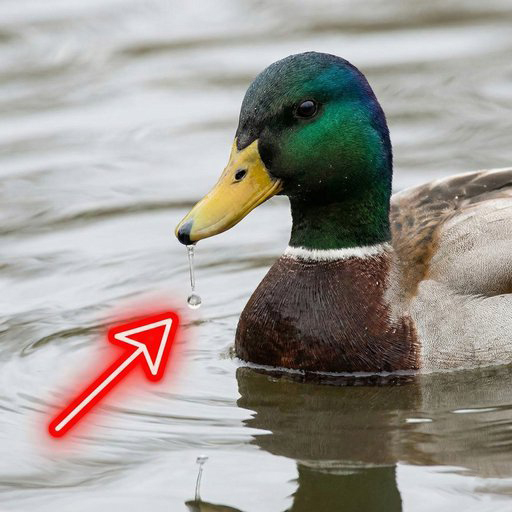}
\end{PromptAR}

\begin{ModelResponse}{gemini-2.5-pro}
{
  "primary_clues": [
    "An image of a duck, which is '(@\foreignlanguage{arabic}{بط}@)' (baTT) in Arabic.",
    "A droplet from the beak interpreted as saliva/drool, which is '(@\foreignlanguage{arabic}{ريق}@)' (reeq) in Arabic."
  ],
  "candidates": [
    "(@\foreignlanguage{arabic}{بطريق}@)", "(@\foreignlanguage{arabic}{بطوطة}@)", "(@\foreignlanguage{arabic}{تنقيط}@)"
  ],
  "final_answer": "(@\foreignlanguage{arabic}{بطريق}@)"
}
\end{ModelResponse}

\begin{ModelResponse}{grok-4.1-fast}
{
  "primary_clues": [
    "(@\foreignlanguage{arabic}{صورة لِـبط}@) (duck).", "(@\foreignlanguage{arabic}{إشارة إلى ريق كجزء لفظي مكمّل}@) (saliva/drool)."
  ],
  "candidates": [
    "(@\foreignlanguage{arabic}{بطريق}@)", "(@\foreignlanguage{arabic}{بط + ريق}@)", "(@\foreignlanguage{arabic}{ماء بط}@)"
  ],
  "final_answer": "(@\foreignlanguage{arabic}{بطريق}@)"
}
\end{ModelResponse}

\begin{PromptAR}[Basic setup]{sidebyside, lefthand ratio=0.7}
Task: Solve the picture word puzzle from the image.\\
Target language: Arabic.\\
Hint type: char\_count.

\begin{tcolorbox}[gtstyle]
        \foreignlanguage{arabic}{مرمر}
\end{tcolorbox}

\tcblower

\centering
\includegraphics[width=\linewidth]{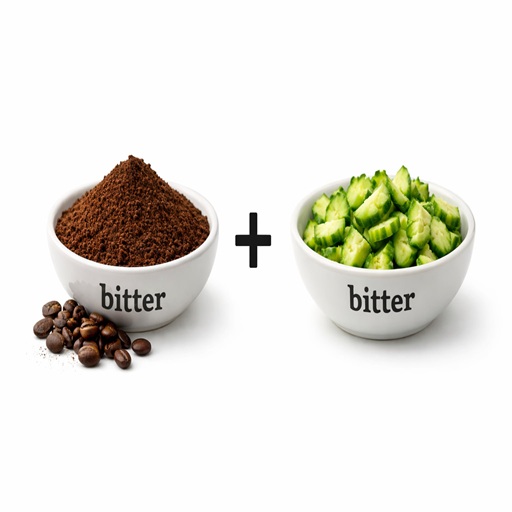}
\end{PromptAR}

\begin{ModelResponse}{gemini-2.5-pro}
{
  "primary_clues": [
    "(@\foreignlanguage{arabic}{قهوة مُرّة}@) (bitter coffee).", "(@\foreignlanguage{arabic}{خيار مُر}@) (bitter cucumber).",
    "The clue '(@\foreignlanguage{arabic}{مُر}@)' (bitter) appears twice, suggesting repetition."
  ],
  "candidates": [
    "(@\foreignlanguage{arabic}{مرمر}@)", "(@\foreignlanguage{arabic}{مرارة}@)", "(@\foreignlanguage{arabic}{مرير}@)"
  ],
  "final_answer": "(@\foreignlanguage{arabic}{مرارة}@)"
}
\end{ModelResponse}

\begin{ModelResponse}{grok-4.1-fast}
{
  "primary_clues": [
    "(@\foreignlanguage{arabic}{قهوة مُرّة}@)", "(@\foreignlanguage{arabic}{خيار مُرّ}@)", "(@\foreignlanguage{arabic}{تكرار معنى مُرّ مرتين}@)"
  ],
  "candidates": [
    "(@\foreignlanguage{arabic}{مرمر}@)", "(@\foreignlanguage{arabic}{مرير}@)", "(@\foreignlanguage{arabic}{مرار}@)"
  ],
  "final_answer": "(@\foreignlanguage{arabic}{مرمر}@)"
}
\end{ModelResponse}

\begin{PromptAR}[Few-shot chain-of-thought prompting]{sidebyside, lefthand ratio=0.7}
Task: Solve the picture word puzzle from the image.\\
Target language: Arabic.\\
Hint type: char\_count.

\begin{tcolorbox}[gtstyle]
        \foreignlanguage{arabic}{سمسم}
\end{tcolorbox}

\tcblower

\centering
\includegraphics[width=\linewidth]{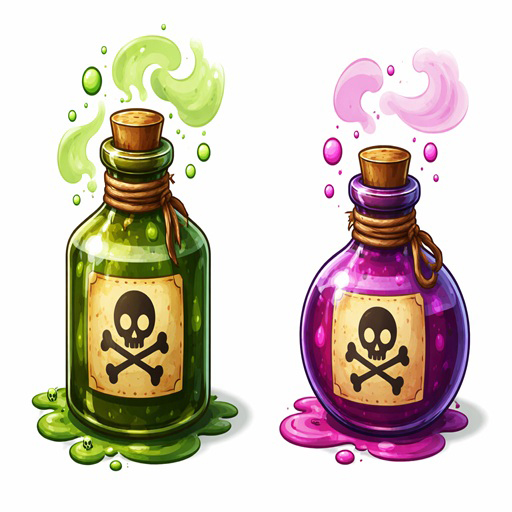}
\end{PromptAR}

\begin{ModelResponse}{gemini-2.5-pro}
{
  "primary_clues": [
    "The image shows two bottles, each marked with a skull and crossbones, indicating they contain poison.",
    "The Arabic word for poison is '(@\foreignlanguage{arabic}{سم}@)' (samm).",
    "The presence of two poison bottles suggests a repetition of the word '(@\foreignlanguage{arabic}{سم}@)'.",
    "Repeating '(@\foreignlanguage{arabic}{سم}@)' twice creates the word '(@\foreignlanguage{arabic}{سمسم}@)' (simsim), which means sesame."
  ],
  "candidates": [
    "(@\foreignlanguage{arabic}{سمسم}@)", "(@\foreignlanguage{arabic}{سموم}@)", "(@\foreignlanguage{arabic}{سمان}@)"
  ],
  "final_answer": "(@\foreignlanguage{arabic}{سمسم}@)"
}
\end{ModelResponse}

\begin{ModelResponse}{grok-4.1-fast}
{
  "primary_clues": [
    "(@\foreignlanguage{arabic}{زجاجتان عليهما علامة خطر وتدلان على السم (سم).}@)", "(@\foreignlanguage{arabic}{وجود زجاجتين يعني تكرار كلمة سم مرتين.}@)"
  ],
  "candidates": [
    "(@\foreignlanguage{arabic}{سمسم}@)", "(@\foreignlanguage{arabic}{سموم}@)", "(@\foreignlanguage{arabic}{سمين}@)"
  ],
  "final_answer": "(@\foreignlanguage{arabic}{سمسم}@)"
}
\end{ModelResponse}

\begin{PromptAR}[Few-shot chain-of-thought prompting]{sidebyside, lefthand ratio=0.7}
Task: Solve the picture word puzzle from the image.\\
Target language: Arabic.\\
Hint type: char\_count.

\begin{tcolorbox}[gtstyle]
        \foreignlanguage{arabic}{منفجر}
\end{tcolorbox}

\tcblower

\centering
\includegraphics[width=\linewidth]{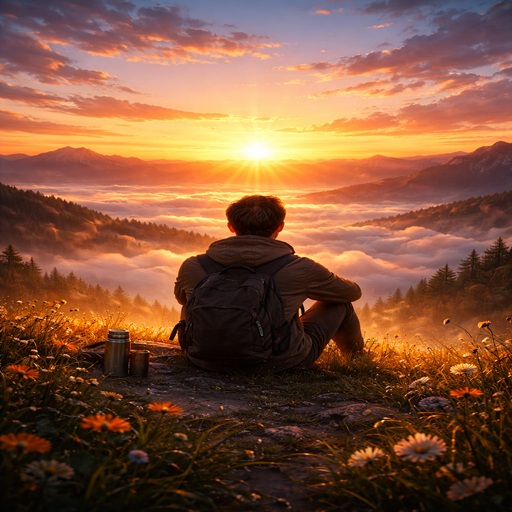}
\end{PromptAR}

\begin{ModelResponse}{gemini-2.5-pro}
{
  "primary_clues": [
    "The image shows a person sitting still and looking towards the horizon.",
    "A sunrise is in progress, which is an event one typically waits for and watches.",
    "The person's posture suggests a state of waiting and anticipation.",
    "The Arabic word for 'one who waits' or 'expectant' is 'muntazir' ((@\foreignlanguage{arabic}{منتظر}@))."
  ],
  "candidates": [
    "(@\foreignlanguage{arabic}{منتظر}@)", "(@\foreignlanguage{arabic}{مسافر}@)", "(@\foreignlanguage{arabic}{شروقي}@)"
  ],
  "final_answer": "(@\foreignlanguage{arabic}{منتظر}@)"
}
\end{ModelResponse}

\begin{ModelResponse}{grok-4.1-fast}
{
  "primary_clues": [
    "A person/man in the scene, suggesting the Arabic component (@\foreignlanguage{arabic}{من}@).",
    "The scene is at dawn/sunrise over clouds, suggesting (@\foreignlanguage{arabic}{فجر}@).",
    "Combining (@\foreignlanguage{arabic}{من}@) + (@\foreignlanguage{arabic}{فجر}@) yields the 5-letter word (@\foreignlanguage{arabic}{منفجر}@)."
  ],
  "candidates": [
    "(@\foreignlanguage{arabic}{منفجر}@)", "(@\foreignlanguage{arabic}{مغامر}@)", "(@\foreignlanguage{arabic}{مسافر}@)"
  ],
  "final_answer": "(@\foreignlanguage{arabic}{منفجر}@)"
}
\end{ModelResponse}

\begin{PromptAR}[Iterative refinement]{sidebyside, lefthand ratio=0.7}
Task: Solve the picture word puzzle from the image.\\
Target language: Arabic.\\
Hint type: char\_count.
\begin{tcolorbox}[gtstyle]
        \foreignlanguage{arabic}{دبابة}
\end{tcolorbox}

\tcblower

\centering
\includegraphics[width=\linewidth]{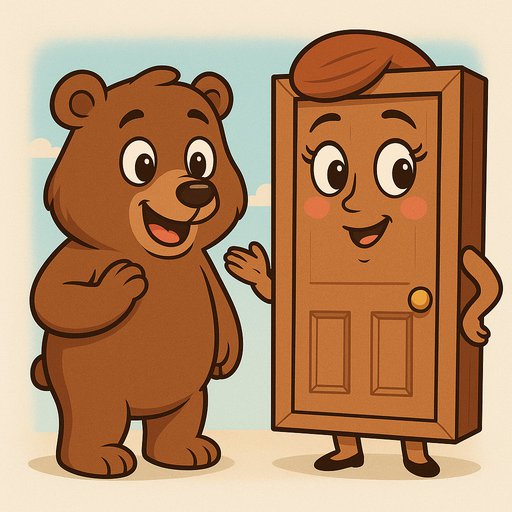}
\end{PromptAR}

\begin{ModelResponse}{gemini-2.5-pro}
"attempts": [
{
  "attempt_idx": 0,
  "model_output": {
    "primary_clues": [
      "(@\foreignlanguage{arabic}{دب (bear)}@)", "(@\foreignlanguage{arabic}{باب (door) personified as female}@)"
    ],
    "candidates": [
      "(@\foreignlanguage{arabic}{دبابة}@)", "(@\foreignlanguage{arabic}{دبدوب}@)", "(@\foreignlanguage{arabic}{باب الدب}@)"
    ],
    "final_answer": "(@\foreignlanguage{arabic}{دبابة}@)"
  }
}
]
\end{ModelResponse}

\begin{ModelResponse}{grok-4.1-fast}
"attempts":
{
  "attempt_idx": 0,
  "model_output": {
    "primary_clues": [
      "(@\foreignlanguage{arabic}{دب}@)", "(@\foreignlanguage{arabic}{باب}@)"
    ],
    "candidates": [
      "(@\foreignlanguage{arabic}{دب باب}@)", "(@\foreignlanguage{arabic}{دُبَّاب}@)", "(@\foreignlanguage{arabic}{دبابة}@)"
    ],
    "final_answer": "(@\foreignlanguage{arabic}{دبابة}@)"
  }
}
\end{ModelResponse}

\begin{PromptAR}[Iterative refinement]{sidebyside, lefthand ratio=0.7}
Task: Solve the picture word puzzle from the image.\\
Target language: Arabic.\\
Hint type: char\_count.

\begin{tcolorbox}[gtstyle]
        \foreignlanguage{arabic}{بيتوتة}
\end{tcolorbox}

\tcblower

\centering
\includegraphics[width=\linewidth]{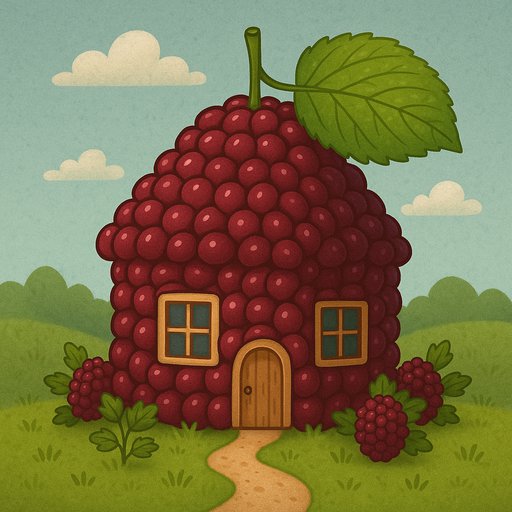}
\end{PromptAR}

\begin{ModelResponse}{gemini-2.5-pro}
"attempts":
{
  "attempt_idx": 0,
  "model_output": {
    "primary_clues": [
      "(@\foreignlanguage{arabic}{بيت}@)", "(@\foreignlanguage{arabic}{توت}@)"
    ],
    "candidates": [
      "(@\foreignlanguage{arabic}{بيت توت}@)", "(@\foreignlanguage{arabic}{بيروت}@)", "(@\foreignlanguage{arabic}{دار التوت}@)"
    ],
    "final_answer": "(@\foreignlanguage{arabic}{بيت توت}@)"
  },
  "feedback": "The answer '(@\foreignlanguage{arabic}{بيت توت}@)' is incorrect."
},
{
  "attempt_idx": 1,
  "model_output": {
    "primary_clues": [
      "(@\foreignlanguage{arabic}{بيت}@)", "(@\foreignlanguage{arabic}{توت}@)"
    ],
    "candidates": [
      "(@\foreignlanguage{arabic}{بيروت}@)", "(@\foreignlanguage{arabic}{بيت توت}@)", "(@\foreignlanguage{arabic}{دار التوت}@)"
    ],
    "final_answer": "(@\foreignlanguage{arabic}{بيروت}@)"
  },
  "feedback": "The answer '' is incorrect."
},
{
  "attempt_idx": 2,
  "model_output": {
    "primary_clues": [
      "(@\foreignlanguage{arabic}{كوخ}@)", "(@\foreignlanguage{arabic}{توت}@)"
    ],
    "candidates": [
      "(@\foreignlanguage{arabic}{كوخ توت}@)", "(@\foreignlanguage{arabic}{بيت توت}@)", "(@\foreignlanguage{arabic}{بيروت}@)"
    ],
    "final_answer": "(@\foreignlanguage{arabic}{كوخ توت}@)"
  }
}
\end{ModelResponse}

\begin{ModelResponse}{grok-4.1-fast}
"attempts":
{
  "attempt_idx": 0,
  "model_output": {
    "primary_clues": [
      "(@\foreignlanguage{arabic}{بيت}@)", "(@\foreignlanguage{arabic}{عنب}@)"
    ],
    "candidates": [
      "(@\foreignlanguage{arabic}{بيت عنب}@)", "(@\foreignlanguage{arabic}{عنب بيت}@)", "(@\foreignlanguage{arabic}{بيت العنب}@)"
    ],
    "final_answer": "(@\foreignlanguage{arabic}{بيت عنب}@)"
  },
  "feedback": "The answer '(@\foreignlanguage{arabic}{بيت عنب}@)' is incorrect."
},
{
  "attempt_idx": 1,
  "model_output": {
    "primary_clues": [
      "(@\foreignlanguage{arabic}{كوخ}@)", "(@\foreignlanguage{arabic}{عنب}@)"
    ],
    "candidates": [
      "(@\foreignlanguage{arabic}{كوخ عنب}@)", "(@\foreignlanguage{arabic}{قبة عنب}@)", "(@\foreignlanguage{arabic}{دار عنب}@)", "(@\foreignlanguage{arabic}{قصر عنب}@)"
    ],
    "final_answer": "(@\foreignlanguage{arabic}{كوخ عنب}@)"
  },
  "feedback": "The answer '(@\foreignlanguage{arabic}{كوخ عنب}@)' is incorrect."
},
{
  "attempt_idx": 2,
  "model_output": {
    "primary_clues": [
      "(@\foreignlanguage{arabic}{دار}@)", "(@\foreignlanguage{arabic}{عنب}@)"
    ],
    "candidates": [
      "(@\foreignlanguage{arabic}{دار عنب}@)", "(@\foreignlanguage{arabic}{عنب دار}@)", "(@\foreignlanguage{arabic}{دار العنب}@)"
    ],
    "final_answer": "(@\foreignlanguage{arabic}{دار عنب}@)"
  }
}
\end{ModelResponse}

\begin{PromptAR}[Partial character reveal]{sidebyside, lefthand ratio=0.7}
Task: Solve the picture word puzzle from the image.\\
Target language: Arabic.\\
Hint type: shuffle\_chars.\\
Hint pattern: \mbox{\foreignlanguage{arabic}{\detokenize{_ر__}}}

\begin{tcolorbox}[gtstyle]
        \foreignlanguage{arabic}{مرمر}
\end{tcolorbox}

\tcblower

\centering
\includegraphics[width=\linewidth]{figures/ar_43.jpg}
\end{PromptAR}

\begin{ModelResponse}{gemini-2.5-pro}
{
  "primary_clues": [
    "A bitter substance (e.g., coffee), suggesting the Arabic word root '(@\foreignlanguage{arabic}{مر}@)'.",
    "Another bitter item (e.g., cucumber), reinforcing '(@\foreignlanguage{arabic}{مر}@)'.",
    "Repetition of '(@\foreignlanguage{arabic}{مر}@)' (bitter) indicated by the two bitter clues."
  ],
  "candidates": [
    "(@\foreignlanguage{arabic}{مرمر}@)", "(@\foreignlanguage{arabic}{مرارة}@)", "(@\foreignlanguage{arabic}{مرير}@)"
  ],
  "final_answer": "(@\foreignlanguage{arabic}{مرمر}@)"
}
\end{ModelResponse}

\begin{ModelResponse}{grok-4.1-fast}
{
  "primary_clues": [
    "(@\foreignlanguage{arabic}{قهوة مرّ}@)", "(@\foreignlanguage{arabic}{خيار مرّ}@)", "(@\foreignlanguage{arabic}{علامة + بينهما}@)"
  ],
  "candidates": [
    "(@\foreignlanguage{arabic}{مر مر}@)", "(@\foreignlanguage{arabic}{مرمر}@)", "(@\foreignlanguage{arabic}{مرارة}@)"
  ],
  "final_answer": "(@\foreignlanguage{arabic}{مرمر}@)"
}
\end{ModelResponse}

\begin{PromptAR}[Partial character reveal]{sidebyside, lefthand ratio=0.7}
Task: Solve the picture word puzzle from the image.\\
Target language: Arabic.\\
Hint type: shuffle\_chars.\\
Hint pattern: \mbox{\foreignlanguage{arabic}{\detokenize{_و___}}}

\begin{tcolorbox}[gtstyle]
        \foreignlanguage{arabic}{سودان}
\end{tcolorbox}

\tcblower

\centering
\includegraphics[width=\linewidth]{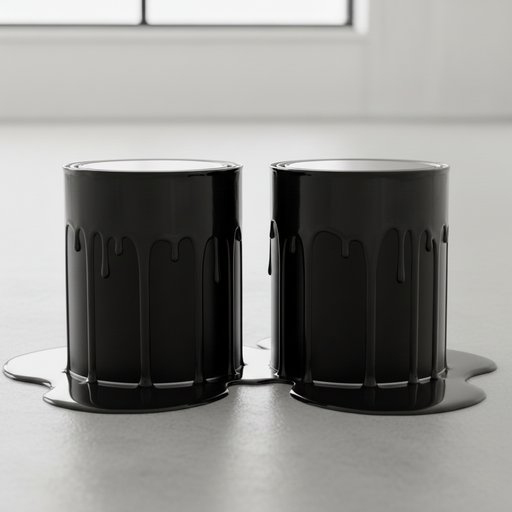}
\end{PromptAR}

\begin{ModelResponse}{gemini-2.5-pro}
{
  "primary_clues": [
    "Two containers/cans: (@\foreignlanguage{arabic}{علبتان}@)", "Black color: (@\foreignlanguage{arabic}{أسود}@)",
    "The hint pattern matches a 5-letter word with (@\foreignlanguage{arabic}{و}@) as the second letter."
  ],
  "candidates": [
    "(@\foreignlanguage{arabic}{سودان}@)", "(@\foreignlanguage{arabic}{سوداء}@)", "(@\foreignlanguage{arabic}{سواد}@)"
  ],
  "final_answer": "(@\foreignlanguage{arabic}{سوداء}@)"
}
\end{ModelResponse}

\begin{ModelResponse}{grok-4.1-fast}
{
  "primary_clues": [
    "(@\foreignlanguage{arabic}{علب طلاء}@)", "(@\foreignlanguage{arabic}{لون أسود}@)", "(@\foreignlanguage{arabic}{اثنان}@)"
  ],
  "candidates": [
    "(@\foreignlanguage{arabic}{سوداء}@)", "(@\foreignlanguage{arabic}{سودان}@)", "(@\foreignlanguage{arabic}{سواد}@)"
  ],
  "final_answer": "(@\foreignlanguage{arabic}{سودان}@)"
}
\end{ModelResponse}

\begin{PromptCL}[Basic setup]{sidebyside, lefthand ratio=0.7}
Task: Solve the picture word puzzle from the image.\\
Target language: Persian (Farsi).\\
Hint type: char\_count.\\
Hint: The answer has 9 characters (excluding spaces).

\begin{tcolorbox}[gtstyle]
        \foreignlanguage{arabic}{کلم بروکلی}
\end{tcolorbox}

\tcblower

\centering
\includegraphics[width=\linewidth]{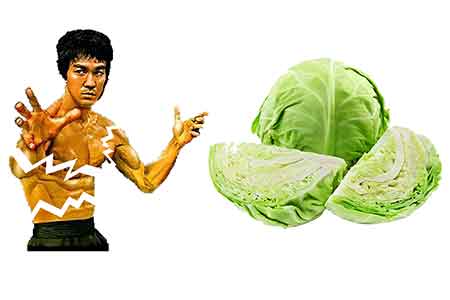}
\end{PromptCL}

\begin{ModelResponse}{gemini-2.5-pro}
{
  "primary_clues": [
    "Bruce Lee", "Cabbage"
  ],
  "candidates": [
    "(@\foreignlanguage{arabic}{کلم بروسلی}@)", "(@\foreignlanguage{arabic}{بروکلی}@)", "(@\foreignlanguage{arabic}{کلم لی}@)"
  ],
  "final_answer": "(@\foreignlanguage{arabic}{کلم بروسلی}@)"
}
\end{ModelResponse}

\begin{ModelResponse}{grok-4.1-fast}
{
  "primary_clues": [
    "(@\foreignlanguage{arabic}{کلم}@)", "(@\foreignlanguage{arabic}{بروس لی}@)"
  ],
  "candidates": [
    "(@\foreignlanguage{arabic}{کلم بروس}@)", "(@\foreignlanguage{arabic}{بروکلی}@)", "(@\foreignlanguage{arabic}{کلم بروکلی}@)"
  ],
  "final_answer": "(@\foreignlanguage{arabic}{کلم بروکلی}@)"
}
\end{ModelResponse}

\begin{PromptCL}[Example: Basic setup]{sidebyside, lefthand ratio=0.7}
Task: Solve the picture word puzzle from the image.\\
Target language: Persian (Farsi).\\
Hint type: char\_count.

\begin{tcolorbox}[gtstyle]
        \foreignlanguage{arabic}{استحکام}
\end{tcolorbox}

\tcblower

\centering
\includegraphics[width=\linewidth]{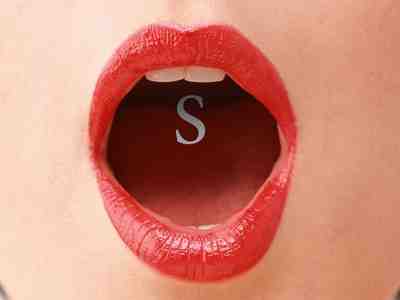}
\end{PromptCL}

\begin{ModelResponse}{gemini-2.5-pro}
{
  "primary_clues": [
    "(@\foreignlanguage{arabic}{حرف S}@)", "(@\foreignlanguage{arabic}{دهان}@)"
  ],
  "candidates": [
    "(@\foreignlanguage{arabic}{اصفهان}@)", "(@\foreignlanguage{arabic}{دهن سرویس}@)", "(@\foreignlanguage{arabic}{سخندان}@)"
  ],
  "final_answer": "(@\foreignlanguage{arabic}{اصفهان}@)"
}
\end{ModelResponse}

\begin{ModelResponse}{grok-4.1-fast}
{
  "primary_clues": [
    "(@\foreignlanguage{arabic}{دهان}@)", "(@\foreignlanguage{arabic}{حرف S}@)"
  ],
  "candidates": [
    "(@\foreignlanguage{arabic}{استحکام}@)", "(@\foreignlanguage{arabic}{استحصال}@)", "(@\foreignlanguage{arabic}{استفهام}@)"
  ],
  "final_answer": "(@\foreignlanguage{arabic}{استحکام}@)"
}
\end{ModelResponse}

\begin{PromptCL}[Example: Few-shot chain-of-thought prompting]{sidebyside, lefthand ratio=0.7}
Task: Solve the picture word puzzle from the image.\\
Target language: Persian (Farsi).\\
Hint type: char\_count.

\begin{tcolorbox}[gtstyle]
        \foreignlanguage{arabic}{دوربین}
\end{tcolorbox}

\tcblower

\centering
\includegraphics[width=\linewidth]{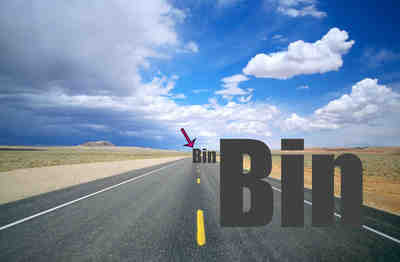}
\end{PromptCL}

\begin{ModelResponse}{gemini-2.5-pro}
{
  "primary_clues": [
    "The image shows the English word 'Bin', which sounds like the Persian root '(@\foreignlanguage{arabic}{بین}@)' (bin), meaning 'to see'.",
    "The word 'Bin' is shown receding into the distance, with an arrow pointing to the furthest one, representing 'far' or '(@\foreignlanguage{arabic}{دور}@)' (dur) in Persian."
  ],
  "candidates": [
    "(@\foreignlanguage{arabic}{دوربین}@)", "(@\foreignlanguage{arabic}{ریزبین}@)", "(@\foreignlanguage{arabic}{بین راه}@)"
  ],
  "final_answer": "(@\foreignlanguage{arabic}{دوربین}@)"
}
\end{ModelResponse}

\begin{ModelResponse}{grok-4.1-fast}
{
  "primary_clues": [
    "(@\foreignlanguage{arabic}{کلمهٔ BIN}@)", "(@\foreignlanguage{arabic}{جاده/راهِ طولانی (دور)}@)"
  ],
  "candidates": [
    "(@\foreignlanguage{arabic}{بین‌راه}@)", "(@\foreignlanguage{arabic}{دورراه}@)", "(@\foreignlanguage{arabic}{دوربین}@)"
  ],
  "final_answer": "(@\foreignlanguage{arabic}{دوربین}@)"
}
\end{ModelResponse}

\begin{PromptCL}[Example: Few-shot chain-of-thought prompting]{sidebyside, lefthand ratio=0.7}
Task: Solve the picture word puzzle from the image.\\
Target language: Persian (Farsi).\\
Hint type: char\_count.

\begin{tcolorbox}[gtstyle]
        \foreignlanguage{arabic}{دیباچه}
\end{tcolorbox}

\tcblower

\centering
\includegraphics[width=\linewidth]{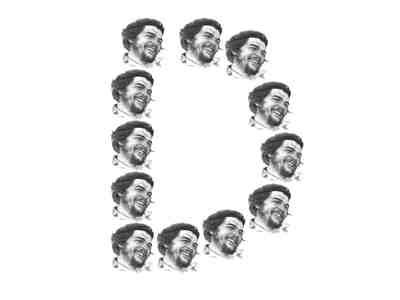}
\end{PromptCL}

\begin{ModelResponse}{gemini-2.5-pro}
{
  "primary_clues": [
    "The image consists of 12 repeated pictures of a laughing face.",
    "The number 12 in Persian is 'Davazdah' ((@\foreignlanguage{arabic}{دوازده}@)).",
    "The pictures are arranged in the shape of the letter 'D', which is the first letter of 'Davazdah'.",
    "The word 'Davazdah' ((@\foreignlanguage{arabic}{دوازده}@)) has 6 characters, which matches the hint provided."
  ],
  "candidates": [
    "(@\foreignlanguage{arabic}{شادی}@)", "(@\foreignlanguage{arabic}{بهروزها}@)", "(@\foreignlanguage{arabic}{دوازده}@)"
  ],
  "final_answer": "(@\foreignlanguage{arabic}{دوازده}@)"
}
\end{ModelResponse}

\begin{ModelResponse}{grok-4.1-fast}
{
  "primary_clues": [
    "(@\foreignlanguage{arabic}{حرف D (خوانشِ فارسی: «دی»)}@)", "(@\foreignlanguage{arabic}{صورت‌های خندانِ زیاد (بچه‌ها)}@)"
  ],
  "candidates": [
    "(@\foreignlanguage{arabic}{دیوانه}@)", "(@\foreignlanguage{arabic}{دیبچه}@)", "(@\foreignlanguage{arabic}{دیباچه}@)"
  ],
  "final_answer": "(@\foreignlanguage{arabic}{دیباچه}@)"
}
\end{ModelResponse}

\begin{PromptCL}[Iterative refinement]{sidebyside, lefthand ratio=0.7}
Task: Solve the picture word puzzle from the image.\\
Target language: Persian (Farsi).\\
Hint type: char\_count.

\begin{tcolorbox}[gtstyle]
        \foreignlanguage{arabic}{کی می خوابی}
\end{tcolorbox}

\tcblower

\centering
\includegraphics[width=\linewidth]{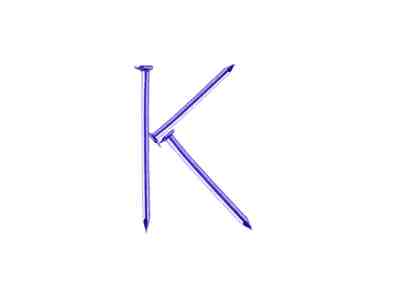}
\end{PromptCL}

\begin{ModelResponse}{gemini-2.5-pro}
"attempts":
{
  "attempt_idx": 0,
  "model_output": {
    "primary_clues": [
      "Nails, which in Persian are '(@\foreignlanguage{arabic}{میخ}@)', arranged as the letter 'K' / '(@\foreignlanguage{arabic}{ک}@)'.",
      "A violet/purple color, which in Persian is '(@\foreignlanguage{arabic}{بنفش}@)'."
    ],
    "candidates": [
      "(@\foreignlanguage{arabic}{میخک}@)", "(@\foreignlanguage{arabic}{میخک بنفش}@)", "(@\foreignlanguage{arabic}{میخک بنفشه}@)"
    ],
    "final_answer": "(@\foreignlanguage{arabic}{میخک بنفشه}@)"
  },
  "feedback": "The answer '(@\foreignlanguage{arabic}{میخک بنفشه}@)' is incorrect."
},
{
  "attempt_idx": 1,
  "model_output": {
    "primary_clues": [
      "The letter 'K'", "Nails '(@\foreignlanguage{arabic}{میخ}@)'", "Purple color '(@\foreignlanguage{arabic}{بنفش}@)'."
    ],
    "candidates": [
      "(@\foreignlanguage{arabic}{میخک بنفشه}@)", "(@\foreignlanguage{arabic}{کاوه آهنگر}@)", "(@\foreignlanguage{arabic}{میخک ارغوانی}@)"
    ],
    "final_answer": "(@\foreignlanguage{arabic}{کاوه آهنگر}@)"
  },
  "feedback": "The answer '' is incorrect."
},
{
  "attempt_idx": 2,
  "model_output": {
    "primary_clues": [
      "Nails '(@\foreignlanguage{arabic}{میخ}@)' arranged as the letter 'K' / '(@\foreignlanguage{arabic}{ک}@)'.", "A distinct purple tone associated with '(@\foreignlanguage{arabic}{ارغوانی}@)'."
    ],
    "candidates": [
      "(@\foreignlanguage{arabic}{میخک بنفشه}@)", "(@\foreignlanguage{arabic}{کاوه آهنگر}@)", "(@\foreignlanguage{arabic}{میخک ارغوان}@)"
    ],
    "final_answer": "(@\foreignlanguage{arabic}{میخک ارغوان}@)"
  }
}
\end{ModelResponse}

\begin{ModelResponse}{grok-4.1-fast}
"attempts": [
{
  "attempt_idx": 0,
  "primary_clues": [
    "(@\foreignlanguage{arabic}{حرف K}@)", "(@\foreignlanguage{arabic}{میخ}@)"
  ],
  "candidates": [
    "(@\foreignlanguage{arabic}{کتیبه میخی}@)", "(@\foreignlanguage{arabic}{نوشته میخی}@)", "(@\foreignlanguage{arabic}{الواح میخی}@)"
  ],
  "final_answer": "(@\foreignlanguage{arabic}{کتیبه میخی}@)"
},
{
  "attempt_idx": 1,
  "primary_clues": [
    "(@\foreignlanguage{arabic}{K = کی}@)", "(@\foreignlanguage{arabic}{میخ}@)"
  ],
  "candidates": [
    "(@\foreignlanguage{arabic}{کی میخ}@)", "(@\foreignlanguage{arabic}{کی می‌خ}@)", "(@\foreignlanguage{arabic}{کی می‌خوابی}@)"
  ],
  "final_answer": "(@\foreignlanguage{arabic}{کی میخ}@)"
},
{
  "attempt_idx": 2,
  "primary_clues": [
    "(@\foreignlanguage{arabic}{K = کی}@)", "(@\foreignlanguage{arabic}{میخ (میخ) = «می‌خ…» شروعِ فعل «خوابی»}@)"
  ],
  "candidates": [
    "(@\foreignlanguage{arabic}{کی می‌خوابی}@)", "(@\foreignlanguage{arabic}{کی می‌خوابید}@)", "(@\foreignlanguage{arabic}{کی می‌خوابن}@)"
  ],
  "final_answer": "(@\foreignlanguage{arabic}{کی می‌خوابی}@)"
}
\end{ModelResponse}

\begin{PromptCL}[Iterative refinement]{sidebyside, lefthand ratio=0.7}
Task: Solve the picture word puzzle from the image.\\
Target language: Persian (Farsi).\\
Hint type: char\_count.

\begin{tcolorbox}[gtstyle]
        \foreignlanguage{arabic}{رمضانی}
\end{tcolorbox}

\tcblower

\centering
\includegraphics[width=\linewidth]{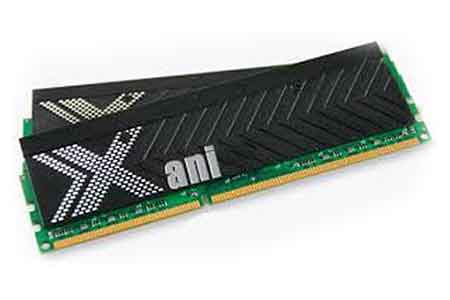}
\end{PromptCL}

\begin{ModelResponse}{gemini-2.5-pro}
"attempts":
{
  "attempt_idx": 0,
  "model_output": {
    "primary_clues": [
      "A RAM memory stick, which is associated with Persian '(@\foreignlanguage{arabic}{رمل}@)' (raml).",
      "The text 'Xani', interpreted as a pun on Persian '(@\foreignlanguage{arabic}{زنی}@)'."
    ],
    "candidates": [
      "(@\foreignlanguage{arabic}{رمل‌زنی}@)", "(@\foreignlanguage{arabic}{حافظه‌خوانی}@)", "(@\foreignlanguage{arabic}{رمالی}@)"
    ],
    "final_answer": "(@\foreignlanguage{arabic}{رمل‌زنی}@)"
  },
  "feedback": "The answer '(@\foreignlanguage{arabic}{رمل‌زنی}@)' is incorrect. Please Step-by-Step."
},
{
  "attempt_idx": 1,
  "model_output": {
    "primary_clues": [
      "A RAM stick (computer memory), i.e., Persian '(@\foreignlanguage{arabic}{حافظه}@)'.",
      "The text 'Xani' as a pun on Persian '(@\foreignlanguage{arabic}{خوانی}@)' (reading), with 'X' used for the '(@\foreignlanguage{arabic}{خ}@)' sound."
    ],
    "candidates": [
      "(@\foreignlanguage{arabic}{حافظه‌خوانی}@)", "(@\foreignlanguage{arabic}{رم‌خوانی}@)", "(@\foreignlanguage{arabic}{رمل‌زنی}@)"
    ],
    "final_answer": "(@\foreignlanguage{arabic}{حافظه‌خوانی}@)"
  },
  "feedback": "The answer '(@\foreignlanguage{arabic}{حافظه‌خوانی}@)' is incorrect. Please Step-by-Step."
},
{
  "attempt_idx": 2,
  "model_output": {
    "primary_clues": [
      "A RAM memory stick, sounding like Persian '(@\foreignlanguage{arabic}{رمل}@)'.",
      "The text 'Xani' as a pun for Persian '(@\foreignlanguage{arabic}{خوانی}@)' (khani), using 'X' for '(@\foreignlanguage{arabic}{خ}@)'."
    ],
    "candidates": [
      "(@\foreignlanguage{arabic}{رمل‌خوانی}@)", "(@\foreignlanguage{arabic}{رم‌خوانی}@)", "(@\foreignlanguage{arabic}{حافظه‌خوانی}@)",  "(@\foreignlanguage{arabic}{رمل‌زنی}@)"
    ],
    "final_answer": "(@\foreignlanguage{arabic}{رمل‌خوانی}@)"
  }
}
\end{ModelResponse}

\begin{ModelResponse}{grok-4.1-fast}
"attempts": [
{
  "attempt_idx": 0,
  "primary_clues": [
    "(@\foreignlanguage{arabic}{رم (RAM)}@)", "(@\foreignlanguage{arabic}{X}@)"
  ],
  "candidates": [
    "(@\foreignlanguage{arabic}{ایکس رم}@)", "(@\foreignlanguage{arabic}{رم ایکس}@)", "(@\foreignlanguage{arabic}{رمضانی}@)"
  ],
  "final_answer": "(@\foreignlanguage{arabic}{ایکس رم}@)"
},
{
  "attempt_idx": 1,
  "primary_clues": [
    "(@\foreignlanguage{arabic}{رم (RAM)}@)", "(@\foreignlanguage{arabic}{الگوی آج/طرح تایر = «ضانی»}@)"
  ],
  "candidates": [
    "(@\foreignlanguage{arabic}{رمضانی}@)", "(@\foreignlanguage{arabic}{رمزنی}@)", "(@\foreignlanguage{arabic}{رمزانی}@)"
  ],
  "final_answer": "(@\foreignlanguage{arabic}{رمضانی}@)"
},
{
  "attempt_idx": 2,
  "primary_clues": [
    "(@\foreignlanguage{arabic}{رم}@)", "(@\foreignlanguage{arabic}{ضانی (از طرح/آج)}@)"
  ],
  "candidates": [
    "(@\foreignlanguage{arabic}{رمضانی}@)", "(@\foreignlanguage{arabic}{رمزانی}@)", "(@\foreignlanguage{arabic}{رمبانی}@)"
  ],
  "final_answer": "(@\foreignlanguage{arabic}{رمضانی}@)"
}
\end{ModelResponse}

\begin{PromptCL}[Partial character reveal]{sidebyside, lefthand ratio=0.7}
Task: Solve the picture word puzzle from the image.\\
Target language: Persian (Farsi).\\
Hint type: shuffle\_chars.\\
Hint pattern: \mbox{\foreignlanguage{arabic}{\detokenize{__ت__ا_}}}

\begin{tcolorbox}[gtstyle]
        \foreignlanguage{arabic}{استحکام}
\end{tcolorbox}

\tcblower

\centering
\includegraphics[width=\linewidth]{figures/cross_102.jpg}
\end{PromptCL}

\begin{ModelResponse}{gemini-2.5-pro}
{
  "primary_clues": [
    "The letter 'S'", "An open mouth"
  ],
  "candidates": [
    "(@\foreignlanguage{arabic}{استدعا}@)", "(@\foreignlanguage{arabic}{استعفا}@)", "(@\foreignlanguage{arabic}{حرف اس}@)"
  ],
  "final_answer": "(@\foreignlanguage{arabic}{استدعا}@)"
}
\end{ModelResponse}

\begin{ModelResponse}{grok-4.1-fast}
{
  "primary_clues": [
    "(@\foreignlanguage{arabic}{حرف S که در فارسی «اِس» خوانده می‌شود}@)", "(@\foreignlanguage{arabic}{تصویر دهان/کام}@)",
    "(@\foreignlanguage{arabic}{ترکیب «اِس» + «تحکام» برای ساخت واژه «استحکام»}@)"
  ],
  "candidates": [
    "(@\foreignlanguage{arabic}{استحکام}@)", "(@\foreignlanguage{arabic}{استحکام‌دار}@)", "(@\foreignlanguage{arabic}{استوار}@)"
  ],
  "final_answer": "(@\foreignlanguage{arabic}{استحکام}@)"
}
\end{ModelResponse}

\begin{PromptCL}[Partial character reveal]{sidebyside, lefthand ratio=0.7}
Task: Solve the picture word puzzle from the image.\\
Target language: Persian (Farsi).\\
Hint type: shuffle\_chars.\\
Hint pattern: \mbox{\foreignlanguage{arabic}{\detokenize{م____ف_}}}

\begin{tcolorbox}[gtstyle]
        \foreignlanguage{arabic}{میکروفن}
\end{tcolorbox}

\tcblower

\centering
\includegraphics[width=\linewidth]{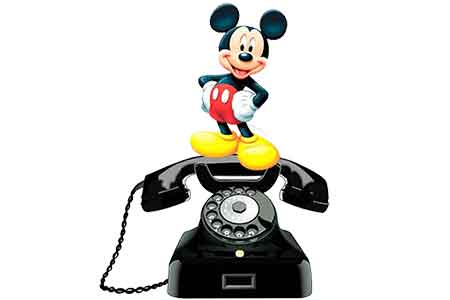}
\end{PromptCL}

\begin{ModelResponse}{gemini-2.5-pro}
{
  "primary_clues": [
    "Mickey Mouse", "Telephone"
  ],
  "candidates": [
    "(@\foreignlanguage{arabic}{میکروفون}@)", "(@\foreignlanguage{arabic}{معروف}@)", "(@\foreignlanguage{arabic}{تلفن میکی}@)"
  ],
  "final_answer": "(@\foreignlanguage{arabic}{میکروفون}@)"
}
\end{ModelResponse}

\begin{ModelResponse}{grok-4.1-fast}
{
  "primary_clues": [
    "(@\foreignlanguage{arabic}{میکی}@)", "(@\foreignlanguage{arabic}{فون}@)", "(@\foreignlanguage{arabic}{ترکیب «میکی» + «فون» برای رسیدن به «میکروفن»}@)"
  ],
  "candidates": [
    "(@\foreignlanguage{arabic}{میکروفن}@)", "(@\foreignlanguage{arabic}{میکروفون}@)",  "(@\foreignlanguage{arabic}{تلفن میکی}@)"
  ],
  "final_answer": "(@\foreignlanguage{arabic}{میکروفن}@)"
}
\end{ModelResponse}

\end{document}